\documentclass[lettersize,journal]{IEEEtran}
\usepackage{amsmath,amsfonts}
\usepackage{algorithm}
\usepackage{algpseudocode}                                  
\usepackage[commentColor=blue,beginComment=//~,beginLComment=/*~, endLComment=~*/]{algpseudocodex}
\usepackage{array}
\usepackage{textcomp}
\usepackage{stfloats}
\usepackage{url}
\usepackage{verbatim}
\usepackage{graphicx}
\usepackage{cite}
\usepackage[pagebackref=true,colorlinks,bookmarks=true,citecolor=blue,linkcolor=blue,urlcolor=blue]{hyperref}

\usepackage{url}            
\usepackage[table]{xcolor}  
\usepackage{amsfonts}       
\usepackage{nicefrac}       
\usepackage{microtype}      
\usepackage{graphicx}
\usepackage{makecell}
\usepackage{multirow}
\usepackage{dsfont}
\usepackage{siunitx}                                        
\usepackage{bm}
\usepackage{booktabs, makecell, multirow, threeparttable}   
\usepackage{overpic}
\usepackage{pict2e}
\usepackage{tikz}                                           
\usepackage{arydshln}                                       
\usepackage{amsmath}
\usepackage{enumitem}
\usepackage[short]{optidef}
\usepackage[svgnames]{xcolor}
\usepackage{rotating}
\usepackage{amssymb}
\usepackage{pifont}
\usepackage{caption}
\usepackage{subcaption}
\usepackage{ragged2e}

\usepackage{setspace}
\usepackage{listings}
\usepackage{anyfontsize}
\usepackage{color, colortbl}                                
\usepackage{titletoc}                                   

\newcommand{\PreserveBackslash}[1]{\let\temp=\\#1\let\\=\temp}
\newcolumntype{C}[1]{>{\PreserveBackslash\centering}p{#1}}
\newcolumntype{R}[1]{>{\PreserveBackslash\raggedleft}p{#1}}
\newcolumntype{L}[1]{>{\PreserveBackslash\raggedright}p{#1}}

\usepackage{transparent}
\newcommand{\semitransp}[2][0.55]{{\transparent{#1}#2}}

\usepackage[capitalize]{cleveref}
\crefname{section}{Sec.}{Secs.}
\Crefname{section}{Section}{Sections}
\Crefname{table}{Table}{Tables}
\crefname{table}{Tab.}{Tabs.}
\newcommand*\rot{\rotatebox{90}}

\definecolor{myblue}{RGB}{169,196,235}
\definecolor{mygreen}{RGB}{213,232,212}
\definecolor{mygray}{RGB}{191,191,191}
\definecolor{mycolor}{RGB}{184,96,41}

\newcommand{\smallstd}[1]{\text{\footnotesize\semitransp{$\pm~#1$}}}

\ifdefined \GramaCheck
\newcommand{\CheckRmv}[1]{}
\newcommand{\figref}[1]{Figure 1}%
\newcommand{\tabref}[1]{Table 1}%
\newcommand{\secref}[1]{Section 1}
\renewcommand{\eqref}[1]{Equation 1}
\else
\newcommand{\CheckRmv}[1]{#1}
\newcommand{\figref}[1]{Fig.~\ref{#1}}%
\newcommand{\tabref}[1]{Tab.~\ref{#1}}%
\newcommand{\secref}[1]{Sec.~\ref{#1}}

\renewcommand{\eqref}[1]{Eqn.~(\ref{#1})}
\fi

\def\ie{\emph{i.e.}}
\def\eg{\emph{e.g.}}

\begin{document}

\title{Effective Prompt Pool Learning for Continual Category Discovery}

\author{Fernando Julio Cendra, Xinghui Li, and Kai Han%
\IEEEcompsocitemizethanks{%
\IEEEcompsocthanksitem Fernando Julio Cendra and Kai Han (Senior~Member,~IEEE) are with the the University of Hong Kong, Hong Kong. \\
Corresponding author: Kai Han (e-mail: \texttt{kaihanx@hku.hk}). \\
Code is publicly available at: \url{https://visual-ai.github.io/promptccd} 
}%
}

\captionsetup{skip=0pt}
\setlength{\textfloatsep}{8.0pt plus 2.0pt minus 4.0pt}
\setlength{\floatsep}{8.0pt plus 2.0pt minus 2.0pt}
\setlength{\intextsep}{8.0pt plus 2.0pt minus 2.0pt}
\setlength{\dbltextfloatsep}{8.0pt plus 2.0pt minus 2.0pt}
\setlength{\dblfloatsep}{8.0pt plus 2.0pt minus 2.0pt}



\maketitle
\IEEEdisplaynontitleabstractindextext

\maketitle
\begin{abstract}
This paper studies effective prompt pool learning for Continual Category Discovery (CCD), a challenging open-world setting where a model must discover novel categories from a continuous stream of unlabelled data containing both known and novel classes, while mitigating catastrophic forgetting of previously learned concepts. We introduce a series of novel prompt-pool-based frameworks for CCD, each exploring a different design of prompt pools. First, we propose \textbf{PromptCCD}, which focuses on global class prototypes via a Gaussian Mixture Prompt (GMP) module. GMP fits a generative Gaussian mixture model over feature embeddings, where each mixture component serves as both a class prototype and a dynamic prompt that conditions the backbone's representations. This design enables label-free prompt selection and on-the-fly estimation of the number of emerging categories. Through a systematic spectrum study, we then show that category count, rather than sample size, is the primary bottleneck for discovery performance, motivating the need for finer-grained representations. Building on this finding, we propose \textbf{PromptCCD++}, which focuses on object-part prototypes via Part-level Prompting (PLP) modules. PLP decomposes prompt pool into multiple, specialized part-level prompt pools. During discovery phase, these pools dynamically assign part-specific prompts to local object regions without the need for manual part annotations, enabling the model to learn object-part representations that boost category discovery. Extensive evaluations on both generic and fine-grained benchmarks, supported by comprehensive ablation studies, demonstrate the effectiveness of our framework for CCD.
\end{abstract}


\section{Introduction}\label{sec:introduction}

\IEEEPARstart{T}{he} success of modern visual recognition systems is largely predicated on a \emph{closed-world} assumption: models are trained offline on exhaustive, manually curated datasets with a fixed set of categories~\cite{deng2009imagenet}. In practice, however, real-world systems encounter a continuous stream of visual data in which new categories emerge unpredictably and manual annotation is often infeasible. A truly capable agent must therefore autonomously discover and incorporate novel categories into its knowledge base without human supervision.

This necessity has catalysed the creation of the Category Discovery field~\cite{he2025category}. Early research into Novel Class Discovery (NCD)~\cite{han2019learning} and its successor, Generalized Category Discovery (GCD)~\cite{vaze2022generalized}, established methods for partitioning unlabelled data by leveraging knowledge transferred from a labelled ``seen'' set. However, these frameworks are primarily static: they assume that the entire data is available for offline processing.

\begin{figure*}[!ht]
    \centering
    \includegraphics[width=0.93\linewidth]{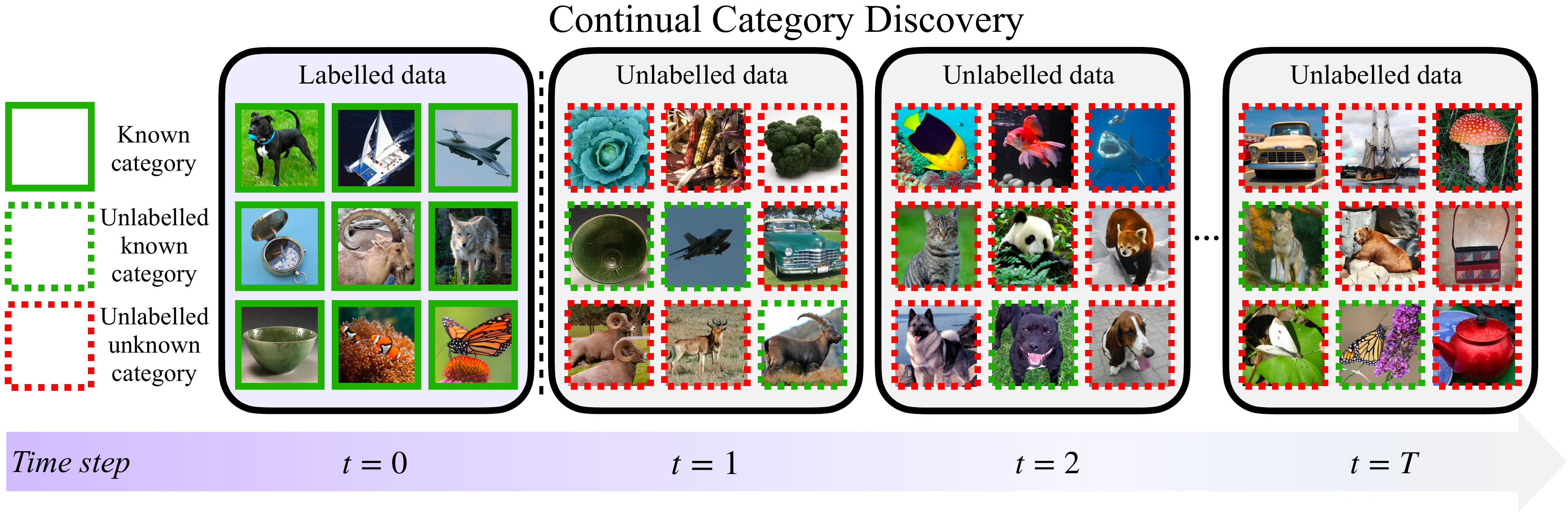}
    \caption{\textbf{Continual Category Discovery task.} In the initial stage, the model learns from labelled data, while in the subsequent stages, the model learns from a continuous data stream containing unlabelled instances from known and novel classes.}
    \label{fig:CCD_task}
\end{figure*}

To bridge this gap, the problem of \emph{Continual Category Discovery (CCD)}~\cite{zhang2022grow} extends category discovery into a continual learning framework. In CCD, a model is first trained with a labelled dataset and subsequently discovers novel categories from a continuous stream of unlabelled data (see Fig.~\ref{fig:CCD_task}). Each discovery stage may contain a mixture of known and novel categories, and the model must integrate this new information under the constraint that past raw data is inaccessible. This exposes a fundamental challenge: the model must remain flexible enough to discover novel categories while preventing catastrophic forgetting~\cite{mccloskey1989catastrophic} of previously learned knowledge, even when the number of emerging classes is unknown.

\begin{figure}[!ht]
    \centering
    \includegraphics[width=\columnwidth]{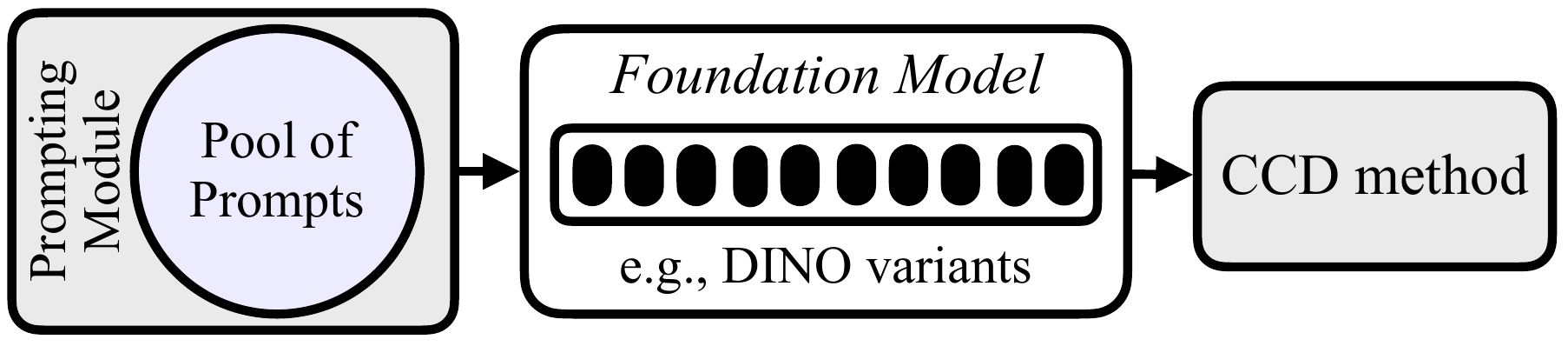}
\caption{\textbf{Prompt pool as memory in CCD.} A dynamic pool of learnable prompts is prepended to the foundation model to encode discovered category structure while preserving previously learned knowledge. Unlike supervised continual learning, CCD provides no labels during the discovery phase, making prompt selection and update without task identity or class labels a key challenge.}
    \label{fig:prompt-based ccd}
\end{figure}

A promising direction for addressing this challenge is prompt-pool-based learning. In continual learning, methods such as L2P~\cite{wang2022learning} and DualPrompt~\cite{wang2022dualprompt} attempt to design a learnable prompt pool that is prepended to a frozen foundation model, conditioning the model's representations across sequential tasks. This paradigm alleviates the need for experience replay buffers or costly parameter expansion strategies, offering a parameter-efficient mechanism for knowledge retention. An overview of this prompt-based formulation for CCD is illustrated in Fig.~\ref{fig:prompt-based ccd}. However, these prompt pool techniques are fundamentally designed for a fully supervised regime: they rely on annotated data streams to learn the mapping between task semantics and prompt selection. In the CCD setting, where incoming data is entirely unlabelled and the category composition of each stage is unknown, such supervision signals are unavailable, rendering these techniques ineffective for CCD. This raises a central question: \emph{how can we design a prompt pool that operates effectively under the open-world challenges of CCD?}

In this paper, we address this question by introducing a series of novel prompt pool designs for CCD. Our first design, \textbf{PromptCCD}, focuses on \textbf{global class prototypes}. It employs a Gaussian Mixture Prompt (GMP) module that fits a Gaussian Mixture Model (GMM) over feature embeddings, where each mixture component serves as both a class prototype and a dynamic prompt that conditions the model's representations. This design enables label-free prompt selection and facilitates knowledge retention across stages through parametric replay, where the model draws samples from previously fitted mixture components. Moreover, GMP enables on-the-fly estimation of the number of categories in the continuous stream of unlabelled data, making it applicable to realistic CCD scenarios where the category count is unknown a priori.

\begin{figure}[!htb]
    \centering
    \includegraphics[width=\columnwidth]{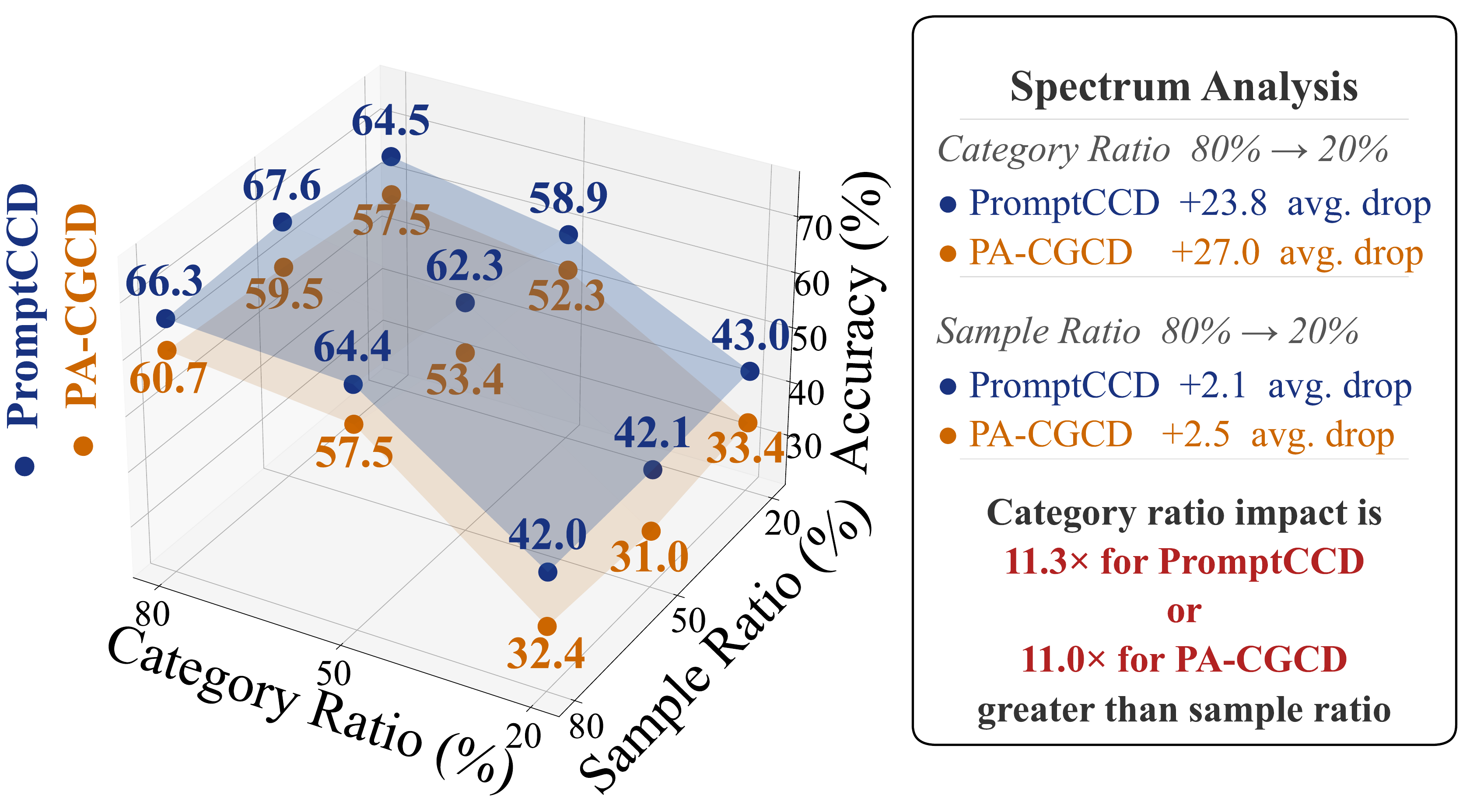}
    \caption{\textbf{Effect of labelled category ratio and sample ratio on novel category accuracy in CCD  task.} Average novel category accuracy across seven datasets under varying labelled category and sample ratios for PromptCCD and PA-CGCD, evaluated following~\cite{cendra2024promptccd} protocol with two discovery stages. Accuracy degrades sharply as the labelled category ratio decreases, while remaining robust to reductions in labelled sample ratio.}
    \label{fig:spectrum_motivation}
\end{figure}

While PromptCCD achieves strong performance, we seek to understand the key factors that influence discovery accuracy and identify opportunities for further improvement. To this end, we conduct a controlled spectrum study across seven CCD datasets, systematically varying the labelled category ratio and sample ratio (Fig.~\ref{fig:spectrum_motivation}). The study reveals a clear finding: \emph{category count}, not sample volume, is the primary bottleneck for category discovery. Reducing the category ratio degrades accuracy far more severely than reducing the sample ratio. This is a significant insight for CCD, because the number of known categories is fixed at initialization while the category space continues to grow over time. Since global class prototypes provide only a single representational anchor per category, the model's ability to discriminate novel classes deteriorates as known categories become insufficient to span the feature space. This finding motivates learning finer-grained representations that provide more discriminative features per category.

This observation naturally raises another question: \emph{despite that the number of known categories is fixed, what representations can be learned to effectively discriminate the growing number of novel categories?} Object parts naturally serve this purpose: they are shared across categories and offer multiple discriminative cues per class, even when the number of known categories is small. However, existing part-level methods in GCD~\cite{wang2024sptnet} operate at rigid pixel-level granularities and lack mechanisms for knowledge retention across stages. To this end, our second design, \textbf{PromptCCD++}, focuses on \textbf{object-part prototypes} via the \textbf{Part-Level Prompt (PLP)} module. PLP decomposes the prompt pool into multiple part-specific prompt pools and uses a context-aware router to dynamically assign part prompts to patch tokens, without requiring manual part annotations during discovery. This enables the model to learn fine-grained, part-level representations that improve the discrimination of novel categories while preventing forgetting of previously learned ones.

We have presented preliminary results of this work in~\cite{cendra2024promptccd}, where we introduced PromptCCD-B, PromptCCD with the GMP module, and PromptCCD-U for the case when the number of categories in the unlabelled data is unknown. This work extends~\cite{cendra2024promptccd} in several aspects. First, we conduct a systematic spectrum study that reveals category count as the primary bottleneck for CCD performance, providing the empirical motivation for finer-grained prompting. Second, we propose PromptCCD++, a part-level prompting framework that decomposes the prompt pool into part-specific prompt pools, and show that it is significantly more resilient to the category-count bottleneck than existing methods. Third, we provide substantially expanded experiments, including comprehensive ablation studies on both GMP and PLP prompting module designs. PromptCCD++ achieves state-of-the-art performance across all CCD benchmarks, demonstrating the effectiveness of our approach.

The remainder of this paper is organized as follows. Section~\ref{sec:related work} reviews related work. Section~\ref{sec:method} details our proposed frameworks: PromptCCD-B, PromptCCD, and PromptCCD++. Section~\ref{sec:experiments} reports experimental results, ablations, and controlled studies. Section~\ref{sec:conclusion} concludes the paper.

\section{Related Work}
\label{sec:related work}

\noindent\textbf{Semi-supervised learning} aims to learn a classifier using both labelled and unlabelled data~\cite{sslbook2006,oliver2018realistic}. Most works assume that the unlabelled data contains instances from the \emph{same} categories as those in the labelled data~\cite{oliver2018realistic}. Pseudo-labeling~\cite{rizve2021defense}, consistency regularization~\cite{sohn2020fixmatch,berthelot2019mixmatch,tarvainen2017mean,laine2016temporal}, and non-parametric classification~\cite{assran2021semi} are among the popular methods. Some recent works do not assume the categories in the unlabelled and labelled set to be the same, such as~\cite{saito2021openmatch,huang2021trash,yu2020multi}, yet their focus is still on improving the performance of the categories 
from the labelled set.

\noindent\textbf{Continual learning} aims to train a model to learn to perform on different tasks while keeping the existing knowledge~\cite{de2021continual}. Catastrophic forgetting~\cite{mccloskey1989catastrophic} refers to the phenomenon whereby training on a new task rapidly degrades performance on previously learned tasks. 
Many attempts have been made to enable models to learn new tasks while retaining knowledge of previous ones~\cite{rebuffi2017icarl,li2017learning,li2019learn,graves2016hybrid,boschini2022class,buzzega2020dark,wang2022learning,wang2022dualprompt}. However, these works all assume that the incoming tasks have all labels provided. In contrast, CCD assumes that the new data is fully unlabelled and can have category overlap with previous tasks. 

\noindent\textbf{Category Discovery} addresses the problem where there are novel categories in the unlabelled data and the goal is to automatically categorize the unlabelled samples, leveraging the labelled samples from the seen categories. Novel Category Discovery (NCD), formalized by DTC~\cite{han2019learning}, assumes no overlap between the unlabelled and labelled data. Several successful NCD methods have emerged, showing promising performance through ranking statistics~\cite{han2020automatically,han2021autonovel,zhao21novel, jia21joint}, data augmentation~\cite{zhong2021openmix}, and specialized objective function~\cite{uno,joseph22spacing}. The problem is later extended to Generalized Category Discovery (GCD)~\cite{vaze2022generalized} by considering that the unlabelled data may contain samples from both known and novel categories. Vaze et al.~\cite{vaze2022generalized}, finetunes a pretrained model using both self-supervised~\cite{chen2020simple} and supervised contrastive losses~\cite{khosla2020supervised} and subsequently obtains the label assignment using a semi-supervised $k$-means algorithm. 
SimGCD~\cite{wen2022simple} introduces a strong parametric baseline based on~\cite{vaze2022generalized} for GCD, obtaining strong performance. 
Recent Subsequent research on GCD propose diverse strategies and explore GCD~\cite{vaze2023no, liu2025debgcd,liu2025hyperbolic,he2025seal,ma2025protogcd} in various contexts, inlcluding methods focus on fine-grained categories~\cite{fei2022xcon}, automatic category estimation~\cite{hao2023cipr,zhao2023learning}, prompt learning~\cite{zhang2023promptcal,wang2024sptnet,yang2025consistent}, leveraging object parts~\cite{Cendra2025PartCo,dai2025adaptive,wang2025learning}, multi-modal learning~\cite{zheng2024textual,wang2025get, ouldnoughi2023clip}, federated environments~\cite{pu2024federated}, long-tailed scenario~\cite{zhao2025lt,hoang2025lt}, and handling domain shifts~\cite{wang2024hilo}.

\noindent\textbf{Continual Category Discovery} extends category discovery to a continual learning setting, where a model is first trained on labelled data and subsequently discovers novel categories from a continuous stream of unlabelled data across multiple stages. Despite its practical importance, CCD remains a relatively under-explored problem. Early efforts such as NCDwF~\cite{joseph2022novel} study NCD under the continual learning setting, where the model first learns from labelled data and subsequently focuses on novel category discovery solely from unlabelled data. NCDwF shows that feature distillation and mutual information-based regularizers are effective for this problem. Concurrent with NCDwF, FRoST~\cite{roy2022class} introduces a replay-based method that stores feature prototypes from labelled data during the discovery phase.
MSc-iNCD~\cite{liu2023large} leverages pretrained self-supervised learning models to address this problem.
Grow \& Merge~\cite{zhang2022grow} studies GCD under the continual learning setting, where the model has access to the labelled data in the initial stage and the unlabelled data in sequences in the subsequent stages.
This method utilizes a growing phase to detect novel categories and a merging phase to distil knowledge from both novel and previously learned categories into a single model.
Other methods addressing the GCD problem under the continual learning setting include PA-CGCD~\cite{kim2023proxy}, which prevents forgetting using a proxy-anchor-based method, and MetaGCD~\cite{wu2023metagcd}, which balances class discovery and prevents forgetting using a meta-learning framework. Happy~\cite{ma2024happy} introduces a debiased learning framework for continual generalized category discovery that addresses the bias toward known categories and improves the discovery of novel categories across stages.
Another method, IGCD~\cite{zhao2023incremental}, studies GCD under the continual learning setting in a slightly different way with an emphasis on the iNaturalist dataset for plant and animal species discovery.
In each stage, IGCD takes a partially labelled set of images as input, rather than fully unlabelled data as in~\cite{zhang2022grow, wu2023metagcd, kim2023proxy}.
In this paper, we consider CCD as the setting studied in~\cite{zhang2022grow, wu2023metagcd, kim2023proxy}. In CCD, the model receives the labelled set at the initial stage and is tasked to discover categories from the continuous stream of unlabelled data in the subsequent stages.
\section{Method}
\label{sec:method}
\begin{figure*}
    \centering
    \includegraphics[width=1.0\linewidth]{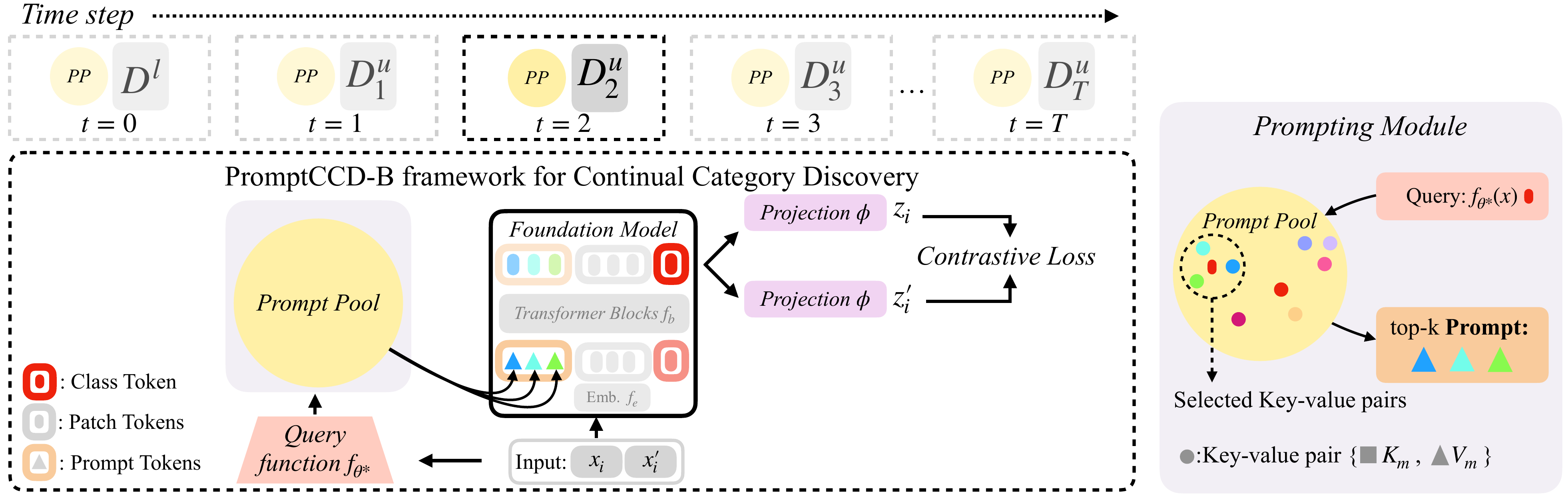}
    \caption{\textbf{Overview of our proposed PromptCCD-B baseline framework}.
    Our baseline CCD framework adopts a prompt-based continual learning technique by utilizing a prompt pool module to adapt the vision foundation model for CCD.}
\label{fig:proCCD_general_design}
\end{figure*}

\noindent\textbf{Problem statement.}
The dataset in CCD contains both labelled data $D^l$ and unlabelled data $D^u$. The labelled data $D^{l} = \{(x_i, y_i)\}^N_{i=1}$ contains tuples of the input ${x_i}\in\mathcal{X}$ and their corresponding labels ${y_i} \in  \mathcal{Y}$ and it is only used in the initial stage for the model to learn useful features for category discovery.
In the following $T$ discovery stages, at each stage, we receive a part of the unlabelled data $D^u_t \subset D^u$ that can be used to train the model. The unlabelled data $D^u_t$ at each stage does not contain the labels, and it contains both known categories from previous stages and also novel categories. The goal of CCD is to train a model $\mathcal{H}_{\theta}: \mathcal{X} \to \mathcal{Z}$ parameterized by $\theta$ that first learns from labelled $D^l$ and then in the following $T$ discovery stages, learns from unlabelled data $D^u_{t}$ such that $\mathcal{H}_{\theta}$ can be used to discover novel classes and assign class labels to all unlabelled instances utilizing representative features without forgetting previous knowledge.

\subsection{PromptCCD-B (\emph{Baseline}): Learning Prompt Pool for CCD}

Prompt learning \cite{wang2022dualprompt,wang2022learning} has been shown effective for supervised continual learning. With properly designed prompts, the necessity of extensive modification for the model when handling the growing data stream can be greatly reduced.
However, these methods cannot be directly applied to the CCD task, as they assume the data stream to be fully annotated, which is not the case for CCD.

To address this gap, we propose a novel baseline prompt learning framework for CCD denoted as PromptCCD-B, taking inspiration from~\cite{wang2022dualprompt,wang2022learning} which learn a pool of prompts to adapt a large-scale pretrained model on ImageNet-21K \cite{ridnik2021imagenet21k} (in a supervised manner) for supervised continual learning. 
This baseline is designed to learn a shared pool of prompts that can effectively adapt the self-supervised foundation model to tackle the CCD challenge. Specifically, the model extracts a feature from a query example using a frozen pretrained model, and the feature will be used to retrieve the \text{top-k} most relevant prompts from the fixed-size $M$ prompts in the shared pool. These prompts are then used to guide the representation learning process by prepending them with the input embeddings, optimised with contrastive learning at each learning stage. 

The overall framework of our baseline is shown in Fig.~\ref{fig:proCCD_general_design}.
Given a model $\mathcal{H}_{\theta}: \{\phi, f_{\theta}\}$, where $\phi$ is a projection head, and $f_{\theta}=\{f_e, f_b\}$ is the transformer-based feature backbone which consists of input embedding layer $f_e$ and self-attention blocks $f_b$. An input image $x \in \mathbb{R}^{H \times W \times 3}$ where $H, W$ represent the height and width of the image, is first split into $L$ tokens (patches) such that $x_q \in {\mathbb{R}}^{L \times (h \times w \times 3)}$ where $h,w$ represent the height and width of the image patches. These patches are then projected by the input embedding layer $x_e = f_{e}(x_q) \in \mathbb{R}^{L \times z}$. A learnable prompt pool with $M$ prompts is denoted as $\mathbb{V} = \{(K_{m}, V_{m})\}_{m=1}^{M}$ where $K_{m} \in \mathbb{R}^{z}$ and $V_{m} \in \mathbb{R}^{L_{pp} \times z}$ are the key-value learnable pairs and $L_{pp}$ is the prompt pool's token length. We define a query function $f_{{\theta}^{*}}$ (\emph{frozen} $f_{{\theta}}$) to map the input image $x$ to the feature space. The query process on the prompt pool operates in a key-value fashion. For a given query $f_{{\theta}^{*}}(x)$, we find the $\text{top-k}$ most similar keys in the prompt pool and retrieve the associated value by:
\begin{equation}
    \mathcal{V}_{\text{top-k}}=\{V_i | K_i \in \mathcal{T}_{\mathbb{V}}^k(f_{{\theta}^{*}}(x))\},
\end{equation}
where $\mathcal{T}_{\mathbb{V}}^k$ is a set of the $\text{top-k}$ similar keys in $\mathbb{V}$. These retrieved prompts are then prepended to the patch embeddings to aid the learning process $x_{total} = [\mathcal{V}_{\text{top-k}}; x_{e}]$. The baseline method is trained with contrastive learning. Let $\{x_i, x'_i\}$ be two randomly augmented views of the same image $x_i$. We obtain their representations as $z_i = \phi(f_{\theta}(x_{i}))$ and $z'_i = \phi(f_{\theta}(x'_{i}))$.
To optimize the prompt pool, we pull the selected keys closer to the corresponding query features by making use of a cosine distance loss:
\begin{equation}
    \mathcal{L}^{\text{cos}}_{i} = \sum_{K_{m}\in \mathcal{T}_{\mathbb{V}}^k(f_{{\theta}^{*}}(x))}\gamma(f_{{\theta}^{*}}(x_i),  K_{m}),
    \label{eq8}
\end{equation}
where $\gamma$ is the cosine distance function. Finally, when the training of stage $t$ is finished, we transfer the current prompt pool $\mathbb{V}$ to the next stage.

\label{sec: optimization}
\noindent\textbf{Model optimization}.
To optimize the model's representation,
we follow the GCD literature to adopt the contrastive loss: 
\begin{equation}
    \mathcal{L}^{\text{rep}}_{i} = - \frac{1}{\mid \mathbb{N}(i) \mid} \sum_{p \in \mathbb{N}(i)}
    \log{\frac{\exp(z_i \cdot z_{p} / \tau)}{\sum_{n} 
     \mathds{1}_{[n \neq i]} \exp(z_i \cdot z_{n} / \tau)} },
     \label{eq5}
\end{equation}
where $\mathds{1}_{[n \neq i]}$ is an indicator function such that it equals $1$ \textit{iff} $n \neq i$, and $\tau$ is the temperature value. If $x_i$ is a labelled image, $\mathbb{N}(i)$ corresponds to images with the same label $y$ in the mini-batch $B$. If $x_i$ is an unlabelled image, $\mathbb{N}(i)$ contains only the index of the other augmented view $x_i'$ of the image, \ie, $z_{p} = z'_i$. For the baseline model optimization, at the initial stage, \ie, $t = 0$, we have our initial labelled set $D^{l}$, and each image may have more than one positive sample; while at the subsequent stages, \ie, $t>0$, we only have access to the unlabelled data $D^{u}_t$, and each image has only one positive sample, \ie, its another augmented view. The prompt parameters are also simultaneously optimized during the model optimization process. 

\noindent \textbf{Limitations of baseline.}
Our baseline can achieve reasonably good performance; However, it has some limitations. 
First, it lacks an explicit mechanism to prevent forgetting. Without label information to guide it, the model may inadvertently bias its representations towards the current unlabelled data during fine-tuning, resulting in representation bias and forgetting. Another limitation arises from the fixed size of the prompt pool in the baseline. The baseline relies on a predefined prompt pool size, which restricts the model's scalability. Consequently, the prompt pool's parameters may hinder the model's ability to discover a growing number of new categories. Lastly, the baseline lacks an efficient mechanism to estimate the number of categories dynamically, which is a crucial challenge for category discovery as per the GCD literature, but remains an open challenge under-explored in CCD.
\begin{figure*}[!ht]
    \centering
    \includegraphics[width=\linewidth]{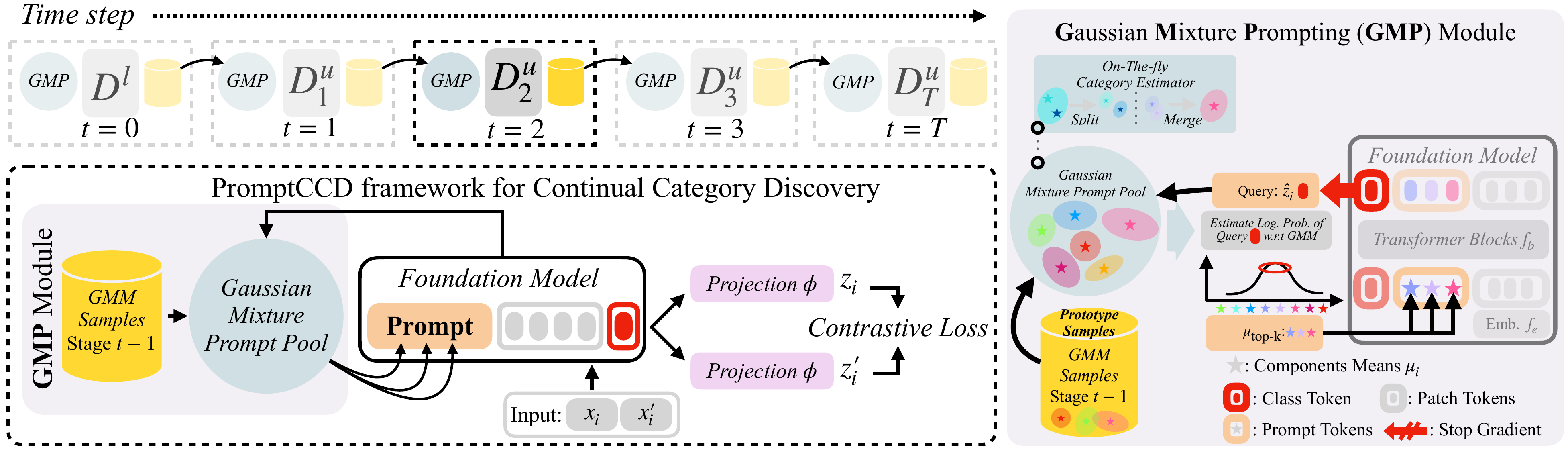}
    \caption{\textbf{Overview of our proposed PromptCCD framework and Gaussian Mixture Prompting (GMP) module}. PromptCCD continually discovers new categories while retaining previously discovered ones by learning a dynamic GMP pool to adapt the vision foundation model for CCD. Specifically, we address CCD by making use of GMP modules to estimate the probability of input $\hat{z_i}$ by calculating the log-likelihood and use the \text{top-k} mean of components $\mu_i$ as prompts to guide the foundation model. Lastly, to retain previously learned prompts, we generate prototype samples from the fitted GMM at time step $t - 1$ and fit the current GMM with these samples at time step $t$.}
\label{fig:proCCD_design}
\end{figure*}

\subsection{PromptCCD: Learning Gaussian Mixture Prompt Pool for CCD}
\label{subsec: promptccd}
To address the aforementioned limitations in our baseline framework PromptCCD-B, here, we propose a novel Gaussian Mixture Prompting (GMP) module, which learns a parameter-efficient Gaussian Mixture Model (GMM) as the prompt pool, leading to a new framework, called PromptCCD~(see Fig.~\ref{fig:proCCD_design}). 

\noindent\textbf{Gaussian mixtures prompting (GMP) module.} 
The GMM is formulated as:
\begin{equation}
    p(z) = \sum^{C}_{c=1}\pi_c\mathcal{N}(z|\mu_c,\Sigma_c) 
    \quad
    \text{s.t.} 
    \quad
    \sum^{C}_{c=1}\pi_c = 1,
\label{eq2}
\end{equation}
where $C$ is the number of Gaussian components, $\pi_c$ is the learnable mixture weight, $\mu_c$ is the mean of each component, and $\Sigma_c$ is the covariance of each component. Given the feature $\hat{z}_i = f_\theta(x_i)$ corresponding to the $\texttt{[CLS]}$ token in the backbone, we calculate the log probability density value of each of the mixture components with the queried feature $\hat{z}_i$ and obtain a set of log-likelihood values for different GMM components. Finally, we find the $\text{top-k}$ components in GMM with the highest log-likelihood values and retrieve the associated components' means:
\begin{equation}
    \mu_{\text{top-k}}=\{\mu_c | c \in \mathcal{T}_{\text{GMM}}^{k}(p(\hat{z}_i))\},
\end{equation}
where $\mathcal{T}^{k}_{\text{GMM}}$ is a set of the $\text{top-k}$ component(s) $c$.
Similar to our baseline framework, PromptCCD-B, a set of embeddings $x_{total} = [\mu_{\text{top-k}}; x_{e}]$ is formed by prepending the selected prompts with the patch embeddings. 
We then apply the same contrastive learning objective as in Eq.~\ref{eq5}, to optimize our PromptCCD~framework. The Gaussian mixture prompt pool serves as the core component that supports continuous category discovery across stages. After the training at stage $t$ is done, we use the fitted GMM to sample a set of samples $\mathcal{Z}^s_t$ with $S$ samples for each component $c$ in the GMM. These samples are used to prevent forgetting previously learned knowledge; we achieve this by using these samples $\mathcal{Z}^s_t$ to fit the GMM of the next stage $t+1$. 
A pseudocode of the training procedure is provided in Sec.~S1 of the supp. material.

Our GMP possesses several unique strengths over the existing prompting techniques for supervised continual learning~\cite{wang2022dualprompt,wang2022learning}. 
First, GMP's prompt serves a dual role, namely (1) as a task prompt to instruct the model and (2) as class prototypes to act as parametric replay sample distribution for discovered classes. The second role, which is unique and important for CCD/GCD, not only allows the model to draw unlimited replay samples to facilitate the representation tuning and class discovery in the next time step but also allows the model to transfer knowledge of previously discovered categories and incorporate this information when making the decision to discover a novel category.
Second, our GMP module enables easy adjustment of parameters and efficient dynamic expansion across stages. This allows our model to enjoy great scalability which is especially important when handling a growing number of categories.

\noindent\textbf{Estimation of unknown number of categories.}
\label{unknown_C}
The GMM-based design of our GMP allows us to equip it with an automatic split-and-merge mechanism, allowing our model to estimate the unknown number of categories in the unlabelled data stream. When the class numbers in the unlabelled data are unknown, one way to approach this problem is to estimate it offline using the non-parametric clustering method introduced in~\cite{vaze2022generalized} at each time step. In CCD, considering the continual learning nature of the problem, it would be more desirable to estimate the class numbers on-the-fly without introducing extra models or an offline process. Inspired by GPC~\cite{zhao2023learning}, which introduces a GMM-based category number estimation method for GCD that automatically splits and merges clusters by assessing their compactness and separability using MCMC, we incorporate this idea into our framework, making use of our GMP module, further enabling the capability of our framework for automatic category number estimation. We denote this extended variant of our framework as PromptCCD-U.

Specifically, considering stage $t = 1$ of CCD, we first extract the features for all the unlabelled samples $D^u_t$ and then use our GMM in our GMP fitted in the previous stage $t = 0$ to generate a set of pseudo features (as a replay for previously learned classes from the labelled data $D^l$). We denote the combined features as $\mathcal{Z}$ and then fit them into the GMM in our GMP. As the class number in $D^u_t$ is unknown, we start the fitting by setting an initial class number of the known class number in $D^l$ and incorporate a \emph{split-and-merge} mechanism as in~\cite{zhao2023learning} to allow for the dynamic adjustment of the GMM. Particularly, for each of the Gaussian components of the GMM, we further decompose it into two sub-components, \ie, $\mu_{c,1}, \mu_{c,2}$ and $\Sigma_{c,1}, \Sigma_{c,2}$. 
We then calculate the Hastings ratio which measures the compactness and separability of the clusters during the fitting iteration. 
The Hastings ratio for splitting a cluster is defined as:
\begin{align}
    H_{s}=\frac{\Gamma(N_{c,1})h(\mathcal{Z}_{c,1}) \Gamma(N_{c,2})h(\mathcal{Z}_{c,2})}{\Gamma(N_c)h(\mathcal{Z}_c)},
\end{align}
where $\Gamma$ is the factorial function, $h$ is the marginal likelihood function of the observed data $\mathcal{Z}$, $\mathcal{Z}_{c,1}$ denotes the data points assigned to the subcluster $\{c,1\}$, and $N_{c,1}$ is the number of data points in the subcluster $\{c,1\}$. Note that $H_{s}$ is in the range of  $(0, +\infty)$, thus we will use $p_s=min(1, H_s)$ as a valid probability for performing the splitting operation.
When the fitting of the GMM is converged, the number of the resulting GMM components is then the class number of all classes seen so far. The number of new classes in  $D^u_t$ can be obtained by simply subtracting the previously learned class number. 

\subsection{PromptCCD++: Learning Part-level Prompt Pools for CCD}
\label{subsec: plp}

\begin{figure*}
\centering
\includegraphics[width=0.85\linewidth]{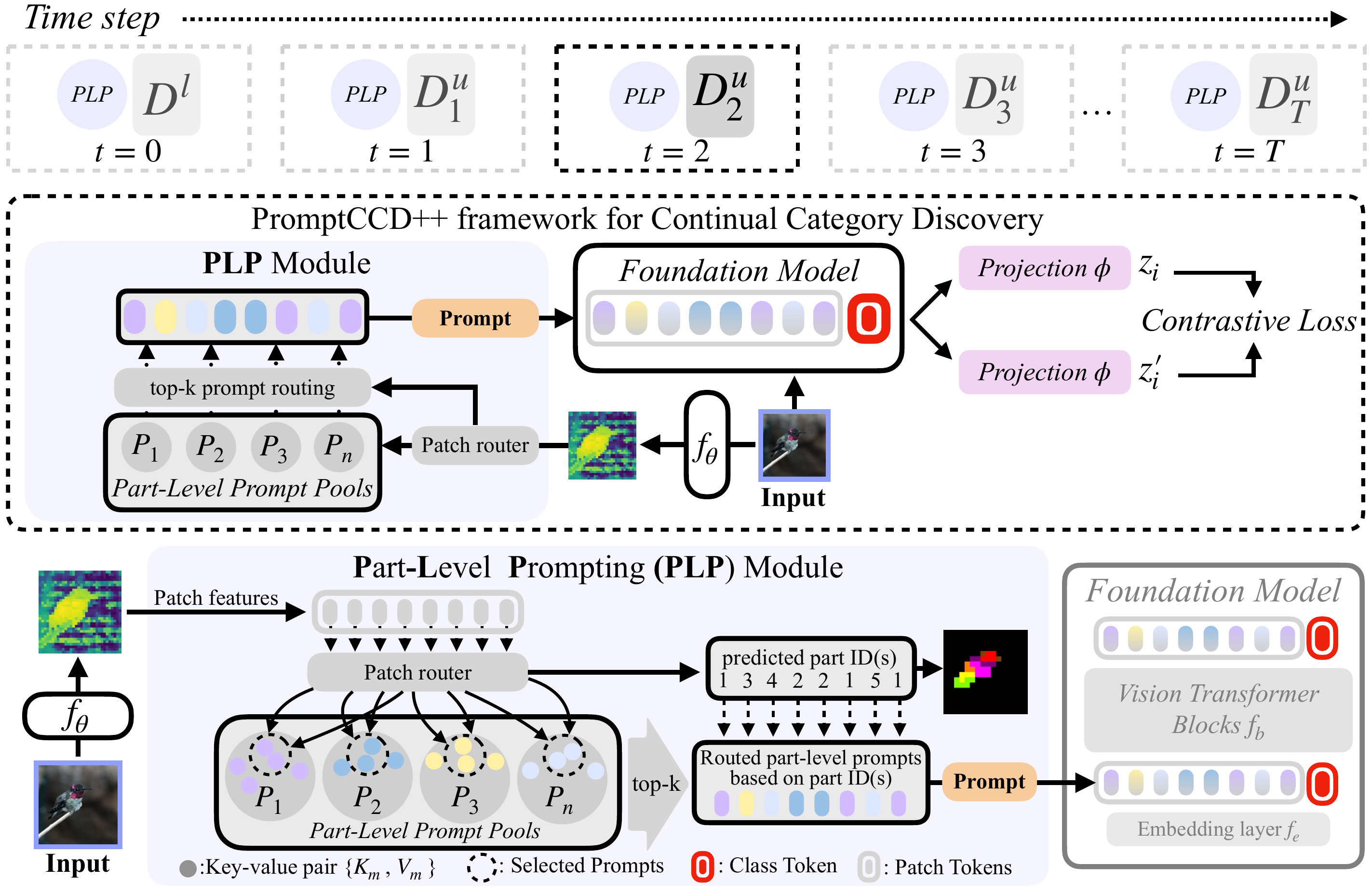}
    \caption{\textbf{Overview of our proposed PromptCCD++ framework and Part-Level Prompting (PLP) module.} The prompt memory is decomposed into $P$ part-specific prompt pools. During discovery stages ($t > 0$), a context-aware router assigns each spatial patch token to its corresponding part pool without requiring manual part annotations. Top-$k$ prompts are retrieved from the assigned pool and applied via multiplicative modulation. Additionally, a frozen teacher model provides feature distillation to mitigate forgetting while the part-specific prompt pools continue to adapt to newly discovered categories.}
\label{fig:proCCD_plp_design}
\end{figure*}

While PromptCCD's GMP module provides an effective global prompt mechanism for CCD, its prompts are derived from class-level prototypes that summarize an entire object with a single mixture component. As the semantic space grows increasingly crowded across stages, such global representations risk conflating categories that share broad visual appearance but differ in fine-grained structural details~(\eg, bird species distinguished primarily by \emph{beak shape} or \emph{wing pattern}). To address this, we propose PromptCCD++, with the overall framework illustrated in Fig.~\ref{fig:proCCD_plp_design}. PromptCCD++ introduces a Part-Level Prompting~(PLP) module that architecturally decomposes the prompt mechanism from a single global pool into multiple, part-specific prompt pools that operate at the spatial patch level.

\noindent\textbf{Part-level prompt pool decomposition.} Rather than maintaining a single shared prompt pool $\mathbb{V}$ as in PromptCCD-B, PLP decomposes the prompt memory into $P$ part-specific prompt pools, one for each semantic part (including a background component). Formally, we define
\begin{equation}
    \mathbb{P} = \bigl\{(K_p, V_p)\bigr\}_{p=1}^{P},
\end{equation}
where $K_p \in \mathbb{R}^{M \times z}$ and $V_p \in \mathbb{R}^{M \times L_{pp} \times z}$ denote the learnable key-value prompt pairs for the $p$-th part pool, $M$ is the number of prompts per pool, $L_{pp}$ is the prompt token length, and $z$ is the embedding dimension. Each pool specialises in capturing the visual characteristics of a particular object part, enabling the framework to learn fine-grained structural dependencies that a single global pool cannot resolve.

\noindent\textbf{Context-aware patch routing.} To assign each spatial patch token to the appropriate part pool \emph{without} requiring part annotations during discovery phase, PLP employs a lightweight routing module. Given the patch tokens $x_e \in \mathbb{R}^{L \times z}$ from $f_e$, we first enrich them with a spatial saliency signal by projecting the CLS-to-patch attention maps $\alpha \in \mathbb{R}^{L \times H}$ from the backbone's last self-attention layer (where $H$ is the number of attention heads) and fusing them with the patch tokens:
\begin{equation}
    x_{\text{aug}} = x_e + \text{AttnProj}(\alpha)
    \in \mathbb{R}^{L \times z},
    \label{eq: attn_fusion}
\end{equation}
where $\text{AttnProj}(\cdot)$ is a lightweight projection consisting of layer normalisation followed by a linear layer and GELU activation, mapping from $H$ dimensions to $z$. This attention fusion provides each patch with information about the backbone's spatial focus, improving the router's ability to distinguish semantically salient regions from background. We then contextualise the augmented tokens via a Transformer encoder:
\begin{equation}
    \tilde{x} = \text{TransEnc}\bigl(\text{LN}(x_{\text{aug}})\bigr)
    \in \mathbb{R}^{L \times z},
\end{equation}
where $\text{LN}(\cdot)$ denotes layer normalisation and $\text{TransEnc}(\cdot)$ is a two-layer Transformer encoder. We then compute routing logits by combining two complementary signals: (i)~attention scores between the contextualised tokens and a set of learnable part queries
$Q = \{q_p\}_{p=1}^{P} \in \mathbb{R}^{P \times z}$, and
(ii)~a projection head $g_\psi$:
\begin{equation}
    r = g_\psi(\tilde{x})
      + \frac{\tilde{x}\, Q^\top}{\sqrt{z}}
    \in \mathbb{R}^{L \times P},
    \label{eq:routing}
\end{equation}
where $g_\psi$ is a two-layer MLP. The routing probabilities are obtained as $\sigma(r) = \text{softmax}(r)$, and each patch is assigned to the part with the highest probability: $p^{*}_l = \arg\max_p\; \sigma(r)_{l,p}$.

\noindent\textbf{Part-specific prompt retrieval and modulation.}
Once patches are assigned to parts, each patch queries its designated part pool in a key-value fashion. For a patch $l$ assigned to part $p^{*}$, we compute the cosine similarity between its contextualised representation $\tilde{x}_l$ and the keys $K_{p^{*}}$, and retrieve the top-$k$ prompts:
\begin{equation}
    \mathcal{V}^{(l)}_{\text{top-k}}
    = \bigl\{V_{p^{*},m}
      \;\big|\; K_{p^{*},m}
      \in \mathcal{T}^{k}_{K_{p^{*}}}(\tilde{x}_l)\bigr\},
\end{equation}
where $\mathcal{T}^{k}_{K_{p^{*}}}$ denotes the set of $k$ keys in pool $p^{*}$ with the highest cosine similarity. The retrieved prompt values are averaged across both the top-$k$ and prompt-length dimensions to produce a compact bias vector $b_l \in \mathbb{R}^{z}$ for each patch. The bias is applied via multiplicative modulation of the original patch embeddings:
\begin{equation} 
    x^{plp}_{e,l} = x_{e,l} \odot (1 + b_l),
    \label{eq:modulation}
\end{equation}
where $\odot$ denotes element-wise multiplication. This multiplicative gating allows each part-specific prompt to selectively amplify or suppress feature dimensions relevant to its assigned part, without altering the overall representational structure.

\noindent\textbf{Router learning and stage-wise freezing.} A key design consideration for the continual setting is how the routing module is trained and maintained across stages. During the initial supervised stage ($t{=}0$), we construct part labels $y^{\text{part}}_l$ from part-level segmentation masks derived from the labelled set $D^l$ (details of the label construction are provided in Sec.S3 of the supp. material). Briefly, we first extract patch-level features from a frozen DINOv2 backbone, use PCA to derive objectness and foreground-aware projections, and then apply k-means clustering to obtain part-level correspondence labels for each dataset. We supervise the router with a standard cross-entropy loss over the patch-level part assignments:
\begin{equation}
    \mathcal{L}^{\text{route}} = \frac{1}{L}\sum_{l=1}^{L}
    \text{CE}\bigl(r_l,\; y^{\text{part}}_l\bigr),
    \label{eq:route_loss}
\end{equation}
where $y^{\text{part}}_l$ is the part annotation for patch $l$. This supervision enables the router to learn a meaningful decomposition of object structure during the initial stage. Crucially, for all subsequent discovery stages ($t > 0$), the router parameters $\psi$ are \emph{frozen}. This design choice serves two purposes: (i)~it provides stable, consistent part assignments across stages, preventing the routing from drifting as new categories are introduced; and (ii)~it leverages the observation that part decompositions learned on base categories transfer well to novel categories, as object parts (\eg, \emph{head, wing, tail}) serve as universal structural primitives shared across related categories~\cite{wang2024sptnet}. While the router remains fixed, the part-specific prompt pools $\mathbb{P}$ continue to be optimised at every stage, allowing the prompt memory to evolve and accommodate newly discovered categories.

\noindent\textbf{Knowledge retention via feature distillation.}
To further mitigate catastrophic forgetting in the prompt pools
and backbone, we employ a dual distillation mechanism for stages
$t > 0$. We maintain a frozen teacher model $f^{(0)}_\theta$
from the initial stage and enforce consistency through:
(i)~a feature-level distillation loss between the student and
teacher CLS representations:
\begin{equation}
    \mathcal{L}^{\text{distill}} =
    \bigl\| \bar{z}_i - \bar{z}^{\text{teacher}}_i \bigr\|^{2}_{2},
    \label{eq:distill}
\end{equation}
where $\bar{z}_i$ and $\bar{z}^{\text{teacher}}_i$ are the normalised projected features from the student and teacher, respectively; and (ii)~a routing-distribution alignment loss that prevents the part assignment statistics from diverging:
\begin{equation}
    \mathcal{L}^{\text{anchor}} =
    D_{\text{KL}}\bigl(
      \bar{\sigma}(r)
      \;\|\;
      \bar{\sigma}(r^{\text{teacher}})
    \bigr),
    \label{eq:anchor_kl}
\end{equation}
where $\bar{\sigma}(r) = \frac{1}{L}\sum_{l=1}^{L}\sigma(r_l)$ is the image-level average routing distribution. This term ensures that the overall part composition perceived by the student remains anchored to the teacher's decomposition, even as the prompt pools adapt to new categories.

\noindent\textbf{Overall objective.} The full training objective for PromptCCD++ at stage $t$ combines the contrastive representation loss from Eq.~\ref{eq5}, the prompt key alignment loss from Eq.~\ref{eq8}, the routing supervision (when $t{=}0$), and the distillation terms (when $t{>}0$):
\begin{equation}
    \mathcal{L} =
    \mathcal{L}^{\text{rep}}
    + \mathcal{L}^{\text{cos}}
    + \lambda_{\text{route}}\,\mathcal{L}^{\text{route}}
    + \lambda_{\text{dist}}\bigl(
        \mathcal{L}^{\text{distill}}
        + \mathcal{L}^{\text{anchor}}
      \bigr),
    \label{eq:total_loss}
\end{equation}
where $\lambda_{\text{route}} > 0$ only at $t{=}0$, and $\lambda_{\text{dist}} > 0$ only at $t{>}0$. After training at each stage, the updated prompt pools $\mathbb{P}$ are carried forward to the next stage, maintaining a persistent yet evolving part-level memory. A pseudocode of the training procedure is provided in Sec.~S2 of the supp. material.

\section{Experiments}
\label{sec:experiments}
In this section, we describe our experimental setups in Sec.~\ref{sec:setup}. Next, we present our main experimental results in Sec.~\ref{sec:main_results}. Finally, in Sec.~\ref{sec:ablation}, we analyze the effectiveness of our PromptCCD and PromptCCD++ framework's components and design choices respectively.

\subsection{Experimental Setups}
\label{sec:setup}
\noindent\textbf{Datasets.}
We conduct our experiments on various benchmark datasets, namely CIFAR100 (C100)~\cite{krizhevsky2009learning}, ImageNet-100 (IN-100)~\cite{russakovsky2015imagenet}, TinyImageNet (Tiny)~\cite{le2015tiny}, Caltech-UCSD Birds-200-2011 (CUB)~\cite{welinder2010caltech}, FGVC-Aircraft~\cite{maji13fine-grained}, Stanford-Cars (SCars)~\cite{krause20133d}, and Caltech-101~(C-101)~\cite{fei2006one}. 
Statistics of the benchmark datasets are shown in Tab.~\ref{tab: data stat}.
CCD task consists of several stages. We set the number of stages to $4$ with data splits presented in Tab.~\ref{tab:data_dist} following \cite{zhang2022grow}. 
\definecolor{SoftP}{rgb}{0.945, 0.933, 0.949}
\begin{table*}[ht]
\centering
\caption{\textbf{Statistics of the CCD benchmark datasets} following the splits in Tab.~\ref{tab:data_dist}.
\label{tab: data stat}}
    \resizebox{1.0\textwidth}{!}{
        \begin{tabular}{lC{1cm}C{1cm}C{1cm}C{1cm}C{1cm}C{1cm}C{1cm}C{1cm}C{1cm}C{1cm}C{1cm}C{1cm}C{1cm}C{1cm}}
        \toprule
        \multicolumn{1}{c}{\multirow{2}{*}{Stages}} 
                                & \multicolumn{2}{c}{C100~\cite{krizhevsky2009learning}} & \multicolumn{2}{c}{IN-100~\cite{russakovsky2015imagenet}} 
                                & \multicolumn{2}{c}{Tiny~\cite{le2015tiny}} & \multicolumn{2}{c}{C-101~\cite{fei2006one}} 
                                & \multicolumn{2}{c}{Aircraft~\cite{maji13fine-grained}} & \multicolumn{2}{c}{SCars~\cite{krause20133d}} 
                                & \multicolumn{2}{c}{CUB~\cite{welinder2010caltech}}  \\  
        
                                 & $C$  & \# & $C$  & \# & $C$  & \# & $C$  & \# 
                                 & $C$  & \# & $C$  & \# & $C$ & \# \\ \midrule
        
        Stage 0 $(D^{l})$        & $70$  & $\textcolor{gray}{30.45K}$ & $70$  & $\textcolor{gray}{77.46K}$ & $140$ & $\textcolor{gray}{60.90K}$ & $71$  & $\textcolor{gray}{4.70K}$
                                & $70$  & $\textcolor{gray}{1.98K}$ & $130$  & $\textcolor{gray}{4.62K}$ & $140$ & $\textcolor{gray}{3.65K}$ \\ 
        
        Stage 1 $(D^{u}_{1})$    & $80$  & $\textcolor{gray}{5.95K}$ & $80$ & $\textcolor{gray}{15.14K}$ & $160$  & $\textcolor{gray}{11.90K}$ & $81$  & $\textcolor{gray}{0.73K}$
                                & $80$  & $\textcolor{gray}{0.37K}$ & $152$ & $\textcolor{gray}{0.98K}$ & $160$ & $\textcolor{gray}{0.71K}$ \\ 
        
        Stage 2 $(D^{u}_{2})$    & $90$  & $\textcolor{gray}{6.55K}$ & $90$  & $\textcolor{gray}{16.66K}$ & $180$ & $\textcolor{gray}{13.10K}$ & $91$ & $\textcolor{gray}{0.65K}$
                                & $90$ & $\textcolor{gray}{0.43K}$ & $174$  & $\textcolor{gray}{1.13K}$ & $180$  & $\textcolor{gray}{0.79K}$ \\ 
        
        Stage 3 $(D^{u}_{3})$    & $100$ & $\textcolor{gray}{7.05K}$ & $100$ & $\textcolor{gray}{17.94K}$ & $200$ & $\textcolor{gray}{14.10K}$ & $101$  & $\textcolor{gray}{1.12K}$
                                & $100$  & $\textcolor{gray}{0.55K}$ & $196$ & $\textcolor{gray}{1.38K}$ & $200$  & $\textcolor{gray}{0.85K}$ \\ \bottomrule
        \end{tabular}
    }
\end{table*}
\definecolor{Gray}{gray}{0.9}
\definecolor{PaleBlue}{rgb}{0.7529, 0.9137, 0.9372}
\definecolor{BeauBlue}{rgb}{0.7686, 0.8470, 0.9529}
\definecolor{Mauve}{rgb}{0.8 , 0.7098, 0.9843}
\definecolor{PaleViolet}{rgb}{0.8156, 0.6431, 1.0}
\definecolor{Salmon}{rgb}{1.0, 0.8980, 0.6920}
\definecolor{Pink}{rgb}{1.0, 0.6902, 0.7908}
\definecolor{Mint}{rgb}{0.6902, 1.0, 0.7451}
\definecolor{SoftP}{rgb}{0.945, 0.933, 0.949}

\newcommand{\tikzxmark}{
\tikz[scale=0.23] {
    \draw[line width=0.7,line cap=round] (0,0) to [bend left=6] (1,1);
    \draw[line width=0.7,line cap=round] (0.2,0.95) to [bend right=3] (0.8,0.05);
}}

\begin{table}[!htb] 
    \centering
    \caption{\textbf{CCD data splits}.}
      \centering
      \resizebox{0.9\columnwidth}{!}{
      \centering
          \begin{tabular}{lcccc} 
            \toprule
            \multicolumn{1}{c}{\large Class splits} & \large $D^{l}$ & \large $D^{u}_{1}$ & \large $D^{u}_{2}$ & \large $D^{u}_{3}$ \\ [0.5ex] 
           \midrule
            $\{y_i \mid y_i \leq 0.7 \ast |\mathcal{Y}| \}$ 
            & $87\%$  & $7\%$  & $3\%$  & $3\%$ \\ [0.5ex] \midrule
            $\{y_i \mid 0.7 \ast |\mathcal{Y}| < y_i \leq 0.8 \ast |\mathcal{Y}| \}$ 
            & $\textcolor{gray}{0\%}$ & $70\%$  & $20\%$  & $10\%$  \\ [0.5ex] \midrule
            $\{y_i \mid 0.8 \ast |\mathcal{Y}| < y_i \leq 0.9 \ast |\mathcal{Y}| \}$ 
            & $\textcolor{gray}{0\%}$ & $\textcolor{gray}{0\%}$ & $90\%$  & $10\%$  \\ [0.5ex] \midrule
            $\{y_i \mid 0.9 \ast |\mathcal{Y}| < y_i \leq |\mathcal{Y}| \}$ 
            & $\textcolor{gray}{0\%}$ & $\textcolor{gray}{0\%}$ & $\textcolor{gray}{0\%}$ & $100\%$  \\ [0.5ex] \bottomrule
          \end{tabular}
    }
    \label{tab:data_dist}
\end{table}

\noindent\textbf{Implementation details.}
We use a $\text{ViT-B}$ backbone \cite{dosovitskiy2020image} pretrained with DINO variants 
\cite{caron2021emerging, oquab2023dinov2} for our experiments. Note that \cite{wang2022learning, wang2022dualprompt} utilized a pretrained model with supervision, which is suitable for the standard supervised continual learning task. However, it is not well-suited to use such pretrained models for the CCD task due to label information leakage. During training, only the final block of the vision transformer is finetuned for $200$ epochs with a batch size of $128$, using SGD optimizer and cosine decay learning rate scheduler with an initial learning rate of $0.1$ and minimum learning rate of $0.0001$, and weight decay of $0.00005$. 
For the GMP module, we optimize the GMM every $30$ epochs and start the prompt learning when the epoch is greater than $30$. We set $\text{top-k}$ to be $5$, and the number of GMM samples to $100$.
For the PLP module, we set the number of part-level prompt pools $P$ according to the number of parts per dataset (\eg, $8$ for CUB including background, for more details see supp. material), with a pool size of $20$, prompt pool length of $10$, and $\text{top-k}$ of $2$. The routing loss weight is set to $\lambda_{\text{route}}{=}0.1$ during the initial stage, and the distillation weight is set to $\lambda_{\text{dist}}{=}0.2$ during discovery stages. The router parameters are frozen after the initial stage.
All input images are resized to $224 \times 224$ and augmented to match the DINO pretrained model settings. For our method, we finetune the last transformer block of the model $f_b$ and the projection head $\phi$ (Sec.~S10 of the supp. material for details)
using the loss introduced in~Sec.~\ref{sec:method}. For other compared methods, we carefully choose the right hyper-parameters following their original papers.
Finally, we dynamically estimate the class number using the method described in~Sec.~\ref{unknown_C}, following a  procedure similar to~\cite{zhao2023learning}. We build our framework with PyTorch on a single NVIDIA RTX $3090$ GPU.

\noindent\textbf{Evaluation metric.}
The model is finetuned at each stage. At test time, the classification token $\texttt{[CLS]}$ features are used for clustering. For the clustering algorithm and label assignment, we use semi-supervised $k$-means (SS-$k$-means) \cite{vaze2022generalized} on the unlabelled sets $D^u_t$ and measure the accuracy given the ground truth $y_i$ and the clustering prediction $\hat{y}_i$ such that:

\begin{equation}
    ACC = \max_{g \in \mathcal{G}(\mathcal{Y}_U)} \frac{1}{|D^u_t|} \sum^{|D^u_t|}_{i=1}\mathds{1}\{y_i = g(\hat{y}_i) \},
\label{eq9}
\end{equation}
where $\mathcal{G(\mathcal{Y}_U)}$ represents a set of all permutations of class labels in the unlabelled set $D^u_t$. For the evaluation across stages in CCD, based on the standard clustering accuracy \textit{ACC} for GCD, we introduce a new metric, called \textit{continual ACC} (\textit{cACC}), for the continual setting considering the sequential data stream. 
Commonly, in GCD, the ACC values are evaluated for \textit{`All'}, \textit{`Old'}, and \textit{`New'} splits of the dataset.  
In CCD, for one time step $t$,
\textit{`All'} indicates the overall accuracy on the entire set $D^u_t$. \textit{`Old'} and \textit{`New'} indicate the accuracy from instances of unlabelled data from $D^{uo}_{t}$ and $D^{un}_t$ respectively. 
The evaluation protocol of cACC is summarized in Alg.~\ref{alg:ccd_eval}. 
Instead of relying solely on the labelled data $D^{l}$ to guide the SS-$k$-means clustering algorithm, \textit{cACC} incorporates labelled data from $\{D^{l}, D^{u^*}_{1}, \dots, D^{u^*}_{t-1}\}$, where $D^{u^*}_{i}$ represents data with assigned labels from previously unlabelled data  $D^{u}_{i}$. 
High-quality label assignments facilitate the subsequent discovery, while low-quality label assignments accumulate errors for the subsequent category discovery.
\begin{figure}[!htb] 
    \begin{minipage}{\columnwidth}
    \centering
        \begin{algorithm}[H]
        \footnotesize
        \hsize=\textwidth 
        \begin{algorithmic}[1]
        \Statex \textbf{Input:} Models $\{f_{\theta}^{t} \mid t=1, \dots, T\}$ and datasets $\{D^l , D^u\}$.
        \Statex \textbf{Output:} \textit{cACC} value. 
        \Statex \textbf{Require:} \Call{SS-$k$-means}{Model, Labelled set, Unlabelled set}. 
        \Statex \textbf{Require:} Initialize set $\mathbb{A}^L \gets D^{l}$.
        \For{$t \in \{1, \cdots, T\}$}
            \State $ACC_t$ $,$ $D^{u^*}_{t}$ $\gets$  \Call{SS-$k$-means}{
                $f_{\theta}^{t}$,
                $\mathbb{A}^L$, 
                $D^u_t$}
            \State $\mathbb{A}^L$ $\leftarrow$ $\mathbb{A}^L$ $\cup$ $D^{u^*}_{t}$  \Comment{append $D^{u^*}_{t}$ ($w/$ assigned labels) to $\mathbb{A}^L$}
        \EndFor
        \State $ACCs \gets \{ACC_{t} \mid t=1, \dots, T\}$
        \State $cACC \gets \Call{Average}{ACCs}$
        \State \Return $cACC$
        \end{algorithmic}
        \caption{\textit{Continual ACC} (\textit{cACC}) evaluation metric}
        \label{alg:ccd_eval}
        \end{algorithm}
    \end{minipage}
\end{figure}

\noindent\textbf{Comparison with other methods.}
We compare our method with the other representative CCD methods: 1) Grow \& Merge (G\&M)~\cite{zhang2022grow}; 2) MetaGCD~\cite{wu2023metagcd}; 3) PA-CGCD~\cite{kim2023proxy};4) Happy~\cite{ma2024happy}, and re-implement GCD methods for CCD task, including 5) ORCA~\cite{cao22orca}; 6) GCD~\cite{vaze2022generalized}; 7) SimGCD \cite{wen2022simple}. As G\&M's encoder is based on ResNet18  network~\cite{he2016deep}, we re-implement their dynamic branch mechanism with the ViT backbone and observe improved performance for their method compared to their original results (see Sec.~S7 of the supp. material). 
We also re-implement GCD and SimGCD for CCD settings by incorporating a replay-based method. At each stage, the model saves samples for discovered classes and mixes them with incoming streamed images. Lastly, we adopt L2P's~\cite{wang2022learning} and DualPrompt's~\cite{wang2022dualprompt} prompt pool modules, following their original prompt pool hyperparameter choices, and integrate them with PromptCCD-B~framework as our baselines.

\subsection{Main Results}
\label{sec:main_results}
\begin{figure*}[!ht]
    \centering
    \includegraphics[width=\linewidth]{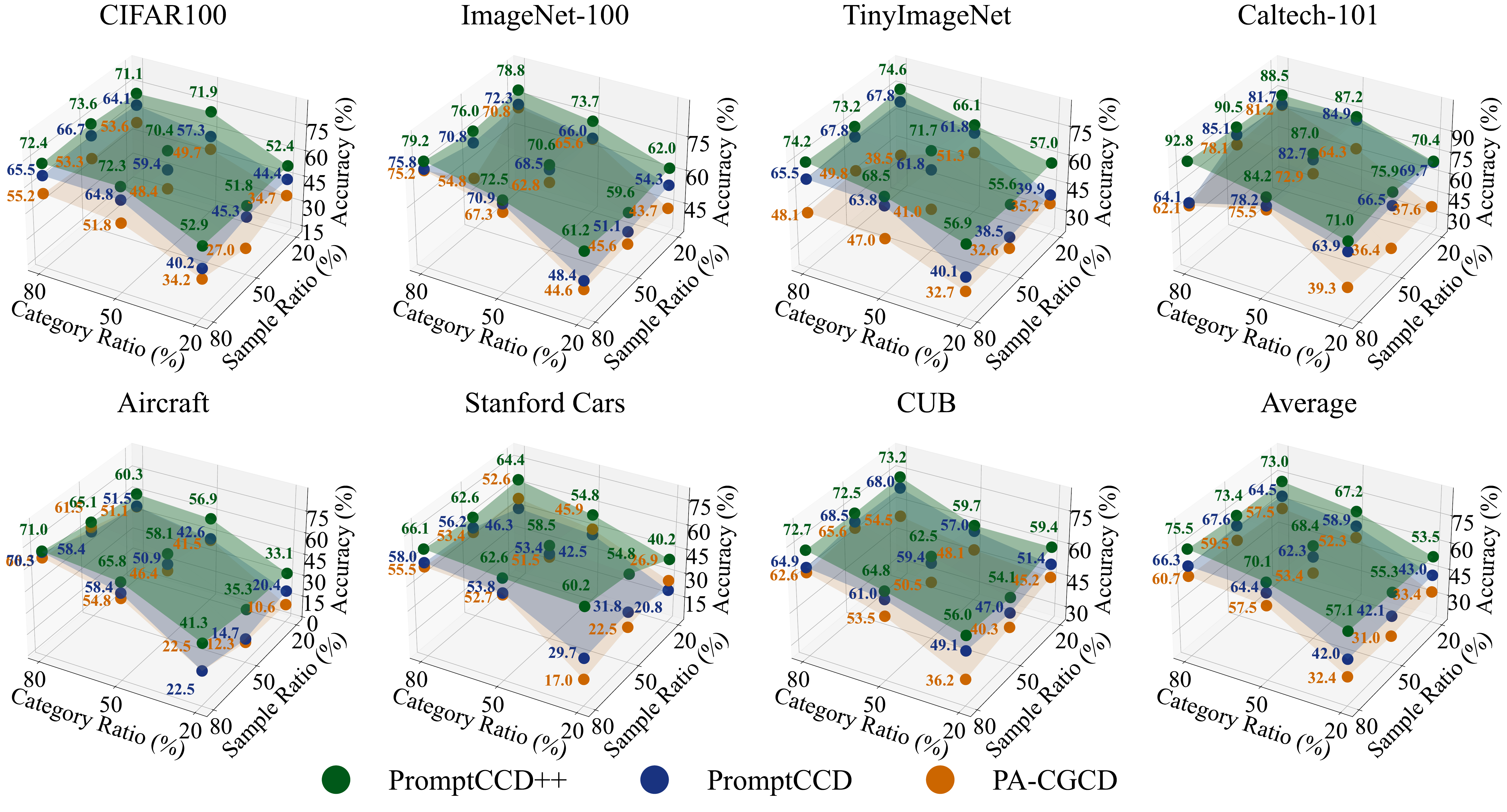}
    \caption{\textbf{Spectrum analysis of all accuracy across varying labelled category and sample ratios on seven benchmarks}. PromptCCD++ $w/$PLP, PromptCCD $w/$GMP, and PA-CGCD are compared across all combinations of labelled category ratio (20\%, 50\%, 80\%) and sample ratio (20\%, 50\%, 80\%) using DINOv2 backbone. PromptCCD++ $w/$PLP consistently achieves superior \textit{`New'} accuracy compared to PromptCCD w/ GMP and PA-CGCD across all seven benchmarks. Notably, even as the category ratio decreases (\ie, fewer known categories are available), PLP maintains robust \textit{`New'} accuracy by leveraging part-level prompts to better discriminate novel categories.}
    \label{fig:spectrum_full}
\end{figure*}

We evaluate our method in two scenarios: when the class number $C$ is known (Tab.~\ref{tab:main_result_avg_baselines}, \ref{tab:main_result_avg_ssb}, and \ref{tab:main_result_avg}) in each unlabelled set at different stages, and when $C$ is unknown (Tab.~\ref{tab:main_result_unknown_avg}). We further conduct a controlled spectrum study (Fig.~\ref{fig:spectrum_full}) to analyse the sensitivity of each method to varying labelled category and sample ratios. We report the \textit{cACC} by averaging the results across all stages. We also provide the breakdown results for each stage in Sec.~S4 of the supp. material. 

\noindent\textbf{Variants of PromptCCD.} 
In our comparison with the baseline PromptCCD-B, as shown in Tab.~\ref{tab:main_result_avg_baselines}, both PromptCCD and PromptCCD++ demonstrate superior performance for CCD. Specifically, both variants outperform the baselines across all datasets in \textit{`All'} accuracy, while the baselines experience performance degradation in later stages (see Sec.~S4). We attribute this decline to the baselines' non-scalable prompt pool parameters, which restrict their ability to \textit{instruct} the model as the category count grows. In contrast, PromptCCD leverages
Gaussian mixture models to construct a flexible pool of prompts and ensures knowledge retention by sampling learned mixture components for fitting subsequent GMMs. PromptCCD++ further improves upon this
by decomposing the prompt memory into part-specific pools, enabling finer-grained discrimination that is particularly beneficial as the semantic space becomes increasingly crowded across stages.
\definecolor{Gray}{gray}{0.9}
\definecolor{PaleBlue}{rgb}{0.7529, 0.9137, 0.9372}
\definecolor{BeauBlue}{rgb}{0.7686, 0.8470, 0.9529}
\definecolor{Mauve}{rgb}{0.8 , 0.7098, 0.9843}
\definecolor{PaleViolet}{rgb}{0.8156, 0.6431, 1.0}
\definecolor{Salmon}{rgb}{1.0, 0.8980, 0.6920}
\definecolor{Pink}{rgb}{1.0, 0.6902, 0.7908}
\definecolor{Mint}{rgb}{0.6902, 1.0, 0.7451}
\definecolor{Cyan}{rgb}{0.906, 0.969, 0.965}
\definecolor{SoftP}{rgb}{0.945, 0.933, 0.949}

\begin{table*}[!ht]
    \caption{\textbf{Experiment on different PromptCCD's frameworks with multiple seeds}. The \textit{cACC} results of our method with different prompt pool designs for CCD on generic and fine-grained benchmark datasets where $C$ is \textit{known} in each unlabelled set. The experiments are conducted five times with different random seeds.}
    \centering
    \resizebox{\textwidth}{!}{%
    \centering
        \begin{tabular}{lcc ccc c ccc}
        \toprule
        \multicolumn{3}{c}{}
        & \multicolumn{3}{c}{CIFAR100}
        && \multicolumn{3}{c}{ImageNet-100} 
        \\ 
        \cmidrule[0.1pt](r{0.80em}){4-6}
        \cmidrule[0.1pt](r{0.80em}){8-10}%

        \multicolumn{1}{c}{Method} &
        \multicolumn{1}{c}{Prompt Pool} &
        \multicolumn{1}{c}{Backbone} &
        \multicolumn{1}{c}{\textit{All} \cellcolor{blue!7!white} } & 
        \multicolumn{1}{c}{\textit{Old}} &
        \multicolumn{1}{c}{\textit{New}} &&
        \multicolumn{1}{c}{\textit{All} \cellcolor{blue!7!white} } & 
        \multicolumn{1}{c}{\textit{Old}} &
        \multicolumn{1}{c}{\textit{New}} 
        \\ \midrule
        \rowcolor{gray!5!white}
        PromptCCD-B~(Ours) & L2P \cite{wang2022learning}
        & DINO
        & $51.59~\smallstd{6.3}$ \cellcolor{blue!7!white}  & $67.27~\smallstd{8.7}$  & $46.14~\smallstd{6.1}$ &
        & $66.14~\smallstd{2.3}$ \cellcolor{blue!7!white}  & $81.05~\smallstd{1.5}$  & $61.36~\smallstd{3.2}$ \\
        \rowcolor{gray!5!white}
        PromptCCD-B~(Ours) & DP \cite{wang2022dualprompt}
        & DINO
        & $59.60~\smallstd{1.2}$ \cellcolor{blue!7!white}  & $78.93~\smallstd{1.3}$  & $54.14~\smallstd{1.6}$ &
        & $70.64~\smallstd{1.3}$ \cellcolor{blue!7!white}  & $83.46~\smallstd{0.4}$  & $67.24~\smallstd{1.8}$ \\
        \rowcolor{gray!5!white}
        PromptCCD~(Ours) & GMP (Ours)
        & DINO
        & $63.97~\smallstd{1.4}$ \cellcolor{blue!7!white}  & $76.67~\smallstd{2.6}$ & $60.01~\smallstd{1.7}$ &
        & $75.38~\smallstd{0.7}$ \cellcolor{blue!7!white}  & $81.16~\smallstd{0.7}$ & $73.71~\smallstd{0.8}$ \\
        \midrule
        \rowcolor{gray!5!white}
        PromptCCD~(Ours) & GMP (Ours)
        & DINOv2
        & $68.75~\smallstd{2.3}$ \cellcolor{blue!7!white}  & $78.35~\smallstd{0.9}$ & $64.63~\smallstd{3.0}$ &&  $76.44~\smallstd{0.9}$ \cellcolor{blue!7!white}  & $81.65~\smallstd{0.9}$ & $74.46~\smallstd{1.3}$\\
        \rowcolor{gray!5!white}
        PromptCCD++~(Ours) & PLP (Ours)
        & DINOv2
        & $\textbf{76.87}~\smallstd{1.0}$ \cellcolor{blue!7!white} & $87.83~\smallstd{2.6}$ & $76.02~\smallstd{1.1}$ &
        & $\textbf{82.29}~\smallstd{0.7}$ \cellcolor{blue!7!white}  & $84.30~\smallstd{0.8}$ & $81.08~\smallstd{0.8}$\\
        \toprule
        \multicolumn{3}{c}{}
        & \multicolumn{3}{c}{TinyImageNet} 
        && \multicolumn{3}{c}{CUB}
        \\ 
        \cmidrule[0.1pt](r{0.80em}){4-6}
        \cmidrule[0.1pt](r{0.80em}){8-10}%
        \multicolumn{1}{c}{Method} &
        \multicolumn{1}{c}{Prompt Pool} &
        \multicolumn{1}{c}{Backbone} &
        \multicolumn{1}{c}{\textit{All} \cellcolor{blue!7!white} } & 
        \multicolumn{1}{c}{\textit{Old}} &
        \multicolumn{1}{c}{\textit{New}} &&
        \multicolumn{1}{c}{\textit{All} \cellcolor{blue!7!white} } & 
        \multicolumn{1}{c}{\textit{Old}} &
        \multicolumn{1}{c}{\textit{New}} 
        \\ \midrule
        \rowcolor{gray!5!white}
        PromptCCD-B~(Ours) & L2P \cite{wang2022learning}
        & DINO
        & $56.66~\smallstd{0.4}$ \cellcolor{blue!7!white}  & $66.05~\smallstd{0.8}$ & $53.69~\smallstd{0.4}$ &
        & $51.31~\smallstd{1.0}$ \cellcolor{blue!7!white}  & $72.43~\smallstd{1.0}$ & $44.27~\smallstd{1.4}$ \\
        \rowcolor{gray!5!white}
        PromptCCD-B~(Ours) & DP \cite{wang2022dualprompt}
        & DINO
        & $58.61~\smallstd{1.5}$ \cellcolor{blue!7!white}  & $66.61~\smallstd{0.6}$ & $55.84~\smallstd{1.7}$ &
        & $56.30~\smallstd{1.1}$ \cellcolor{blue!7!white} & $78.64~\smallstd{1.7}$ & $48.91~\smallstd{1.1}$ \\
        \rowcolor{gray!5!white}
        PromptCCD~(Ours) & GMP (Ours)
        & DINO
        & $61.15~\smallstd{1.0}$ \cellcolor{blue!7!white}  & $66.29~\smallstd{2.0}$ & $58.83~\smallstd{1.0}$  &
        & $56.65~\smallstd{1.0}$ \cellcolor{blue!7!white}  & $79.88~\smallstd{2.5}$ & $48.96~\smallstd{0.8}$ \\
        \midrule
        \rowcolor{gray!5!white}
        PromptCCD~(Ours) & GMP (Ours)
        & DINOv2
        & $67.19~\smallstd{0.9}$ \cellcolor{blue!7!white}  & $75.16~\smallstd{0.7}$ & $63.89~\smallstd{1.1}$ &
        & $67.98~\smallstd{2.0}$ \cellcolor{blue!7!white}  & $85.33~\smallstd{2.1}$ & $61.52~\smallstd{2.9}$\\
        \rowcolor{gray!5!white}
        PromptCCD++~(Ours) & PLP (Ours)
        & DINOv2
        & $\textbf{72.05}~\smallstd{0.9}$ \cellcolor{blue!7!white}  & $79.61~\smallstd{0.5}$ & $69.37~\smallstd{1.3}$ &
        &  $\textbf{75.52}~\smallstd{0.4}$  \cellcolor{blue!7!white}  & $88.45~\smallstd{1.6}$ & $70.48~\smallstd{0.8}$ \\
        \bottomrule
        \end{tabular}
    }
    \label{tab:main_result_avg_baselines}
\end{table*}
\begin{table*}[!htb]
  \centering
  \caption{\textbf{Comparison of CCD methods on the fine-grained benchmark}. Results are reported in \textit{cACC} across the `\textit{All}', `\textit{Old}' and `\textit{New}' categories.}
  \label{tab:main_result_avg_ssb}
  
  \resizebox{\textwidth}{!}{
    \begin{tabular}{
      l c ccc c ccc c ccc c ccc
    }
      \toprule
      &
      & \multicolumn{3}{c}{Aircraft} 
      & & \multicolumn{3}{c}{Stanford Cars} 
      & & \multicolumn{3}{c}{CUB} 
      & & \multicolumn{3}{c}{Average fine-grained}\\
      \cmidrule(lr){3-5} \cmidrule(lr){7-9} \cmidrule(lr){11-13} \cmidrule(lr){15-17}
      Method & Backbone & \cellcolor{blue!7!white}\textit{All} & \textit{Old} & \textit{New} &&  \cellcolor{blue!7!white}\textit{All} & \textit{Old} & \textit{New} &&  \cellcolor{blue!7!white}\textit{All} & \textit{Old} & \textit{New} &&  \cellcolor{blue!7!white}\textit{All} & \textit{Old} & \textit{New}\\
      \midrule
        \rowcolor{gray!5!white}
        ORCA~\cite{cao22orca}& DINO & \cellcolor{blue!7!white}30.77 & 25.71 & 32.44 && \cellcolor{blue!7!white}20.79 & 33.40 & 17.60 && \cellcolor{blue!7!white}41.73 & 66.19 & 34.14 && \cellcolor{blue!7!white}31.10 &41.77  &28.06  \\
        \rowcolor{gray!5!white}
        GCD~\cite{vaze2022generalized}& DINO & \cellcolor{blue!7!white}47.37 & 61.43 & 42.53 && \cellcolor{blue!7!white}39.21 & 58.29 & 33.45 && \cellcolor{blue!7!white}54.98 & 75.47 & 48.15 && \cellcolor{blue!7!white}47.19 &65.06 &41.38  \\
        \rowcolor{gray!5!white}
        SimGCD~\cite{wen2022simple}& DINO & \cellcolor{blue!7!white}29.03 & 35.72 & 25.61 && \cellcolor{blue!7!white}21.01 & 40.93 & 16.48 && \cellcolor{blue!7!white}39.89 & 59.25 & 33.75 && \cellcolor{blue!7!white}29.98 &45.30 &25.28  \\
        \rowcolor{gray!5!white}
        GCD $w/$replay& DINO & \cellcolor{blue!7!white}45.63 & 62.38 & 39.89 && \cellcolor{blue!7!white}39.87 & 58.18 & 33.89 && \cellcolor{blue!7!white}54.66 & 74.64 & 47.81 && \cellcolor{blue!7!white}46.72 &65.07 &40.53  \\
        \rowcolor{gray!5!white}
        SimGCD $w/$replay& DINO & \cellcolor{blue!7!white}37.44 & 61.43 & 28.96 && \cellcolor{blue!7!white}22.76 & 49.04 & 16.65 && \cellcolor{blue!7!white}42.08 & 72.65 & 31.92 && \cellcolor{blue!7!white}34.09 &61.04  &25.84  \\
        \midrule
        \rowcolor{gray!5!white}
        Grow \& Merge~\cite{zhang2022grow} & DINO & \cellcolor{blue!7!white}31.06 & 33.33 & 30.78 && \cellcolor{blue!7!white}21.90 & 35.29 & 18.17 && \cellcolor{blue!7!white}38.87 & 65.00 & 30.29 && \cellcolor{blue!7!white}30.61 &44.54  &26.41  \\
        \rowcolor{gray!5!white}
        MetaGCD~\cite{wu2023metagcd}& DINO & \cellcolor{blue!7!white}44.63 & 59.05 & 39.39 && \cellcolor{blue!7!white}35.98 & 56.97 & 29.96 && \cellcolor{blue!7!white}44.59 & 74.40 & 35.40 && \cellcolor{blue!7!white}41.73 & 63.47  & 34.92  \\
        \rowcolor{gray!5!white}
        PA-CGCD~\cite{kim2023proxy}& DINO & \cellcolor{blue!7!white}48.24 & 73.09 & 40.60 && \cellcolor{blue!7!white}43.88 & 80.43 & 33.54 && \cellcolor{blue!7!white}52.48 & 77.26 & 44.74 && \cellcolor{blue!7!white}48.20 &76.93  &39.63  \\
        \rowcolor{gray!5!white}
        Happy~\cite{ma2024happy} & DINO & \cellcolor{blue!7!white}45.77 & 59.05 & 40.00 && \cellcolor{blue!7!white}45.36 & 68.96 & 35.49 && \cellcolor{blue!7!white}57.74 & 75.42 & 51.37 && \cellcolor{blue!7!white}49.62 & 67.81 & 42.29 \\
        \rowcolor{blue!3!white}
        PromptCCD (Ours)& DINO & \cellcolor{blue!7!white}52.64 & 60.48 & 50.23 && \cellcolor{blue!7!white}44.07 & 66.36 & 36.83 && \cellcolor{blue!7!white}55.45 & 75.48 & 48.56 && \cellcolor{blue!7!white}50.72 &67.44  &45.21  \\
        \midrule
        \rowcolor{gray!5!white}
        GCD~\cite{vaze2022generalized}& DINOv2  & \cellcolor{blue!7!white}57.87 & 63.80 & 55.39 && \cellcolor{blue!7!white}58.52 & 71.65 & 53.80 && \cellcolor{blue!7!white}66.70 & 83.33 & 60.81 && \cellcolor{blue!7!white}61.03 &72.93 &56.67  \\
        \rowcolor{gray!5!white}
        MetaGCD~\cite{wu2023metagcd}& DINOv2 & \cellcolor{blue!7!white}54.90 & 64.29 & 52.08 && \cellcolor{blue!7!white}57.16 & 71.87 & 52.01 && \cellcolor{blue!7!white}62.19 & 82.50 & 55.13 && \cellcolor{blue!7!white}58.08 &67.44 &45.21  \\
        \rowcolor{gray!5!white}
        PA-CGCD\cite{kim2023proxy}& DINOv2 & \cellcolor{blue!7!white}58.15 & 77.62 & 51.08 && \cellcolor{blue!7!white}64.91 & 89.64 & 57.84 && \cellcolor{blue!7!white}66.88 & 92.62 & 58.48 && \cellcolor{blue!7!white}63.31 &86.63 & 55.80  \\
        \rowcolor{gray!5!white}
        Happy~\cite{ma2024happy} & DINOv2 & \cellcolor{blue!7!white}64.53 & 73.81 & 59.08 && \cellcolor{blue!7!white}66.49 & 76.88 & 61.32 && \cellcolor{blue!7!white}72.13 & 86.07 & 66.74  && \cellcolor{blue!7!white}67.72 & 78.92 & 62.38 \\
        \rowcolor{blue!3!white}
        PromptCCD (Ours)& DINOv2 & \cellcolor{blue!7!white}62.71 & 68.33 & 60.82 && \cellcolor{blue!7!white}65.08 & 76.60 & 60.75 && \cellcolor{blue!7!white}67.81 & 81.55 & 62.81 && \cellcolor{blue!7!white}65.20 &75.50 & 61.46 \\
        \rowcolor{blue!3!white}
        PromptCCD++ (Ours)& DINOv2 & \cellcolor{blue!7!white}\textbf{71.40} & 78.57 & 68.92 && \cellcolor{blue!7!white}\textbf{70.35} & 85.92 & 65.49 && \cellcolor{blue!7!white}\textbf{76.02} & 86.31 & 71.73 && \cellcolor{blue!7!white}\textbf{72.59} & 83.60 & 68.71 \\
      \bottomrule
    \end{tabular}
  }
\end{table*}
\begin{table*}[!htb]
  \centering
  \caption{\textbf{Comparison of CCD methods on the generic benchmark}. Results are reported in \textit{cACC} across the `\textit{All}', `\textit{Old}' and `\textit{New}' categories.}
  \label{tab:main_result_avg}
  
  \resizebox{\textwidth}{!}{
    \begin{tabular}{
      l c ccc c ccc c ccc c ccc c ccc
    }
      \toprule
      &
      & \multicolumn{3}{c}{CIFAR100} 
      & & \multicolumn{3}{c}{ImageNet-100} 
      & & \multicolumn{3}{c}{TinyImageNet} 
      & & \multicolumn{3}{c}{Caltech-101}
      & & \multicolumn{3}{c}{Average generic}\\
      \cmidrule(lr){3-5} \cmidrule(lr){7-9} \cmidrule(lr){11-13} \cmidrule(lr){15-17} \cmidrule(lr){19-21}
      Method & Backbone & \cellcolor{blue!7!white}\textit{All} & \textit{Old} & \textit{New} &&  \cellcolor{blue!7!white}\textit{All} & \textit{Old} & \textit{New} &&  \cellcolor{blue!7!white}\textit{All} & \textit{Old} & \textit{New} &&  \cellcolor{blue!7!white}\textit{All} & \textit{Old} & \textit{New} &&  \cellcolor{blue!7!white}\textit{All} & \textit{Old} & \textit{New}\\
      \midrule
       \rowcolor{gray!5!white}
        ORCA~\cite{cao22orca} & DINO & \cellcolor{blue!7!white}60.91 & 66.61 & 58.33 && \cellcolor{blue!7!white}40.29 & 45.85 & 35.40 && \cellcolor{blue!7!white}54.71 & 63.13 & 51.93 && \cellcolor{blue!7!white}76.77 & 82.80 & 73.20  && \cellcolor{blue!7!white}58.17 &64.60 &54.72 \\
        \rowcolor{gray!5!white}
        GCD~\cite{vaze2022generalized}& DINO & \cellcolor{blue!7!white}58.18 & 72.27 & 52.83 && \cellcolor{blue!7!white}69.41 & 81.56 & 65.65 && \cellcolor{blue!7!white}55.20 & 65.87 & 51.61 && \cellcolor{blue!7!white}78.27 & 86.60 & 72.92  && \cellcolor{blue!7!white}65.27 &76.58 &60.75 \\
        \rowcolor{gray!5!white}
        SimGCD~\cite{wen2022simple}& DINO & \cellcolor{blue!7!white}25.56 & 38.76 & 20.43 && \cellcolor{blue!7!white}31.38 & 40.47 & 27.44 && \cellcolor{blue!7!white}33.40 & 29.11 & 34.74 && \cellcolor{blue!7!white}33.65 & 37.53 & 31.62  && \cellcolor{blue!7!white}31.00 &36.47 &28.56 \\
        \rowcolor{gray!5!white}
        GCD $w/$replay & DINO & \cellcolor{blue!7!white}49.93 & 73.15 & 41.47 && \cellcolor{blue!7!white}72.04 & 83.75 & 69.01 && \cellcolor{blue!7!white}56.33 & 67.54 & 52.60 && \cellcolor{blue!7!white}76.51 & 86.14 & 72.48  && \cellcolor{blue!7!white}63.70 &77.65 &58.90 \\
        \rowcolor{gray!5!white}
        SimGCD $w/$replay& DINO & \cellcolor{blue!7!white}40.13 & 66.72 & 30.91 && \cellcolor{blue!7!white}47.53 & 67.86 & 39.18 && \cellcolor{blue!7!white}37.45 & 58.15 & 30.36 && \cellcolor{blue!7!white}49.38 & 52.72 & 47.99  && \cellcolor{blue!7!white}43.62 &61.36  &37.11 \\
        \midrule
        \rowcolor{gray!5!white}
        Grow \& Merge~\cite{zhang2022grow} & DINO & \cellcolor{blue!7!white}57.43 & 63.68 & 55.31 && \cellcolor{blue!7!white}67.84 & 75.10 & 66.60 && \cellcolor{blue!7!white}52.14 & 59.68 & 49.96 && \cellcolor{blue!7!white}75.75 & 83.66 & 71.59  && \cellcolor{blue!7!white}63.29 &70.53  &60.87 \\
        \rowcolor{gray!5!white}
        MetaGCD~\cite{wu2023metagcd}& DINO & \cellcolor{blue!7!white}55.49 & 69.38 & 48.98 && \cellcolor{blue!7!white}66.41 & 80.54 & 60.65 && \cellcolor{blue!7!white}55.26 & 66.12 & 50.79 && \cellcolor{blue!7!white}80.75 & 89.02 & 75.86  && \cellcolor{blue!7!white}64.48 &76.27 &59.07 \\
        \rowcolor{gray!5!white}
        PA-CGCD~\cite{kim2023proxy}& DINO & \cellcolor{blue!7!white}58.25 & 87.11 & 49.04 && \cellcolor{blue!7!white}64.79 & 91.15 & 57.83 && \cellcolor{blue!7!white}51.13 & 74.95 & 43.52 && \cellcolor{blue!7!white}77.96 & 94.75 & 69.66  && \cellcolor{blue!7!white}63.03 &86.99  &55.01 \\
        \rowcolor{gray!5!white}
        Happy~\cite{ma2024happy} & DINO & 64.27\cellcolor{blue!7!white} & 77.54 & 59.05 && 74.78\cellcolor{blue!7!white} & 82.54 & 72.38 && 58.63\cellcolor{blue!7!white} & 65.71 & 55.99 && 
        80.94\cellcolor{blue!7!white} & 89.98 &  77.03 && 69.66\cellcolor{blue!7!white} & 78.94 & 66.11\\
        \rowcolor{blue!3!white}
        PromptCCD (Ours)& DINO & \cellcolor{blue!7!white}64.17 & 75.57 & 60.34 && \cellcolor{blue!7!white}76.16 & 81.76 & 74.35 && \cellcolor{blue!7!white}61.84 & 66.54 & 60.26 && \cellcolor{blue!7!white}82.44 & 89.08 & 79.72  && \cellcolor{blue!7!white}71.15 &78.24  &68.67 \\
        \midrule
        \rowcolor{gray!5!white}
        GCD~\cite{vaze2022generalized} & DINOv2 & \cellcolor{blue!7!white}65.35 & 77.06 & 60.46 && \cellcolor{blue!7!white}71.58 & 83.02 & 68.05 && \cellcolor{blue!7!white}59.05 & 77.44 & 53.41 && \cellcolor{blue!7!white}83.00 & 88.65 & 79.80  && \cellcolor{blue!7!white}69.75 &81.54  &65.43 \\
        \rowcolor{gray!5!white}
        MetaGCD~\cite{wu2023metagcd}& DINOv2 & \cellcolor{blue!7!white}52.10 & 79.64 & 43.13 && \cellcolor{blue!7!white}70.20 & 82.62 & 64.66 && \cellcolor{blue!7!white}56.15 & 74.69 & 49.37 && \cellcolor{blue!7!white}83.05 & 88.08 & 80.89  && \cellcolor{blue!7!white}65.38 &81.26  &59.51 \\
        \rowcolor{gray!5!white}
        PA-CGCD~\cite{kim2023proxy} & DINOv2 & \cellcolor{blue!7!white}54.36 & 79.19 & 45.65 && \cellcolor{blue!7!white}74.82 & 88.20 & 72.02 && \cellcolor{blue!7!white}52.10 & 68.07 & 46.32 && \cellcolor{blue!7!white}83.06 & 94.07 & 77.55  && \cellcolor{blue!7!white}66.09 &82.38  &60.39 \\
        \rowcolor{gray!5!white}
        Happy~\cite{ma2024happy} & DINOv2 & \cellcolor{blue!7!white}73.13 & 85.35 & 69.53 && \cellcolor{blue!7!white}80.36 & 87.36 & 77.80 && \cellcolor{blue!7!white}67.28 & 78.80 & 63.47 && \cellcolor{blue!7!white}88.27 & 90.98 & 87.69 && \cellcolor{blue!7!white}77.26 & 85.62 & 74.62\\
        \rowcolor{blue!3!white}
        PromptCCD (Ours)& DINOv2 & \cellcolor{blue!7!white}69.73 & 78.01 & 66.16 && \cellcolor{blue!7!white}76.28 & 82.61 & 74.53 && \cellcolor{blue!7!white}68.20 & 75.56 & 65.23 && \cellcolor{blue!7!white}83.86 & 87.93 & 81.42  && \cellcolor{blue!7!white}74.52 &81.03  &71.84 \\
        \rowcolor{blue!3!white}
        PromptCCD++ (Ours) & DINOv2 & \cellcolor{blue!7!white}\textbf{77.68} & 83.18 & 75.89 && \cellcolor{blue!7!white}\textbf{83.39} & 85.06 & 82.17 && \cellcolor{blue!7!white}\textbf{73.02} & 80.15 & 70.54  && \cellcolor{blue!7!white}\textbf{91.85} & 93.86 & 91.72 && \cellcolor{blue!7!white}\textbf{81.49} & 85.56 & 80.08  \\
      \bottomrule
    \end{tabular}
  }
\end{table*}

\noindent\textbf{Comparison with known class numbers.} The CCD benchmark results for generic and fine-grained datasets are presented in Tab.~\ref{tab:main_result_avg_ssb} and \ref{tab:main_result_avg}.
PromptCCD achieves competitive performance, obtaining results comparable to Happy~\cite{ma2024happy} across most benchmarks. Building on this foundation, PromptCCD++ achieves state-of-the-art performance, outperforming all other approaches across both generic and fine-grained datasets in terms of overall accuracy \textit{`All'} while maintaining a strong balance between \textit{`Old'} and \textit{`New'} accuracy. As our framework builds upon GCD~\cite{vaze2022generalized}, these results demonstrate that integrating part-level prompting enables effective adaptation to the CCD
setting, highlighting the robustness and versatility of our frameworks across benchmarks.

\label{sec:spectrum}
\noindent\textbf{Controlled spectrum study.} As shown in Fig.~\ref{fig:spectrum_motivation}, the labelled category count is the primary bottleneck for continual category discovery. To assess whether PromptCCD++ $w/$~PLP alleviates this limitation, we repeat the same spectrum analysis, now including PLP alongside GMP and PA-CGCD~\cite{kim2023proxy}. Following the same two-stage protocol with DINOv2 backbone, we vary both the labelled category ratio and sample ratio across all seven datasets. Fig.~\ref{fig:spectrum_full} shows the results. Two trends are evident. First, all three methods are relatively insensitive to sample ratio: reducing it from 80\% to 20\% costs only $\sim$2-3~pp on average. Second, reducing the category ratio from 80\% to 20\% is far more damaging, but PLP suffers notably less: it loses $\sim$18.7~pp, versus $\sim$23.8~pp for GMP and $\sim$27.0~pp for PA-CGCD. At the hardest setting (20\% category ratio), PLP reaches 55.3\% average accuracy, surpassing GMP (42.4\%) by +12.9~pp and PA-CGCD (32.2\%) by +23.1~pp. The gap narrows at higher category ratios, confirming that PLP's advantage is largest precisely where it is most needed. These results provide direct empirical support for the architectural motivation behind PromptCCD++. When the category ratio is low, global class prototypes lack sufficient semantic anchors to discriminate novel classes effectively, leading to steep accuracy degradation. Object-part prototypes, by contrast, decompose the representation into finer-grained structural components that can be shared across categories. Even with a small pool of known categories, these part-level features provide transferable visual primitives, such as object parts, shapes, and local structures, that generalize to novel classes, yielding a flatter and more robust accuracy surface along the category-ratio axis. This validates our hypothesis that \emph{fine-grained, part-level representations are more resilient to the category-count bottleneck} inherent to CCD.

\noindent\textbf{Performance with on-the-fly category number estimation.}
Tab.~\ref{tab:main_result_unknown_avg} reports results in the more realistic setting where the number of categories $C$ is unknown, comparing five representative methods~\cite{vaze2022generalized, zhang2022grow, kim2023proxy, wu2023metagcd} and PromptCCD-U. Our method consistently outperforms all compared methods by a large margin, demonstrating the effectiveness of our approach when the category count is unknown.
\definecolor{Gray}{gray}{0.9}
\definecolor{PaleBlue}{rgb}{0.7529, 0.9137, 0.9372}
\definecolor{BeauBlue}{rgb}{0.7686, 0.8470, 0.9529}
\definecolor{Mauve}{rgb}{0.8 , 0.7098, 0.9843}
\definecolor{PaleViolet}{rgb}{0.8156, 0.6431, 1.0}
\definecolor{Salmon}{rgb}{1.0, 0.8980, 0.6920}
\definecolor{Pink}{rgb}{1.0, 0.6902, 0.7908}
\definecolor{Mint}{rgb}{0.6902, 1.0, 0.7451}
\definecolor{Cyan}{rgb}{0.906, 0.969, 0.965}
\definecolor{SoftP}{rgb}{0.945, 0.933, 0.949}

\begin{table*}[htb]
    \caption{\textbf{CCD benchmark when $C$ is \textit{unknown}.} Comparison with other methods for CCD leveraging the pretrained DINO model when the class number $C$ in each unlabelled set is \textit{unknown}.}
    \resizebox{\textwidth}{!}{%
    \centering
        \begin{tabular}{lc ccc ccc ccc ccc}
        \toprule
        \multicolumn{1}{c}{}
        & \multicolumn{1}{c}{Est. method}
        & \multicolumn{3}{c}{CIFAR100}
        & \multicolumn{3}{c}{ImageNet-100} 
        & \multicolumn{3}{c}{TinyImageNet} 
        & \multicolumn{3}{c}{CUB} 
        \\ 
        \cmidrule[0.1pt](r{0.80em}){3-5}
        \cmidrule[0.1pt](r{0.80em}){6-8}
        \cmidrule[0.1pt](r{0.80em}){9-11}
        \cmidrule[0.1pt](r{0.80em}){12-14}

        \multicolumn{2}{c}{Category discovery at stage $\dashrightarrow$} &
        \multicolumn{1}{c}{1} & 
        \multicolumn{1}{c}{2} &
        \multicolumn{1}{c}{3} &
        \multicolumn{1}{c}{1} & 
        \multicolumn{1}{c}{2} &
        \multicolumn{1}{c}{3} &
        \multicolumn{1}{c}{1} & 
        \multicolumn{1}{c}{2} &
        \multicolumn{1}{c}{3} &
        \multicolumn{1}{c}{1} & 
        \multicolumn{1}{c}{2} &
        \multicolumn{1}{c}{3} \\ \midrule
        Estimated category $C$ & GPC
        & $85$ & $100$ & $115$
        & $83$ & $98$ & $113$
        & $155$ & $170$ & $185$
        & $161$ & $180$ & $198$  \\
        Ground truth category $C$  & --
        & $\underline{80}$  & $\underline{90}$  & $\underline{100}$ 
        & $\underline{80}$  & $\underline{90}$ & $\underline{100}$ 
        & $\underline{160}$  & $\underline{180}$  & $\underline{200}$
        & $\underline{160}$  & $\underline{180}$  & $\underline{200}$  \\
        \midrule
        
        \multicolumn{1}{c}{Methods} &
        \multicolumn{1}{c}{} &
        \multicolumn{1}{c}{\textit{All}  \cellcolor{blue!7!white} } & 
        \multicolumn{1}{c}{\textit{Old}} &
        \multicolumn{1}{c}{\textit{New}} &
        \multicolumn{1}{c}{\textit{All}  \cellcolor{blue!7!white} } & 
        \multicolumn{1}{c}{\textit{Old}} &
        \multicolumn{1}{c}{\textit{New}} &
        \multicolumn{1}{c}{\textit{All}  \cellcolor{blue!7!white} } & 
        \multicolumn{1}{c}{\textit{Old}} &
        \multicolumn{1}{c}{\textit{New}} &
        \multicolumn{1}{c}{\textit{All}  \cellcolor{blue!7!white} } & 
        \multicolumn{1}{c}{\textit{Old}} &
        \multicolumn{1}{c}{\textit{New}} 
        \\ \midrule
        \rowcolor{gray!5!white}
        GCD \cite{vaze2022generalized} & GPC
        & $53.78$  \cellcolor{blue!7!white}  & $74.05$  & $46.37$
        & $68.55$  \cellcolor{blue!7!white}  & $82.05$  & $63.96$ 
        & $55.28$  \cellcolor{blue!7!white}  & $65.04$  & $52.15$  
        & $50.69$  \cellcolor{blue!7!white}  & $72.43$  & $43.16$   \\
        \rowcolor{gray!5!white}
        Grow \& Merge \cite{zhang2022grow} & GPC
        & $53.33$  \cellcolor{blue!7!white}  & $66.64$  & $49.61$
        & $66.40$  \cellcolor{blue!7!white}  & $74.52$  & $64.01$ 
        & $52.40$  \cellcolor{blue!7!white}  & $57.87$  & $51.00$  
        & $38.12$  \cellcolor{blue!7!white}  & $62.21$  & $30.00$   \\
        \rowcolor{gray!5!white}
        MetaGCD \cite{wu2023metagcd} & GPC
        & $47.55$  \cellcolor{blue!7!white}  & $70.79$  & $38.57$
        & $63.48$  \cellcolor{blue!7!white}  & $80.82$  & $56.28$ 
        & $56.21$  \cellcolor{blue!7!white}  & $68.33$  & $50.99$  
        & $44.30$  \cellcolor{blue!7!white}  & $70.69$  & $35.83$   \\
        \rowcolor{gray!5!white}
        PA-CGCD \cite{kim2023proxy} & GPC
        & $55.66$  \cellcolor{blue!7!white}  & $90.21$  & $44.99$
        & $66.74$  \cellcolor{blue!7!white}  & $91.28$  & $58.97$ 
        & $50.55$  \cellcolor{blue!7!white}  & $72.44$  & $43.38$  
        & $52.27$  \cellcolor{blue!7!white}  & $76.38$  & $44.24$   \\
        \rowcolor{blue!3!white}
        PromptCCD-U (Ours) & GPC
        & $\textbf{59.12}$  \cellcolor{blue!7!white}  & $77.62$  & $53.70$
        & $\textbf{70.12}$  \cellcolor{blue!7!white}  & $81.84$  & $66.12$
        & $\textbf{57.76}$  \cellcolor{blue!7!white}  & $64.57$  & $55.37$
        & $\textbf{55.20}$  \cellcolor{blue!7!white}  & $73.19$  & $48.82$  \\ 
        \bottomrule
        \end{tabular}
    }
    \label{tab:main_result_unknown_avg}
\end{table*}

\subsection{Model Component Analysis}
\label{sec:ablation}

\noindent \textbf{The choices of numbers for \text{top-k} and GMM samples in GMP.}
To investigate the effectiveness of our GMP module, we analyzed each component in our prompt module and present the results in Tab.~\ref{tab:ablation_result}. The results show a clear advantage of adopting the GMP into our framework. The number of \text{top-k} prompts and the number of GMM samples are identified as important factors. The optimal configuration is \text{top-$5$} for prompt selection, and $100$ samples for sampling, which appears to be a good trade-off. 
\definecolor{Gray}{gray}{0.9}
\definecolor{PaleBlue}{rgb}{0.7529, 0.9137, 0.9372}
\definecolor{BeauBlue}{rgb}{0.7686, 0.8470, 0.9529}
\definecolor{Mauve}{rgb}{0.8 , 0.7098, 0.9843}
\definecolor{PaleViolet}{rgb}{0.8156, 0.6431, 1.0}
\definecolor{Salmon}{rgb}{1.0, 0.8980, 0.6920}
\definecolor{Pink}{rgb}{1.0, 0.6902, 0.7908}
\definecolor{Mint}{rgb}{0.6902, 1.0, 0.7451}
\definecolor{SoftP}{rgb}{0.945, 0.933, 0.949}

\def\boxit#1{
  \smash{\color{red}\fboxrule=1pt\relax\fboxsep=2pt\relax%
  \llap{\rlap{\fbox{\vphantom{0}\makebox[#1]{}}}~}}\ignorespaces
}
\begin{table}[!htb]
        \captionof{table}{\textbf{Ablation study on different components of our GMP module} on C100 and CUB.}
        \centering
        \resizebox{1.0\columnwidth}{!}{%
        \centering
            \begin{tabular}{cc ccc ccc}
            \toprule
            \multicolumn{1}{c}{$\text{top-k}$} {\hskip 0.1in}
            & \multicolumn{1}{c}{GMM} {\hskip 0.1in}
            & \multicolumn{3}{c}{C100 Avg. \textit{ACC}}
            & \multicolumn{3}{c}{CUB Avg. \textit{ACC}} \\
            \multicolumn{1}{c}{Prompts} {\hskip 0.1in} &
            \multicolumn{1}{c}{Samples} {\hskip 0.1in} & 
            \multicolumn{1}{c}{\textit{All}  \cellcolor{blue!7!white}} & 
            \multicolumn{1}{c}{\textit{Old}} & 
            \multicolumn{1}{c}{\textit{New}} &
            \multicolumn{1}{c}{\textit{All}  \cellcolor{blue!7!white}} & 
            \multicolumn{1}{c}{\textit{Old}} & 
            \multicolumn{1}{c}{\textit{New}} 
            \\ \midrule
            \rowcolor{gray!5!white}
            $0$ & $0$ & $58.18$  \cellcolor{blue!7!white} & $72.27$ & $52.83$ & $54.98$  \cellcolor{blue!7!white} & $75.47$ & $48.15$ \\ \midrule
            \rowcolor{gray!5!white}
            $5$  & $0$ & $61.48$  \cellcolor{blue!7!white} & $74.68$ & $57.55$ & $53.54$  \cellcolor{blue!7!white} & $74.28$ & $46.47$ \\
            \rowcolor{gray!5!white}
            $5$ & $20$ & $62.21$  \cellcolor{blue!7!white} & $75.71$ & $57.90$ & $54.37$  \cellcolor{blue!7!white} & $74.88$ & $46.41$ \\
            \rowcolor{gray!5!white}
            $5$ & $200$ & $61.00$  \cellcolor{blue!7!white} & $72.46$ & $57.08$ & $51.67$  \cellcolor{blue!7!white} & $73.33$ & $44.08$ \\ \midrule
            \rowcolor{gray!5!white}
            $2$ & $100$ & $61.39$   \cellcolor{blue!7!white} & $73.04$ & $57.64$ & $53.36$  \cellcolor{blue!7!white} & $73.45$ & $46.04$ \\
            \rowcolor{blue!3!white}
            $\underline{5}$ & $\underline{100}$ & $\textbf{64.17}$  \cellcolor{blue!7!white} & $75.57$ & $60.34$ &  $\textbf{55.45}$  \cellcolor{blue!7!white} & $75.48$ & $48.56$ \\
            \rowcolor{gray!5!white}
            $10$ & $100$ & $61.03$  \cellcolor{blue!7!white} & $72.91$ & $56.97$ & $52.76$  \cellcolor{blue!7!white} & $71.67$ & $46.02$ \\ 
            \bottomrule
            \end{tabular}
        }
        \label{tab:ablation_result}
\end{table}

\noindent \textbf{The choices of numbers for \text{top-k}, pool length, and pool size in PLP.}
To investigate the effectiveness of our PLP module, we analyzed each component in our prompt module and present the results in Tab.~\ref{tab:ablation_result_plp}. The results show a clear advantage of adopting the PLP into our framework. The number of \text{top-k} prompts, pool length and the number of pool size are identified as important factors. The optimal configuration is \text{top-$2$} for prompt selection, pool length of $10$, and pool size of $20$, which appears to be a good trade-off. 
\definecolor{Gray}{gray}{0.9}
\definecolor{PaleBlue}{rgb}{0.7529, 0.9137, 0.9372}
\definecolor{BeauBlue}{rgb}{0.7686, 0.8470, 0.9529}
\definecolor{Mauve}{rgb}{0.8 , 0.7098, 0.9843}
\definecolor{PaleViolet}{rgb}{0.8156, 0.6431, 1.0}
\definecolor{Salmon}{rgb}{1.0, 0.8980, 0.6920}
\definecolor{Pink}{rgb}{1.0, 0.6902, 0.7908}
\definecolor{Mint}{rgb}{0.6902, 1.0, 0.7451}
\definecolor{SoftP}{rgb}{0.945, 0.933, 0.949}

\def\boxit#1{
  \smash{\color{red}\fboxrule=1pt\relax\fboxsep=2pt\relax%
  \llap{\rlap{\fbox{\vphantom{0}\makebox[#1]{}}}~}}\ignorespaces
}
\begin{table}[!htb]
        \captionof{table}{\textbf{Ablation study on different components of our PLP module} on CUB and Aircraft.}
        \centering
        \resizebox{\columnwidth}{!}{%
        \centering
            \begin{tabular}{ccc ccc ccc}
            \toprule
            \multicolumn{1}{c}{$\text{top-k}$}
            & \multicolumn{1}{c}{Pool}
            & \multicolumn{1}{c}{Pool}
            & \multicolumn{3}{c}{CUB Avg. \textit{ACC}}
            & \multicolumn{3}{c}{Aircraft Avg. \textit{ACC}} \\
            \multicolumn{1}{c}{Prompts}  &
            \multicolumn{1}{c}{Length} & 
            \multicolumn{1}{c}{Size} & 
            \multicolumn{1}{c}{\textit{All}  \cellcolor{blue!7!white}} & 
            \multicolumn{1}{c}{\textit{Old}} & 
            \multicolumn{1}{c}{\textit{New}} &
            \multicolumn{1}{c}{\textit{All}  \cellcolor{blue!7!white}} & 
            \multicolumn{1}{c}{\textit{Old}} & 
            \multicolumn{1}{c}{\textit{New}} 
            \\ \midrule
            \rowcolor{gray!5!white}
            $0$ & $0$ & $0$ & $66.70$  \cellcolor{blue!7!white} & $83.33$ & $60.81$ & $57.87$  \cellcolor{blue!7!white} & $63.80$ & $55.39$ \\ \midrule
            \rowcolor{gray!5!white}
            $1$  & $10$ & $20$ & $73.21$  \cellcolor{blue!7!white} & $85.00$ & $68.84$ & $61.85$  \cellcolor{blue!7!white} & $73.19$ & $57.94$ \\ \midrule
            \rowcolor{gray!5!white}
            $2$  & $5$ & $10$ & $72.27$  \cellcolor{blue!7!white} & $85.47$ & $67.38$ & $62.40$  \cellcolor{blue!7!white} & $77.86$ & $56.29$ \\
            \rowcolor{gray!5!white}
            $2$  & $10$ & $10$ & $74.37$  \cellcolor{blue!7!white} & $85.95$ & $70.12$ & $67.82$  \cellcolor{blue!7!white} & $75.71$ & $63.21$ \\
            \rowcolor{gray!5!white}
            $2$ & $20$ & $10$ & $73.66$  \cellcolor{blue!7!white} & $85.12$ & $69.58$ & $64.13$  \cellcolor{blue!7!white} & $74.29$ & $59.93$ \\ \midrule
            \rowcolor{gray!5!white}
            $2$ & $5$ & $20$ & $74.23$  \cellcolor{blue!7!white} & $84.76$ & $69.62$ & $66.13$  \cellcolor{blue!7!white} & $78.28$ & $62.62$ \\
            \rowcolor{blue!3!white}
            $\underline{2}$ & $\underline{10}$ & $\underline{20}$ & $\textbf{76.02}$  \cellcolor{blue!7!white} & $\textbf{86.31}$ & $\textbf{71.73}$ &  $\textbf{71.40}$ \cellcolor{blue!7!white} & $\textbf{78.57}$ & $\textbf{68.92}$ \\
            \rowcolor{gray!5!white}
            $2$ & $20$ & $20$ & $72.33$  \cellcolor{blue!7!white} & $84.17$ & $68.07$ & $68.30$  \cellcolor{blue!7!white} & $78.05$ & $64.61$ \\ \midrule
            \rowcolor{gray!5!white}
            $5$ & $10$ & $10$ & $72.95$  \cellcolor{blue!7!white} & $84.28$ & $68.52$ & $63.45$  \cellcolor{blue!7!white} & $74.29$ & $59.58$ \\
            \rowcolor{gray!5!white}
            $5$ & $10$ & $20$ & $73.02$ \cellcolor{blue!7!white} & $85.35$ & $68.55$ & $66.11$  \cellcolor{blue!7!white} & $78.52$ & $61.80$ \\
            \bottomrule
            \end{tabular}
        }
        \label{tab:ablation_result_plp}
\end{table}

\noindent \textbf{\text{Top-k} vs random prompts.}
In Tab.~\ref{tab: study on random-k}, to validate the effectiveness of using top-k prompts in our GMP \& PLP modules, we compare the results by using $\text{top-k}$ and random-k prompts.
We observe that using random-k prompts hurts the performance, as evidenced by the fact that random-$k$ performance is worse than using no prompts at all. In contrast, our top-k strategy leads to significantly improved performance, especially for the \textit{`New'} \textit{ACC}. This observation suggests that prompting with $\text{top-k}$ prompts indeed aids category discovery.
\definecolor{Gray}{gray}{0.9}
\definecolor{PaleBlue}{rgb}{0.7529, 0.9137, 0.9372}
\definecolor{BeauBlue}{rgb}{0.7686, 0.8470, 0.9529}
\definecolor{Mauve}{rgb}{0.8 , 0.7098, 0.9843}
\definecolor{PaleViolet}{rgb}{0.8156, 0.6431, 1.0}
\definecolor{Salmon}{rgb}{1.0, 0.8980, 0.6920}
\definecolor{Pink}{rgb}{1.0, 0.6902, 0.7908}
\definecolor{Mint}{rgb}{0.6902, 1.0, 0.7451}
\definecolor{SoftP}{rgb}{0.945, 0.933, 0.949}

\begin{table}[!htb]
    \caption{\textbf{Study on the effectiveness of GMP \& PLP's~\text{top-k} prompts} compared with random prompts.}
    \centering
    \resizebox{\columnwidth}{!}{%
    \centering
        \begin{tabular}{lcccccc}
        \toprule
        \multicolumn{1}{c}{}
        & \multicolumn{3}{c}{CIFAR100  Avg. \textit{ACC}}
        & \multicolumn{3}{c}{ImageNet-100  Avg. \textit{ACC}} 
        \\ 
        \cmidrule[0.1pt](r{0.80em}){2-4}
        \cmidrule[0.1pt](r{0.80em}){5-7}
        \multicolumn{1}{l}{PromptCCD} &
        \multicolumn{1}{c}{\textit{All}  \cellcolor{blue!7!white} } & 
        \multicolumn{1}{c}{\textit{Old}} &
        \multicolumn{1}{c}{\textit{New}} &
        \multicolumn{1}{c}{\textit{All}  \cellcolor{blue!7!white} } & 
        \multicolumn{1}{c}{\textit{Old}} &
        \multicolumn{1}{c}{\textit{New}} \\
        \midrule  
        \rowcolor{gray!5!white}
        $w/o$~GMP
        & $58.18$  \cellcolor{blue!7!white} & $72.27$  & $52.83$ 
        & $69.41$  \cellcolor{blue!7!white} & $81.56$  & $65.65$  \\
        \rowcolor{gray!5!white}
        GMP (random-k)
        & $59.98$  \cellcolor{blue!7!white}  & $73.81^{\textcolor{Green}{\textbf{+1.54}}}$  & $55.68^{\textcolor{Green}{\textbf{+2.85}}}$  
        & $68.30$  \cellcolor{blue!7!white}  & $80.09^{\textcolor{Red}{\textbf{-1.47}}}$  & $63.50^{\textcolor{Red}{\textbf{-2.15}}}$ \\
        \rowcolor{blue!3!white}
        GMP (top-k) (Ours)
        & $\textbf{64.17}$  \cellcolor{blue!7!white}  & $75.57^{\textcolor{Green}{\textbf{+3.30}}}$  & $60.34^{\textcolor{Green}{\textbf{+7.51}}}$ 
        & $\textbf{76.16}$  \cellcolor{blue!7!white}  & $81.76^{\textcolor{Green}{\textbf{+0.20}}}$  & $74.35^{\textcolor{Green}{\textbf{+8.70}}}$ \\
        \midrule \midrule
        PromptCCD++ & \textit{All} \cellcolor{blue!7!white} & \textit{Old} & \textit{New} & \textit{All} \cellcolor{blue!7!white} & \textit{Old} & \textit{New}\\
        \midrule  
        \rowcolor{gray!5!white}
        $w/o$~PLP
        & $65.36$  \cellcolor{blue!7!white} & $77.06$  & $60.46$ 
        & $71.58$  \cellcolor{blue!7!white} & $83.02$  & $68.05$  \\
        \rowcolor{gray!5!white}
        PLP (random-k)
        & $71.46$  \cellcolor{blue!7!white}  & $80.26^{\textcolor{Green}{\textbf{+3.20}}}$  & $63.73^{\textcolor{Green}{\textbf{+3.27}}}$  
        & $73.09$  \cellcolor{blue!7!white}  & $78.77^{\textcolor{Red}{\textbf{-4.25}}}$  & $73.49^{\textcolor{Green}{\textbf{+5.44}}}$ \\
        \rowcolor{blue!3!white}
        PLP (top-k) (Ours)
        & $\textbf{77.68}$  \cellcolor{blue!7!white}  & $83.18^{\textcolor{Green}{\textbf{+6.12}}}$  & $75.89^{\textcolor{Green}{\textbf{+15.43}}}$ 
        & $\textbf{83.39}$  \cellcolor{blue!7!white}  & $85.06^{\textcolor{Green}{\textbf{+2.04}}}$  & $82.17^{\textcolor{Green}{\textbf{+14.12}}}$ \\
        \toprule
        \multicolumn{1}{c}{}
        & \multicolumn{3}{c}{TinyImageNet Avg. \textit{ACC}} 
        & \multicolumn{3}{c}{CUB  Avg. \textit{ACC}} 
        \\ 
        \cmidrule[0.1pt](r{0.80em}){2-4}
        \cmidrule[0.1pt](r{0.80em}){5-7}
        \multicolumn{1}{l}{PromptCCD} &
        \multicolumn{1}{c}{\textit{All}  \cellcolor{blue!7!white}} & 
        \multicolumn{1}{c}{\textit{Old}} &
        \multicolumn{1}{c}{\textit{New}} &
        \multicolumn{1}{c}{\textit{All}  \cellcolor{blue!7!white}} & 
        \multicolumn{1}{c}{\textit{Old}} &
        \multicolumn{1}{c}{\textit{New}} \\
        \midrule  
        \rowcolor{gray!5!white}
        $w/o$~GMP
        & $55.20$  \cellcolor{blue!7!white} & $65.87$  & $51.61$ 
        & $54.98$  \cellcolor{blue!7!white} & $75.47$  & $48.15$  \\
        \rowcolor{gray!5!white}
        GMP (random-k)
        & $55.69$  \cellcolor{blue!7!white} & $63.95^{\textcolor{Red}{\textbf{-1.92}}}$  & $52.52^{\textcolor{Green}{\textbf{+0.91}}}$ 
        & $51.46$  \cellcolor{blue!7!white} & $73.10^{\textcolor{Red}{\textbf{-2.37}}}$  & $43.90^{\textcolor{Red}{\textbf{-4.25}}}$  \\
        \rowcolor{blue!3!white}
        GMP (top-k) (Ours)
        & $\textbf{61.84}$  \cellcolor{blue!7!white}  & $66.54^{\textcolor{Green}{\textbf{+0.67}}}$  & $60.26^{\textcolor{Green}{\textbf{+8.65}}}$ 
        & $\textbf{55.45}$  \cellcolor{blue!7!white}  & $75.48^{\textcolor{Green}{\textbf{+0.01}}}$  & $48.56^{\textcolor{Green}{\textbf{+0.41}}}$  \\
        \midrule \midrule
        PromptCCD++ & \textit{All} \cellcolor{blue!7!white} & \textit{Old} & \textit{New} & \textit{All} \cellcolor{blue!7!white} & \textit{Old} & \textit{New}\\
        \midrule 
        \rowcolor{gray!5!white}
        $w/o$~PLP
        & $59.05$  \cellcolor{blue!7!white} & $77.44$  & $53.41$ 
        & $66.70$  \cellcolor{blue!7!white} & $83.33$  & $60.81$  \\
        \rowcolor{gray!5!white}
        PLP (random-k)
        & $63.31$  \cellcolor{blue!7!white}  & $78.80^{\textcolor{Green}{\textbf{+1.36}}}$  & $60.47^{\textcolor{Green}{\textbf{+7.06}}}$  
        & $66.74$  \cellcolor{blue!7!white}  & $84.52^{\textcolor{Green}{\textbf{+1.19}}}$  & $59.79^{\textcolor{Red}{\textbf{-1.02}}}$ \\
        \rowcolor{blue!3!white}
        PLP (top-k) (Ours)
        & $\textbf{73.02}$  \cellcolor{blue!7!white}  & $80.15^{\textcolor{Green}{\textbf{+2.71}}}$  & $70.54^{\textcolor{Green}{\textbf{+17.13}}}$ 
        & $\textbf{76.02}$  \cellcolor{blue!7!white}  & $86.31^{\textcolor{Green}{\textbf{+2.98}}}$  & $71.73^{\textcolor{Green}{\textbf{+10.92}}}$ \\
        \bottomrule 
        \end{tabular}
     }
    \label{tab: study on random-k}
\end{table}

\noindent\textbf{Knowledge retention teacher weight.}
Tab.~\ref{tab: plp teacher weight} studies the distillation weight $\lambda_{\text{dist}}$ that balances knowledge retention and adaptation during discovery stages. Disabling distillation ($\lambda_{\text{dist}}{=}0$) causes notable degradation on both datasets, confirming that unconstrained prompt pool updates lead to representational drift. A moderate weight of $\lambda_{\text{dist}}{=}0.2$ achieves the best performance (77.68\% on C100, 76.02\% on CUB), while larger values ($\geq 0.7$) over-regularise the model and reduce \textit{`New'} \textit{ACC}, limiting its ability to learn discriminative features for newly discovered categories.
\definecolor{Gray}{gray}{0.9}
\definecolor{PaleBlue}{rgb}{0.7529, 0.9137, 0.9372}
\definecolor{BeauBlue}{rgb}{0.7686, 0.8470, 0.9529}
\definecolor{Mauve}{rgb}{0.8 , 0.7098, 0.9843}
\definecolor{PaleViolet}{rgb}{0.8156, 0.6431, 1.0}
\definecolor{Salmon}{rgb}{1.0, 0.8980, 0.6920}
\definecolor{Pink}{rgb}{1.0, 0.6902, 0.7908}
\definecolor{Mint}{rgb}{0.6902, 1.0, 0.7451}
\definecolor{SoftP}{rgb}{0.945, 0.933, 0.949}

\begin{table}[!htb]
    \centering
        \caption{\textbf{Study on the PLP knowledge retention teacher's weight} on C100 and CUB.}
    \centering
    \resizebox{0.9\columnwidth}{!}{%
    \centering
        \begin{tabular}{c ccc ccc}
        \toprule
        \multicolumn{1}{c}{}
        & \multicolumn{3}{c}{C100 Avg. \textit{ACC}}
        & \multicolumn{3}{c}{CUB Avg. \textit{ACC}} 
        \\ 
        \cmidrule[0.1pt](r{0.80em}){2-4}
        \cmidrule[0.1pt](r{0.80em}){5-7}
        \multicolumn{1}{c}{$\lambda_{dist}$} &
        \multicolumn{1}{c}{\textit{All}  \cellcolor{blue!7!white} } & 
        \multicolumn{1}{c}{\textit{Old}} &
        \multicolumn{1}{c}{\textit{New}} &
        \multicolumn{1}{c}{\textit{All}  \cellcolor{blue!7!white} } & 
        \multicolumn{1}{c}{\textit{Old}} &
        \multicolumn{1}{c}{\textit{New}} \\
        \midrule  
        \rowcolor{gray!5!white}
        $0$
        & $73.85$  \cellcolor{blue!7!white} & $79.11$  & $71.71$ 
        & $71.78$  \cellcolor{blue!7!white} & $84.43$  & $65.79$  \\
        \rowcolor{blue!3!white}
        $0.2$
        & $\textbf{77.68}$  \cellcolor{blue!7!white} & $83.18$  & $75.89$ 
        & $\textbf{76.02}$  \cellcolor{blue!7!white} & $86.31$  & $71.73$  \\
        \rowcolor{gray!5!white}
        $0.7$
        & $76.50$  \cellcolor{blue!7!white} & $82.67$  & $74.09$ 
        & $74.15$  \cellcolor{blue!7!white} & $85.95$  & $69.93$  \\
        \rowcolor{gray!5!white}
        $1.0$
        & $75.89$  \cellcolor{blue!7!white} & $83.44$  & $73.84$ 
        & $73.32$  \cellcolor{blue!7!white} & $85.59$  & $68.96$  \\
        \bottomrule 
        \end{tabular}
    }
    \label{tab: plp teacher weight}
\end{table} 

\noindent\textbf{Routing loss weight.}
Tab.~\ref{tab: plp routing weight} studies the routing supervision weight $\lambda_{\text{route}}$, which controls the strength of the cross-entropy loss over patch-to-part assignments during the initial stage. A small weight ($\lambda_{\text{route}}{=}0.01$) under-supervises the router, resulting in suboptimal part assignments that propagate to weaker CCD performance. Conversely, large values ($\lambda_{\text{route}} \geq 0.3$) over-emphasise routing accuracy at the expense of the contrastive representation objective, degrading CCD accuracy on CUB by up to 4~pp. A moderate value of $\lambda_{\text{route}}{=}0.1$ yields the best overall performance across both datasets (77.68\% on C100, 76.02\% on CUB), balancing accurate part decomposition with effective representation learning.
\begin{table}[!htb]
    \centering
        \caption{\textbf{Study on the PLP routing loss weight} on C100 and CUB.}
    \centering
    \resizebox{0.9\columnwidth}{!}{%
    \centering
        \begin{tabular}{c ccc ccc}
        \toprule
        \multicolumn{1}{c}{}
        & \multicolumn{3}{c}{C100 Avg. \textit{ACC}}
        & \multicolumn{3}{c}{CUB Avg. \textit{ACC}} 
        \\ 
        \cmidrule[0.1pt](r{0.80em}){2-4}
        \cmidrule[0.1pt](r{0.80em}){5-7}
        \multicolumn{1}{c}{$\lambda_{route}$} &
        \multicolumn{1}{c}{\textit{All}  \cellcolor{blue!7!white} } & 
        \multicolumn{1}{c}{\textit{Old}} &
        \multicolumn{1}{c}{\textit{New}} &
        \multicolumn{1}{c}{\textit{All}  \cellcolor{blue!7!white} } & 
        \multicolumn{1}{c}{\textit{Old}} &
        \multicolumn{1}{c}{\textit{New}} \\
        \midrule  
        \rowcolor{gray!5!white}
        $0.01$
        & $76.88$  \cellcolor{blue!7!white} & $82.70$  & $74.97$ 
        & $73.98$  \cellcolor{blue!7!white} & $85.60$  & $68.89$  \\
        \rowcolor{blue!3!white}
        $0.1$
        & $\textbf{77.68}$  \cellcolor{blue!7!white} & $83.18$  & $75.89$ 
        & $\textbf{76.02}$  \cellcolor{blue!7!white} & $86.31$  & $71.73$  \\
        \rowcolor{gray!5!white}
        $0.3$
        & $76.96$  \cellcolor{blue!7!white} & $82.98$  & $75.02$ 
        & $74.23$  \cellcolor{blue!7!white} & $84.88$  & $69.75$ \\
        \rowcolor{gray!5!white}
        $0.5$
        & $75.17$  \cellcolor{blue!7!white} & $82.88$  & $72.68$ 
        & $72.08$  \cellcolor{blue!7!white} & $85.95$  & $66.51$  \\
        \bottomrule 
        \end{tabular}
    }
    \label{tab: plp routing weight}
\end{table}

\noindent\textbf{Ablation of the part prompt router design.}
Tab.~\ref{tab:ppr_ablation} ablates the two routing signals in PLP's context-aware patch router: attention fusion (Eq.~\ref{eq: attn_fusion}) via the Transformer encoder with learnable part queries, and attention-score (Eq.~\ref{eq:routing}) routing via scaled dot-product affinity. In addition to the downstream CCD accuracy, we evaluate the part routing accuracy by comparing the router's predicted part assignments against pseudo part labels. \textit{`All'} metrics are averaged over three discovery stages. Starting from the MLP-only baseline~(a), adding either signal individually~(b,~c) improves both part routing and CCD accuracy. Combining both~(d) consistently achieves the best CCD performance on both datasets (77.68\% on C100, 76.02\% on CUB) along with the highest part routing accuracy, confirming that the two signals are complementary: attention fusion captures spatial context, while attention scores provide direct part affinity, together yielding more discriminative patch-to-part assignments for Continual Category Discovery.
\newcommand{\cmark}{\ding{51}}%
\newcommand{\xmark}{\ding{55}}%
\begin{table}[!htb]
    \centering
    \caption{\textbf{Ablation study on PLP's part prompt routing design components.}
    We ablate the attention fusion and attention-score routing
    of the part prompt router and report the average accuracy
    on C100 and CUB.}
    \resizebox{\columnwidth}{!}{%
    \begin{tabular}{clcc ccc ccc}
        \toprule
        \multicolumn{2}{c}{}
        & &
        & \multicolumn{3}{c}{Part routing Avg.\ \textit{ACC}}
        & \multicolumn{3}{c}{CCD Avg.\ \textit{ACC}} \\
        \cmidrule[0.1pt](r{0.80em}){5-7}
        \cmidrule[0.1pt](r{0.80em}){8-10}
        \multicolumn{1}{l}{} &
        \multicolumn{1}{c}{} &
        Attn.\ Fusion & Attn.\ Scores
        & \textit{All}  \cellcolor{blue!7!white}
        & \textit{Old}
        & \textit{New}
        & \textit{All}  \cellcolor{blue!7!white}
        & \textit{Old}
        & \textit{New} \\
        \cmidrule[0.1pt](r{0.80em}){1-2}
        \cmidrule[0.1pt](r{0.80em}){3-4}
        \cmidrule[0.1pt](r{0.80em}){5-7}
        \cmidrule[0.1pt](r{0.80em}){8-10}
        \rowcolor{gray!5!white}
        & (a) & \xmark & \xmark
        & $64.71$ \cellcolor{blue!7!white} & $64.49$ & $66.73$ 
        & $73.27$ \cellcolor{blue!7!white} & $79.26$ & $73.06$ \\
        \rowcolor{gray!5!white}
        & (b) & \xmark & \cmark
        & $66.08$ \cellcolor{blue!7!white} & $66.76$ & $68.26$
        & $74.49$ \cellcolor{blue!7!white} & $80.34$ & $74.80$ \\
        \rowcolor{gray!5!white}
        & (c) & \cmark & \xmark
        & $67.50$ \cellcolor{blue!7!white} & $67.19$ & $67.62$
        & $76.94$ \cellcolor{blue!7!white} & $83.43$ & $74.92$ \\
        \rowcolor{blue!3!white}
        \rot{\rlap{C100}}
        & (d) & \cmark & \cmark
        & $\textbf{68.55}$ \cellcolor{blue!7!white} & $68.41$ & $70.11$
        & $\textbf{77.68}$ \cellcolor{blue!7!white} & $83.18$ & $75.89$ \\
        \midrule
        \rowcolor{gray!5!white}
        & (a) & \xmark & \xmark
        & $83.55$ \cellcolor{blue!7!white} & $82.42$ & $83.95$  
        & $72.02$ \cellcolor{blue!7!white} & $85.00$ & $67.05$ \\
        \rowcolor{gray!5!white}
        & (b) & \xmark & \cmark
        & $89.57$ \cellcolor{blue!7!white} & $89.44$ & $89.96$
        & $74.72$ \cellcolor{blue!7!white} & $86.07$ & $70.31$ \\
        \rowcolor{gray!5!white}
        & (c) & \cmark & \xmark
        & $89.51$ \cellcolor{blue!7!white} & $89.38$ & $89.91$
        & $73.70$ \cellcolor{blue!7!white} & $86.55$ & $68.88$ \\
        \rowcolor{blue!3!white}
        \rot{\rlap{CUB}}
        & (d) & \cmark & \cmark
        & $\textbf{89.94}$ \cellcolor{blue!7!white} & $89.85$ & $90.26$
        & $\textbf{76.02}$ \cellcolor{blue!7!white} & $86.31$ & $71.73$ \\
        \bottomrule
    \end{tabular}
    }
    \label{tab:ppr_ablation}
\end{table}
\section{Conclusion}
\label{sec:conclusion}
In this paper, we study effective prompt pool learning for Continual Category Discovery (CCD), introducing a series of prompt pool designs that address the challenge of discovering novel categories from a continuous stream of unlabelled data while mitigating catastrophic forgetting. Our first design, PromptCCD, focuses on global class prototypes via the Gaussian Mixture Prompting (GMP) module, which provides a principled, label-free approach to prompt selection and enables on-the-fly category number estimation, eliminating the need for prior knowledge of the category count. Through a systematic spectrum study, we reveal that category count, rather than sample volume, is the primary bottleneck for discovery performance, motivating the need for finer-grained representations. Our second design, PromptCCD++, focuses on object-part prototypes via the Part-Level Prompting (PLP) module, which decomposes the prompt pool into multiple part-specific prompt pools. During discovery phase, a context-aware router automatically assigns spatial patches to part pools without manual annotations, enabling the model to capture fine-grained structural dependencies that transfer effectively to novel categories. Extensive evaluations on seven generic and fine-grained benchmarks, using our continual accuracy metric \textit{cACC}, demonstrate that our prompt pool designs achieve state-of-the-art performance, confirming that effective prompt pool learning provides a robust foundation for CCD.

\section*{Acknowledgement}
This work is supported by the Hong Kong Research Grants Council - General Research Fund (Grant No.: $17211024$).

\ifCLASSOPTIONcaptionsoff
  \newpage
\fi

\bibliographystyle{IEEEtran}
\bibliography{egbib}
\ifCLASSOPTIONcaptionsoff
  \newpage
\fi

\clearpage
\onecolumn

\begin{center}
    \Large{\textbf{Effective Prompt Pool Learning for Continual Category Discovery}} \\
    \textit{\textbf{\Large{--Supplementary Material--}}}
\end{center}

\setcounter{table}{0}
\setcounter{figure}{0}
\setcounter{algorithm}{0}
\setcounter{equation}{0}
\renewcommand{\thetable}{\Alph{table}}
\renewcommand\thefigure{\Alph{figure}} 
\renewcommand{\thealgorithm}{\Alph{algorithm}} 
\renewcommand{\thesection}{S\arabic{section}}
\renewcommand\theequation{\alph{equation}}

We provide this supplementary material to further support our main paper. 
We begin by providing the pseudo-code implementations for PromptCCD \& PromptCCD++ in Sec.~\ref{pseudo code promptccd}~\&~\ref{pseudo code promptccd++} respectively. In Sec.~\ref{supp:constructing_labels}, we provide details of PLP's part labels construction. Next, in Sec.~\ref{supp: complete}, we present the breakdown CCD results, benchmarking the performance of each compared model across different stages for cases when number of categories $C$ are known and unknown. Additionally, we delve into inductive evaluation scenarios (Sec.~\ref{supp: trans and induc}), evaluation on standard GCD metric (Sec.~\ref{supp: adapt gcd}), additional comparison with other CCD settings (Sec.~\ref{supp: recent works}), analysis on different class splits scenarios (Sec.~\ref{supp: diff class ratio}), and qualitative results (Sec.~\ref{supp: extra quali}) in separate sections. Our implementation details cover aspects such as our implementation of Grow \& Merge to enhance it with ViT in Sec.~\ref{supp: gm on vit}, fine-tuning method and parameter analysis in Sec.~\ref{supp: learn parameters}. Finally, we discuss both the impacts and limitations of our model for future studies in Sec.~\ref{supp: limitations}.

\vspace{5mm}
\makeatletter
\let\supp@orig@addcontentsline\addcontentsline
\renewcommand{\addcontentsline}[3]{%
  \def\supp@ext{#1}%
  \def\supp@toc{toc}%
  \ifx\supp@ext\supp@toc
    \supp@orig@addcontentsline{stc}{#2}{#3}%
  \else
    \supp@orig@addcontentsline{#1}{#2}{#3}%
  \fi
}
\makeatother

\begingroup
\let\clearpage\relax
\setcounter{tocdepth}{2}
\hypersetup{linkcolor=black}
\begin{center}
    \textbf{Contents}
\end{center}
\makeatletter
\@starttoc{stc}
\makeatother
\hypersetup{linkcolor=blue}
\endgroup
\newpage

\clearpage
\section{Pseudo Code for PromptCCD}
\label{pseudo code promptccd}
\begin{algorithm}[H]
\begin{algorithmic}[1]
\caption{PromptCCD's Pseudo Code.}
\Statex \textbf{Require:} $\mathcal{H}_{\theta}: \{\phi, f_{\theta}\}$ where $f_{\theta}:\{f_e, f_b\}$.
\Statex \textbf{Require:} \Call{GMP}{} prompt module where it contains $\text{GMM}_t$.
\Statex \textbf{Require:} Dataloader $\mathcal{B}$ for dataset $D_t$ at stage $t$.
\State Set $\alpha \gets$  integer value for the incremental update epoch.
\State Set $\beta \gets$ integer value for the warmup epoch.
\Procedure {PromptCCD}{$\mathcal{H}_{\theta}$, \Call{GMP}{}, $\mathcal{B}$} at stage $t$.
\LComment{************************ Start training ************************}
\For{$e \in Epochs$}
    \State \hfill
    \LComment{fit $GMM_t$ every n increment of epoch.}
    \If{$0 \equiv e \pmod{\alpha}$} 
        \State  $\mathcal{Z}_t \gets \{f_{\theta}(x)|x\in D_{t} \}$ \Comment{extract features $\texttt{[stop gradient]}$}
        \If{$t > 0$}
            \State $\mathcal{Z}^{s}_{t-1}$ $\gets$ $\Call{Generate-random-samples}{{\text{GMM}}_{t-1}}$.

            \State $\mathcal{Z}_{t}$ $\gets$ $\mathcal{Z}_{t}\cup\mathcal{Z}^{s}_{t-1}$ 
            \Comment{combine with generated samples from $\text{GMM}_{t-1}$.}
        \EndIf
        \State $\Call{Optimize}{\Call{GMP}{}}$ by Fitting $\text{GMM}_t$ with $\mathcal{Z}_t$
    \EndIf

    \State \hfill
    \For{$B:\{x_i, x'_i\} \in \mathcal{B}$} \Comment{assume a batch $B$ only contains a set $\{x_i, x^{'}_i\}$.}
    
        \State \hfill
        \BeginBox[draw=blue, dashed] 
        \LComment{the next lines covered in this box describe how to acquire $\mu_{\text{top-k}}$.}
        \If{$e > \beta$} \Comment{when the model reaches the warm-up epoch.}
            \State $\hat{z}_i \gets f_\theta(x_i)$ \Comment{extract features $\texttt{[stop gradient]}$.}
            \State $\mu_{\text{top-k}}$ $\gets$ $\Call{GMP}{\hat{z}_i | \text{GMM}_t}$ \Comment{see Fig.~$3$
            in main paper for details.}
        \Else
            \State $\mu_{\text{top-k}} \gets \textit{None}$
        \EndIf 
        \State \hfill
        \EndBox

        \BeginBox[draw=blue, dashed]
        \LComment{the next lines covered in this box describe how $x_i$ and $\mu_{\text{top-k}}$ are projected into the model 
        [note: same operation for $x^{'}_i$].}
        \State $x_q \gets \Call{patchify}{x_i}$ \Comment{patchify image $x_i$ into $L$ patches.}
        \State $x_e \gets f_{e}(x_q)$ \Comment{project to pretrained patch embedding layer.}
        \State $x_{total} \gets [\mu_{\text{top-k}}; x_{e}]$ \Comment{concatenate $x_e$ with the $\mu_{\text{top-k}}$ prompts.}
        \State $z_i \gets \phi(f_{b}(x_{total}))$ \Comment{project to self-attention blocks and projection head.}
        \EndBox

        \LComment{to summarize above operations, from $\mathcal{H}_{\theta}: \{\phi, f_{\theta}\}$, we got:}
        \State $z_i \gets \phi(f_{\theta}(x_{i}))$ $\And{z'_i \gets \phi(f_{\theta}(x'_{i}))}$

    \EndFor
    \State \hfill
    \LComment{optimize $\mathcal{H}_{\theta}$, (see Sec.~$3.1$ in main paper) and do $\texttt{[gradient update]}$.}
    \State \Call{Optimize}{$\mathcal{H}_{\theta}$}
\EndFor

\LComment{************************* End training *************************}
\EndProcedure
\end{algorithmic}
\label{alg:pseudocode}
\end{algorithm}
\clearpage
\section{Pseudo Code for PromptCCD++}
\label{pseudo code promptccd++}
\begin{algorithm}[H]
\begin{algorithmic}[1]
\caption{PromptCCD++'s Pseudo Code.}
\Statex \textbf{Require:} $\mathcal{H}_{\theta}: \{\phi, f_{\theta}\}$ where $f_{\theta}:\{f_e, f_b\}$.
\Statex \textbf{Require:} \Call{PLP}{} prompt module with part-level pools $\mathbb{P}=\{(K_p, V_p)\}_{p=1}^{P}$ and router $\mathcal{R}_{\psi}=\{\text{AttnProj}, \text{TransEnc}, g_{\psi}, Q\}$.
\Statex \textbf{Require:} Dataloader $\mathcal{B}$ for dataset $D_t$ at stage $t$.
\Statex \textbf{Require:} Frozen teacher model $\mathcal{H}_{\theta}^{(0)}$ for stages $t>0$.
\State Set $\lambda_{\text{route}} \gets$ routing loss weight.
\State Set $\lambda_{\text{dist}} \gets$ distillation loss weight.

\Procedure {PromptCCD++}{$\mathcal{H}_{\theta}$, \Call{PLP}{}, $\mathcal{B}$} at stage $t$.
\LComment{************************ Start training ************************}

\LComment{Router learning and stage-wise freezing.}
\If{$t > 0$}
    \State \Call{Freeze}{$\psi$} \Comment{freeze router after initial stage}
\EndIf

\For{$e \in Epochs$}
    \For{$B:\{x_i, x'_i\} \in \mathcal{B}$} \Comment{assume a batch $B$ contains $\{x_i, x'_i\}$.}

        \State \hfill
        \BeginBox[draw=blue, dashed]
        \LComment{the next lines covered in this box describe PLP forward for $x_i$.}
        \LComment{Part-level prompt pool decomposition + context-aware patch routing + part-specific modulation.}
        \State $x_q \gets \Call{patchify}{x_i}$
        \State $x_e \gets f_e(x_q)$
        \State $\alpha \gets \Call{CLS-to-patch-attention}{x_i}$ \Comment{attention from last self-attention layer}
        \State $x_{\text{aug}} \gets x_e + \Call{AttnProj}{\alpha}$
        \State $\tilde{x}_i \gets \Call{TransEnc}{\Call{LN}{x_{\text{aug}}}}$
        \State $r_i \gets g_{\psi}(\tilde{x}_i) + \tilde{x}_i Q^\top / \sqrt{z}$
        \State $\sigma(r_i) \gets \Call{softmax}{r_i}$
        \State $p^{*}_{l} \gets \arg\max_{p}\ \sigma(r_i)_{l,p},\ \forall l \in \{1,\dots,L\}$
        \For{$l \in \{1,\dots,L\}$}
            \State $\mathcal{V}^{(l)}_{\text{top-k}} \gets \{V_{p^{*}_{l},m}\ |\ K_{p^{*}_{l},m}\in \mathcal{T}^{k}_{K_{p^{*}_{l}}}(\tilde{x}_{i,l})\}$
            \State $b_l \gets \Call{Average}{\mathcal{V}^{(l)}_{\text{top-k}}}$ \Comment{average over top-k and prompt length}
            \State $x^{plp}_{e,l} \gets x_{e,l} \odot (1 + b_l)$
        \EndFor
        \State $z_i \gets \phi(f_b(x^{plp}_e)),\ \bar{z}_i \gets \Call{Normalize}{z_i}$
        \State $\bar{\sigma}(r_i) \gets \frac{1}{L}\sum_{l=1}^{L}\sigma(r_{i,l})$
        \EndBox


        \If{$t = 0$}
            \LComment{Initial stage: supervise router with pseudo part labels.}
            \State $y^{\text{part}} \gets \Call{Get-Pseudo-Part-Labels}{B}$
            \State $\mathcal{L}^{\text{route}} \gets \frac{1}{L}\sum_{l=1}^{L}\Call{CE}{r_{i,l},\ y^{\text{part}}_{l}}$ 
            \State $\mathcal{L}^{\text{distill}} \gets 0,\ \mathcal{L}^{\text{anchor}} \gets 0$
        \Else
            \LComment{Knowledge retention via feature/routing distillation from frozen teacher.}
            \State $(\bar{z}_i^{teacher}, r_i^{teacher}) \gets \Call{Forward-Teacher}{x_i,\ \mathcal{H}_{\theta}^{(0)}}$ 
            \State $\bar{\sigma}(r_i^{teacher}) \gets \frac{1}{L}\sum_{l=1}^{L}\sigma(r^{teacher}_{i,l})$ 
            \State $\mathcal{L}^{\text{distill}} \gets \|\bar{z}_i-\bar{z}_i^{teacher}\|_2^2$
            \State $\mathcal{L}^{\text{anchor}} \gets D_{\text{KL}}(\bar{\sigma}(r_i)\|\bar{\sigma}(r_i^{teacher}))$
            \State $\mathcal{L}^{\text{route}} \gets 0$
        \EndIf

        \LComment{Overall objective (see Eq.~$16$, main paper) and do $\texttt{[gradient update]}$.}
        \State \Call{Optimize}{$\mathcal{H}_{\theta}$ and \Call{PLP}{}} \Comment{$\psi$ remains frozen when $t>0$}
    \EndFor
\EndFor

\If{$t = 0$}
    \State $\mathcal{H}_{\theta}^{(0)} \gets \Call{Copy-and-Freeze}{\mathcal{H}_{\theta}}$ \Comment{save frozen teacher}
\EndIf
\State Transfer updated part-level prompt pools $\mathbb{P}$ to stage $t+1$.
\LComment{************************* End training *************************}
\EndProcedure
\end{algorithmic}
\label{alg:pseudocode_promptccdpp}
\end{algorithm}
\clearpage
\section{Constructing Part-Level Correspondence Labels}
\label{supp:constructing_labels}

\begin{figure}[!ht]
    \centering
    \includegraphics[width=0.9\textwidth]{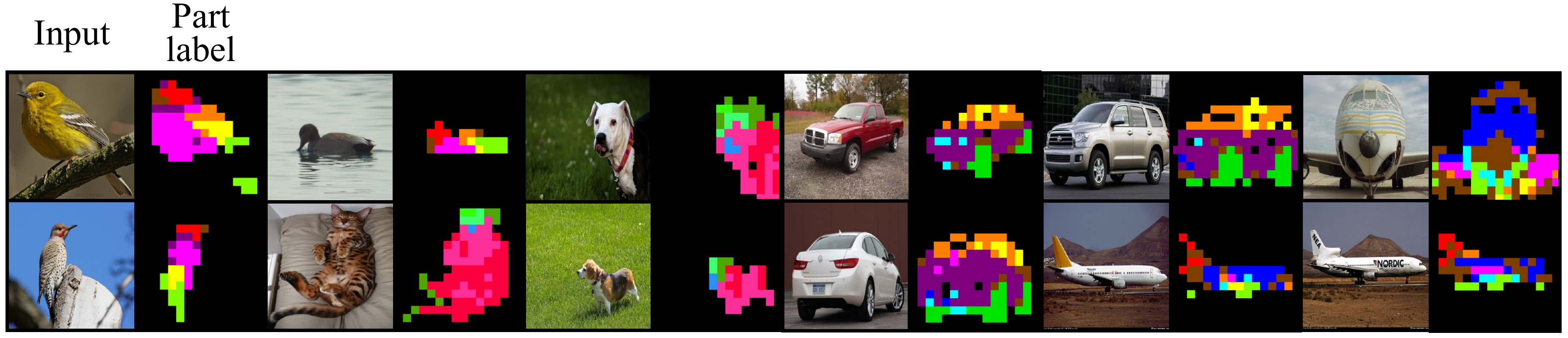}
    \caption{
    \textbf{Visualization of PLP's part labels.}
    }
    \label{fig:label_vis}
\end{figure}

We follow PartCo~\cite{Cendra2025PartCo} to construct the part-level correspondence labels used to supervise the PLP router at the initial stage. The goal is to obtain a stable patch-wise partition for each dataset using only frozen DINOv2 features, so that the same label construction procedure can be applied consistently across labeled and unlabeled data. As illustrated in Fig.~\ref{fig:label_vis}, the construction consists of two stages: first, we derive object-aware and fine-grained projections from patch tokens via PCA; second, we apply \textit{k}-means clustering to assign part labels at patch resolution.

\noindent\textbf{Stage 1: PCA-based feature projection.}
We begin by sampling a subset of labeled images from the labeled split $D^{l}$, selecting one image for each labeled known category. For each image $x_i$, we extract patch-token features from the frozen DINOv2 backbone~\cite{oquab2023dinov2}. Let $\mathbf{f}_{i,n} \in \mathbb{R}^{d}$ denote the feature of the $n$-th patch token in image $x_i$, where $d$ is the feature dimension. By stacking all sampled patch tokens, we obtain a feature matrix: $\mathbf{F} \in \mathbb{R}^{T \times d}$, where $T$ is the total number of sampled patch tokens. We then compute the first principal direction:
\begin{equation}
    \mathbf{w}_{\text{obj}} = \arg\max_{\mathbf{w}} \frac{\mathbf{w}^\top \mathbf{\Sigma} \mathbf{w}}{\mathbf{w}^\top \mathbf{w}},
\end{equation}
where $\mathbf{\Sigma}$ is the covariance matrix of $\mathbf{F}$. This projection captures the dominant variation in the patch features and is used to estimate an objectness score for each patch: $s^{\text{obj}}_{i,n} = \mathbf{f}_{i,n}^{\top}\mathbf{w}_{\text{obj}}$. We threshold the scores with $\tau_{\text{obj}} = 0.6$ to obtain a binary foreground mask:
\begin{equation}
    m_{i,n} = \mathds{1}\bigl(s^{\text{obj}}_{i,n} > \tau_{\text{obj}}\bigr),
\end{equation}
which separates foreground patches from background patches. Next, we perform a second PCA on the masked features to obtain a compact fine-grained representation. Specifically, we compute:
\begin{equation}
    \mathbf{f}^{\text{fg}}_{i,n} = \bigl(\mathbf{f}_{i,n} \odot m_{i,n}\bigr)\mathbf{W}_{\text{fg}},
\end{equation}
where $\mathbf{W}_{\text{fg}} \in \mathbb{R}^{d \times 3}$ contains the top three principal components. This maps each patch feature into a three-dimensional space that is suitable for clustering and visualization. Before clustering, we normalize the projected features $\mathbf{f}^{\text{fg}}_{i,n}$ to reduce scale sensitivity.

\noindent\textbf{Stage 2: Label construction by \textit{k}-means.}
Given the normalized fine-grained patch features, we sweep over a candidate set of cluster numbers and apply \textit{k}-means clustering to obtain part-level assignments for the labeled split $D^{l}$. For each candidate $k$, we evaluate the clustering solution using two criteria: (i) the minimum distance between cluster centers, which encourages well-separated clusters, and (ii) the balance of cluster sizes, which avoids degenerate solutions where a few clusters dominate. We select the optimal number of parts $k^*$ by maximizing:
\begin{equation}
    \min_{i \neq j} \|\mathbf{k}_i - \mathbf{k}_j\| \times
    \left(\frac{\min_i |S_i|}{\max_j |S_j|}\right),
\end{equation}
where $\mathbf{k}_i$ denotes the centroid of cluster $i$ and $|S_i|$ is the number of patch samples assigned to cluster $i$.

With $k^*$ fixed, we assign part-level correspondence labels to all patches in the full dataset $\mathbf{D}$. For each patch $l$, the final part label is given by the nearest cluster center:
\begin{equation}
    y^{\text{part}}_{l} = \arg\min_{\mathbf{k} \in \mathcal{K}} \|\mathbf{f}^{\text{fg}}_{l} - \mathbf{k}\|,
\end{equation}
where $\mathcal{K} = \{\mathbf{k}_1, \mathbf{k}_2, \dots, \mathbf{k}_{k^*}\}$ is the set of optimal cluster centers. The resulting patch-wise assignments are reshaped back to the ViT patch grid to form the final part-label maps used to supervise the router in PromptCCD++.

\begin{table}[!ht]
    \centering
    \caption{Number of parts (including background) used for each dataset in PLP.}
    \label{tab:num_parts}
    \begin{tabular}{lccccccc}
        \toprule
        & Aircraft & Stanford Cars & CUB & CIFAR-100 & ImageNet-100 & TinyImageNet & Caltech-101 \\
        \midrule
        Number of Parts & 10 & 9 & 8 & 6 & 7 & 6 & 7 \\
        \bottomrule
    \end{tabular}
\end{table}
\clearpage
\section{Breakdown CCD Benchmark Results}
\label{supp: complete}
\let\labelitemi\labelitemii
We provide the breakdown results which include the \textit{continual ACC} (\textit{cACC}) (\textit{`All'}, \textit{`Old'}, \textit{`New'}) for each stage following the data splits in~\cite{zhang2022grow}. 
\begin{enumerate}
    \item Comparison with known class numbers:
    \begin{itemize}
        \item Table~\ref{tab:main_complete_result}, comparison on generic datasets with DINO.
        \item  Table~\ref{tab:main_complete_result_w_dinov2}, comparison on generic datasets with DINOv2.
        \item Table~\ref{tab:main_complete_result_ssb}, comparison on fine-grained datasets with DINO.
        \item Table~\ref{tab:main_complete_result_ssb_w_dinov2}, comparison on fine-grained datasets with DINOv2.
        \item Table~$\{$\ref{tab:main_complete_result_baselines},\ref{tab:main_complete_result_baselines_seed_7},\ref{tab:main_complete_result_baselines_seed_10},\ref{tab:main_complete_result_baselines_seed_2000},\ref{tab:main_complete_result_baselines_seed_2024}$\}$, multiple runs (\textit{5 seeds}) results on variants of PromptCCD with different prompt pool designs on generic and CUB datasets with DINO.
            \item Table~$\{$\ref{tab:main_complete_result_baselines_seed_1_dinov2},\ref{tab:main_complete_result_baselines_seed_7_dinov2},\ref{tab:main_complete_result_baselines_seed_10_dinov2},\ref{tab:main_complete_result_baselines_seed_2000_dinov2},\ref{tab:main_complete_result_baselines_seed_2024_dinov2}$\}$, multiple runs (\textit{5 seeds}) results on PromptCCD and PromptCCD++ on generic and CUB datasets with DINOv2.
    \end{itemize}
    \item Comparison with unknown class numbers, using DINO:
    \begin{itemize}
        \item Table~\ref{tab:main_result_with_unknown_k_gpc}, comparison using our GPC-based-estimator \cite{zhao2023learning}.
        \item Table~\ref{tab:main_result_with_unknown_k}, comparison using the $k$-means-based estimator in \cite{vaze2022generalized}.
    \end{itemize}
\end{enumerate}
\vspace{-2mm}
\definecolor{Gray}{gray}{0.9}
\definecolor{PaleBlue}{rgb}{0.7529, 0.9137, 0.9372}
\definecolor{BeauBlue}{rgb}{0.7686, 0.8470, 0.9529}
\definecolor{Mauve}{rgb}{0.8 , 0.7098, 0.9843}
\definecolor{PaleViolet}{rgb}{0.8156, 0.6431, 1.0}
\definecolor{Salmon}{rgb}{1.0, 0.8980, 0.6920}
\definecolor{Pink}{rgb}{1.0, 0.6902, 0.7908}
\definecolor{Mint}{rgb}{0.6902, 1.0, 0.7451}
\definecolor{cyan}{rgb}{0.906, 0.969, 0.965}
\definecolor{SoftP}{rgb}{0.945, 0.933, 0.949}

\begin{table*}[!ht]
    \caption{Breakdown results of various methods for CCD leveraging pretrained DINO model on generic datasets with the \textit{known} $C$ in each unlabelled set.}
    \centering
    \resizebox{\columnwidth}{!}{%
    \centering
        \begin{tabular}{clccc @{\hskip 0.1in} ccc@{\hskip 0.1in}ccc @{\hskip 0.1in} ||ccc}
        \toprule
        \multicolumn{2}{c}{}
        & \multicolumn{3}{c}{Stage 1 \textit{ACC} (\%)} 
        & \multicolumn{3}{c}{Stage 2 \textit{ACC} (\%)} 
        & \multicolumn{3}{c}{Stage 3 \textit{ACC} (\%)} 
        & \multicolumn{3}{c}{Average \textit{ACC} (\%)} 
        \\ 
        \multicolumn{1}{l}{} &
        \multicolumn{1}{c}{Method} &
        \multicolumn{1}{c}{\textit{All}} & 
        \multicolumn{1}{c}{\textit{Old}} &
        \multicolumn{1}{c}{\textit{New}} &
        \multicolumn{1}{c}{\textit{All}} & 
        \multicolumn{1}{c}{\textit{Old}} & 
        \multicolumn{1}{c}{\textit{New}} &
        \multicolumn{1}{c}{\textit{All}} & 
        \multicolumn{1}{c}{\textit{Old}} & 
        \multicolumn{1}{c}{\textit{New}} &
        \multicolumn{1}{c}{\textit{All} \cellcolor{blue!7!white}} & 
        \multicolumn{1}{c}{\textit{Old}} & 
        \multicolumn{1}{c}{\textit{New}} 
        \\ 
        \cmidrule[0.1pt](r{0.80em}){1-2}%
        \cmidrule[0.1pt](r{0.80em}){3-5}%
        \cmidrule[0.1pt](r{0.80em}){6-8}%
        \cmidrule[0.1pt](r{0.80em}){9-11}%
        \cmidrule[0.1pt](r{0.80em}){12-14}%
        \rowcolor{gray!5!white}
        &ORCA \cite{cao22orca}
        & $62.59$ & $71.55$ & $56.31$ 
        & $63.05$ & $66.38$ & $62.42$ 
        & $57.09$ & $61.90$ & $56.25$
        & $60.91$ \cellcolor{blue!7!white} & $66.61$ & $58.33$ \\
        \rowcolor{gray!5!white}
        &GCD \cite{vaze2022generalized} 
        & $67.65$ & $83.59$ & $56.49$ 
        & $52.89$ & $68.38$ & $49.93$ 
        & $53.99$ & $64.86$ & $52.08$
        & $58.18$ \cellcolor{blue!7!white} & $72.27$ & $52.83$ \\
        \rowcolor{gray!5!white}
        &SimGCD \cite{wen2022simple} 
        & $35.04$ & $50.65$ & $24.11$ 
        & $22.41$ & $39.05$ & $19.24$ 
        & $19.23$ & $26.57$ & $17.95$
        & $25.56$ \cellcolor{blue!7!white} & $38.76$ & $20.43$ \\
        \rowcolor{gray!5!white}
        &GCD $w/$replay
        & $55.68$ & $80.12$ & $38.57$ 
        & $45.16$ & $67.62$ & $40.87$ 
        & $48.96$ & $71.71$ & $44.98$
        & $49.93$ \cellcolor{blue!7!white} & $73.15$ & $41.47$ \\
        \rowcolor{gray!5!white}
        &SimGCD $w/$replay
        & $48.84$ & $74.16$ & $31.11$ 
        & $35.28$ & $61.43$ & $30.29$ 
        & $36.28$ & $64.57$ & $31.33$
        & $40.13$ \cellcolor{blue!7!white} & $66.72$ & $30.91$ \\
        \rowcolor{gray!5!white}
        &Grow \& Merge \cite{zhang2022grow} 
        & $64.77$ & $70.49$ & $60.77$ 
        & $58.32$ & $62.95$ & $57.44$ 
        & $49.21$ & $57.62$ & $47.73$
        & $57.43$ \cellcolor{blue!7!white} & $63.68$ & $55.31$ \\
        \rowcolor{gray!5!white}
        &MetaGCD \cite{wu2023metagcd} 
        & $56.20$ & $79.59$ & $39.83$ 
        & $56.63$ & $65.81$ & $55.05$ 
        & $53.65$ & $62.76$ & $52.05$
        & $55.49$ \cellcolor{blue!7!white} & $69.38$ & $48.98$ \\
        \rowcolor{gray!5!white}
        &PA-CGCD \cite{kim2023proxy} 
        & $57.43$ & $80.29$ & $41.43$ 
        & $61.69$ & $92.38$ & $55.84$ 
        & $55.63$ & $88.67$ & $49.85$
        & $58.25$ \cellcolor{blue!7!white} & $87.11$ & $49.04$ \\
        \rowcolor{gray!5!white}
        &Happy~\cite{ma2024happy}  
        & $71.70$ & $87.49$ & $59.54$ 
        & $60.22$ & $75.90$ & $57.89$  
        & $60.88$ & $69.24$ & $59.72$ 
        & $\textbf{64.27}$ \cellcolor{blue!7!white} & $77.54$ & $59.05$   \\
        \rowcolor{blue!3!white}
        \rot{\rlap{CIFAR100}}
        &PromptCCD (Ours) 
        & $70.69$ & $80.90$ & $63.54$ 
        & $64.08$ & $73.14$ & $62.35$ 
        & $57.73$ & $72.67$ & $55.12$
        & $64.17$ \cellcolor{blue!7!white} & $75.57$ & $60.34$ \\ \midrule
        \rowcolor{gray!5!white}
        &ORCA \cite{cao22orca}
        & $47.63$ & $69.84$ & $32.09$ 
        & $34.46$ & $38.95$ & $33.60$ 
        & $38.77$ & $28.76$ & $40.52$
        & $40.29$ \cellcolor{blue!7!white} & $45.85$ & $35.40$ \\
        \rowcolor{gray!5!white}
        &GCD \cite{vaze2022generalized} 
        & $75.65$ & $84.69$ & $69.31$ 
        & $71.21$ & $80.67$ & $69.40$ 
        & $61.38$ & $79.33$ & $58.23$
        & $69.41$ \cellcolor{blue!7!white} & $81.56$ & $65.65$ \\
        \rowcolor{gray!5!white}
        &SimGCD \cite{wen2022simple} 
        & $36.96$ & $51.43$ & $28.29$ 
        & $32.46$ & $40.71$ & $30.36$ 
        & $24.73$ & $29.29$ & $23.67$
        & $31.38$ \cellcolor{blue!7!white} & $40.47$ & $27.44$ \\
        \rowcolor{gray!5!white}
        &GCD $w/$replay
        & $79.46$ & $84.78$ & $75.74$ 
        & $72.64$ & $81.81$ & $70.89$ 
        & $64.01$ & $84.67$ & $60.40$
        & $72.04$ \cellcolor{blue!7!white} & $83.75$ & $69.01$ \\
        \rowcolor{gray!5!white}
        &SimGCD $w/$replay
        & $48.76$ & $77.14$ & $31.71$ 
        & $48.55$ & $67.14$ & $43.82$ 
        & $45.27$ & $59.29$ & $42.00$
        & $47.53$ \cellcolor{blue!7!white} & $67.86$ & $39.18$ \\
        \rowcolor{gray!5!white}
        &Grow \& Merge \cite{zhang2022grow} 
        & $75.34$ & $76.78$ & $74.34$ 
        & $63.76$ & $73.67$ & $62.87$ 
        & $64.43$ & $74.86$ & $62.60$
        & $67.84$ \cellcolor{blue!7!white} & $75.10$ & $66.60$ \\
        \rowcolor{gray!5!white}
        &MetaGCD \cite{wu2023metagcd} 
        & $65.61$ & $83.92$ & $52.80$ 
        & $67.36$ & $83.14$ & $64.35$ 
        & $66.26$ & $74.57$ & $64.80$
        & $66.41$ \cellcolor{blue!7!white} & $80.54$ & $60.65$ \\
        \rowcolor{gray!5!white}
        &PA-CGCD \cite{kim2023proxy} 
        & $70.05$ & $82.61$ & $61.26$ 
        & $68.34$ & $96.57$ & $62.95$ 
        & $55.99$ & $94.29$ & $49.28$
        & $64.79$ \cellcolor{blue!7!white} & $91.15$ & $57.83$ \\
        \rowcolor{gray!5!white}
        &Happy~\cite{ma2024happy} 
        & $80.40$ & $86.08$ & $76.29$ 
        & $71.82$ & $81.24$ & $70.02$  
        & $72.11$ & $80.30$ & $70.83$ 
        & $74.78$ \cellcolor{blue!7!white} & $82.54$ & $72.38$   \\
        \rowcolor{blue!3!white}
        \rot{\rlap{ImageNet-100}}
        &PromptCCD (Ours) 
        & $79.56$ & $84.24$ & $76.29$ 
        & $78.58$ & $79.71$ & $78.36$ 
        & $70.33$ & $81.33$ & $68.40$
        & $\textbf{76.16}$ \cellcolor{blue!7!white} & $81.76$ & $74.35$ \\ \midrule
        \rowcolor{gray!5!white}
        &ORCA \cite{cao22orca} 
        & $62.64$ & $68.63$ & $58.44$ 
        & $50.76$ & $61.38$ & $48.25$ 
        & $50.72$ & $59.38$ & $49.11$
        & $54.71$ \cellcolor{blue!7!white} & $63.13$ & $51.93$ \\
        \rowcolor{gray!5!white}
        &GCD \cite{vaze2022generalized} 
        & $63.62$ & $73.14$ & $56.96$ 
        & $51.08$ & $64.19$ & $48.58$ 
        & $50.91$ & $60.29$ & $49.28$
        & $55.20$ \cellcolor{blue!7!white} & $65.87$ & $51.61$ \\
        \rowcolor{gray!5!white}
        &SimGCD \cite{wen2022simple} 
        & $37.96$ & $34.76$ & $40.20$ 
        & $32.18$ & $26.62$ & $33.24$ 
        & $30.05$ & $25.95$ & $30.77$
        & $33.40$ \cellcolor{blue!7!white} & $29.11$ & $34.74$ \\
        \rowcolor{gray!5!white}
        &GCD $w/$replay
        & $66.40$ & $76.06$ & $59.64$ 
        & $50.28$ & $66.38$ & $47.21$ 
        & $52.32$ & $60.19$ & $50.94$
        & $56.33$ \cellcolor{blue!7!white} & $67.54$ & $52.60$ \\
        \rowcolor{gray!5!white}
        &SimGCD $w/$replay
        & $46.04$ & $65.22$ & $32.61$ 
        & $34.21$ & $56.00$ & $30.05$ 
        & $32.11$ & $53.24$ & $28.42$
        & $37.45$ \cellcolor{blue!7!white} & $58.15$ & $30.36$ \\
        \rowcolor{gray!5!white}
        &Grow \& Merge \cite{zhang2022grow} 
        & $59.52$ & $64.24$ & $56.21$ 
        & $51.50$ & $58.19$ & $50.23$ 
        & $45.39$ & $56.62$ & $43.43$
        & $52.14$ \cellcolor{blue!7!white} & $59.68$ & $49.96$ \\
        \rowcolor{gray!5!white}
        &MetaGCD \cite{wu2023metagcd} 
        & $59.41$ & $73.90$ & $49.27$ 
        & $57.21$ & $63.71$ & $55.96$ 
        & $49.17$ & $60.76$ & $47.14$
        & $55.26$ \cellcolor{blue!7!white} & $66.12$ & $50.79$ \\
        \rowcolor{gray!5!white}
        &PA-CGCD \cite{kim2023proxy} 
        & $56.01$ & $74.96$ & $42.74$ 
        & $41.89$ & $65.38$ & $37.40$ 
        & $55.50$ & $84.52$ & $50.42$
        & $51.13$ \cellcolor{blue!7!white} & $74.95$ & $43.52$ \\
        \rowcolor{gray!5!white}
        &Happy~\cite{ma2024happy} 
        & $64.48$ & $72.24$ & $59.04$ 
        & $57.14$ & $65.43$ & $55.55$  
        & $54.29$ & $59.48$ & $53.38$ 
        & $58.63$ \cellcolor{blue!7!white} & $65.71$ & $55.99$   \\
        \rowcolor{blue!3!white}
        \rot{\rlap{TinyImageNet}}
        &PromptCCD (Ours) 
        & $68.67$ & $72.84$ & $65.76$ 
        & $59.69$ & $65.67$ & $58.55$ 
        & $57.16$ & $61.10$ & $56.47$
        & $\textbf{61.84}$ \cellcolor{blue!7!white} & $66.54$ & $60.26$ \\ \midrule
        \rowcolor{gray!5!white}
        &ORCA \cite{cao22orca} 
        & $80.79$ & $80.08$ & $75.52$ 
        & $77.34$ & $85.96$ & $74.27$ 
        & $72.18$ & $82.38$ & $69.82$
        & $76.77$ \cellcolor{blue!7!white} & $82.80$ & $73.20$ \\
        \rowcolor{gray!5!white}
        &GCD \cite{vaze2022generalized} 
        & $82.17$ & $93.94$ & $70.49$ 
        & $74.73$ & $85.38$ & $70.95$ 
        & $77.91$ & $80.48$ & $77.31$
        & $78.27$ \cellcolor{blue!7!white} & $86.60$ & $72.92$ \\
        \rowcolor{gray!5!white}
        &SimGCD \cite{wen2022simple} 
        & $34.57$ & $38.29$ & $30.87$ 
        & $34.00$ & $38.60$ & $32.37$ 
        & $32.38$ & $35.71$ & $31.61$
        & $33.65$ \cellcolor{blue!7!white} & $37.53$ & $31.62$ \\
        \rowcolor{gray!5!white}
        &GCD $w/$replay
        & $87.11$ & $92.56$ & $81.69$ 
        & $75.50$ & $85.38$ & $71.99$ 
        & $66.91$ & $80.48$ & $63.77$
        & $76.51$ \cellcolor{blue!7!white} & $86.14$ & $72.48$ \\
        \rowcolor{gray!5!white}
        &SimGCD $w/$replay
        & $59.95$ & $62.26$ & $57.65$ 
        & $41.50$ & $42.11$ & $41.29$ 
        & $46.69$ & $53.81$ & $45.04$
        & $49.38$ \cellcolor{blue!7!white} & $52.72$ & $47.99$ \\
        \rowcolor{gray!5!white}
        &Grow \& Merge \cite{zhang2022grow} 
        & $80.80$ & $88.15$ & $73.50$ 
        & $75.96$ & $87.13$ & $71.99$ 
        & $70.48$ & $75.71$ & $69.27$
        & $75.75$ \cellcolor{blue!7!white} & $83.66$ & $71.59$ \\
        \rowcolor{gray!5!white}
        &MetaGCD \cite{wu2023metagcd} 
        & $86.15$ & $95.59$ & $76.78$ 
        & $75.96$ & $90.06$ & $70.95$ 
        & $80.14$ & $81.43$ & $79.85$
        & $80.75$ \cellcolor{blue!7!white} & $89.02$ & $75.86$ \\
        \rowcolor{gray!5!white}
        &PA-CGCD \cite{kim2023proxy} 
        & $78.33$ & $92.56$ & $64.21$ 
        & $79.33$ & $98.83$ & $72.41$ 
        & $76.21$ & $92.86$ & $72.36$
        & $77.96$ \cellcolor{blue!7!white} & $94.75$ & $69.66$ \\
        \rowcolor{gray!5!white}
        &Happy~\cite{ma2024happy} 
        & $91.84$ & $94.21$ & $85.76$ 
        & $79.59$ & $88.24$ & $75.44$  
        & $71.39$ & $87.50$ & $69.88$ 
        & $80.94$ \cellcolor{blue!7!white} & $89.98$ & $77.03$   \\
        \rowcolor{blue!3!white}
        \rot{\rlap{Caltech-101}}
        &PromptCCD (Ours) 
        & $89.57$ & $92.84$ & $86.34$ 
        & $78.87$ & $87.72$ & $75.73$ 
        & $78.89$ & $86.67$ & $77.09$
        & $\textbf{82.44}$ \cellcolor{blue!7!white} & $89.08$ & $79.72$ \\
        \bottomrule
        \end{tabular}
    }
    \label{tab:main_complete_result}
\end{table*}
\definecolor{Gray}{gray}{0.9}
\definecolor{PaleBlue}{rgb}{0.7529, 0.9137, 0.9372}
\definecolor{BeauBlue}{rgb}{0.7686, 0.8470, 0.9529}
\definecolor{Mauve}{rgb}{0.8 , 0.7098, 0.9843}
\definecolor{PaleViolet}{rgb}{0.8156, 0.6431, 1.0}
\definecolor{Salmon}{rgb}{1.0, 0.8980, 0.6920}
\definecolor{Pink}{rgb}{1.0, 0.6902, 0.7908}
\definecolor{Mint}{rgb}{0.6902, 1.0, 0.7451}
\definecolor{cyan}{rgb}{0.906, 0.969, 0.965}
\definecolor{SoftP}{rgb}{0.945, 0.933, 0.949}

\begin{table*}[!ht]
    \caption{Breakdown results of different methods for CCD leveraging pretrained DINOv2 model on generic datasets with the \textit{known} $C$ in each unlabelled set.}
    \centering
    \resizebox{\columnwidth}{!}{%
    \centering
        \begin{tabular}{clccc @{\hskip 0.1in} ccc@{\hskip 0.1in}ccc @{\hskip 0.1in} ||ccc}
        \toprule
        \multicolumn{2}{c}{}
        & \multicolumn{3}{c}{Stage 1 \textit{ACC} (\%)} 
        & \multicolumn{3}{c}{Stage 2 \textit{ACC} (\%)} 
        & \multicolumn{3}{c}{Stage 3 \textit{ACC} (\%)} 
        & \multicolumn{3}{c}{Average \textit{ACC} (\%)} 
        \\ 
        \multicolumn{1}{l}{} &
        \multicolumn{1}{c}{Method} &
        \multicolumn{1}{c}{\textit{All}} & 
        \multicolumn{1}{c}{\textit{Old}} &
        \multicolumn{1}{c}{\textit{New}} &
        \multicolumn{1}{c}{\textit{All}} & 
        \multicolumn{1}{c}{\textit{Old}} & 
        \multicolumn{1}{c}{\textit{New}} &
        \multicolumn{1}{c}{\textit{All}} & 
        \multicolumn{1}{c}{\textit{Old}} & 
        \multicolumn{1}{c}{\textit{New}} &
        \multicolumn{1}{c}{\textit{All} \cellcolor{blue!7!white}} & 
        \multicolumn{1}{c}{\textit{Old}} & 
        \multicolumn{1}{c}{\textit{New}} 
        \\ 
        \cmidrule[0.1pt](r{0.80em}){1-2}%
        \cmidrule[0.1pt](r{0.80em}){3-5}%
        \cmidrule[0.1pt](r{0.80em}){6-8}%
        \cmidrule[0.1pt](r{0.80em}){9-11}%
        \cmidrule[0.1pt](r{0.80em}){12-14}%
        \rowcolor{gray!5!white}
        &GCD \cite{vaze2022generalized} 
        & $74.18$ & $90.16$ & $63.00$ 
        & $63.89$ & $73.52$ & $62.05$ 
        & $57.99$ & $67.52$ & $56.32$ 
        & $65.35$ \cellcolor{blue!7!white} & $77.06$ & $60.46$  \\
        \rowcolor{gray!5!white}
        &MetaGCD \cite{wu2023metagcd} 
        & $62.03$ & $84.94$ & $46.00$ 
        & $51.34$ & $78.95$ & $46.07$ 
        & $42.94$ & $75.05$ & $37.32$ 
        & $52.10$ \cellcolor{blue!7!white} & $79.64$ & $43.13$  \\
        \rowcolor{gray!5!white}
        &PA-CGCD \cite{kim2023proxy} 
        & $59.75$ & $84.04$ & $42.74$ 
        & $47.60$ & $69.33$ & $43.45$ 
        & $55.73$ & $84.19$ & $50.75$ 
        & $54.36$ \cellcolor{blue!7!white} & $79.19$ & $45.65$  \\
        \rowcolor{gray!5!white}
        &Happy~\cite{ma2024happy} 
        & $80.86$ & $88.33$ & $75.63$ 
        & $72.00$ & $84.29$ & $69.39$  
        & $66.52$ & $83.43$ & $63.57$ 
        & $73.13$ \cellcolor{blue!7!white} & $85.35$ & $69.53$   \\
        \rowcolor{blue!3!white}
        &PromptCCD (Ours) 
        & $78.24$ & $90.04$ & $69.97$ 
        & $65.27$ & $74.67$ & $63.46$ 
        & $65.69$ & $69.33$ & $65.05$ 
        & $69.73$ \cellcolor{blue!7!white} & $78.01$ & $66.16$  \\
        \rowcolor{blue!3!white}
        \rot{\rlap{C100}}
        &PromptCCD++ (Ours) 
        & $82.59$ & $87.27$ & $79.31$ 
        & $81.30$ & $82.95$ & $80.98$  
        & $69.15$ & $79.33$ & $67.37$ 
        & $\textbf{77.68}$ \cellcolor{blue!7!white} & $83.18$ & $75.89$   \\ \midrule
        \rowcolor{gray!5!white}
        &GCD \cite{vaze2022generalized} 
        & $78.72$ & $86.49$ & $73.29$ 
        & $73.92$ & $81.90$ & $72.02$ 
        & $62.09$ & $80.67$ & $58.83$ 
        & $71.58$ \cellcolor{blue!7!white} & $83.02$ & $68.05$  \\
        \rowcolor{gray!5!white}
        &MetaGCD \cite{wu2023metagcd} 
        & $71.56$ & $86.49$ & $61.11$ 
        & $70.14$ & $80.14$ & $68.75$ 
        & $68.90$ & $81.24$ & $64.12$ 
        & $70.20$ \cellcolor{blue!7!white} & $82.62$ & $64.66$  \\
        \rowcolor{gray!5!white}
        &PA-CGCD \cite{kim2023proxy} 
        & $79.83$ & $87.92$ & $74.17$ 
        & $67.53$ & $86.29$ & $63.95$ 
        & $77.11$ & $90.38$ & $77.93$ 
        & $74.82$ \cellcolor{blue!7!white} & $88.20$ & $72.02$  \\
        \rowcolor{gray!5!white}
        &Happy~\cite{ma2024happy} 
        & $79.98$ & $87.43$ & $74.77$ 
        & $81.97$ & $87.05$ & $81.00$  
        & $79.12$ & $87.62$ & $77.63$ 
        & $80.36$ \cellcolor{blue!7!white} & $87.36$ & $77.80$   \\
        \rowcolor{blue!3!white}
        &PromptCCD (Ours) 
        & $80.35$ & $86.78$ & $75.86$ 
        & $75.65$ & $80.86$ & $74.65$ 
        & $72.85$ & $80.19$ & $73.08$ 
        & $76.28$ \cellcolor{blue!7!white} & $82.61$ & $74.53$ \\ 
        \rowcolor{blue!3!white}
        \rot{\rlap{IN-100}}
        &PromptCCD++ (Ours) 
        & $81.09$ & $86.33$ & $77.43$ 
        & $82.63$ & $84.95$ & $82.18$  
        & $86.45$ & $83.90$ & $86.90$ 
        & $\textbf{83.39}$ \cellcolor{blue!7!white} & $85.06$ & $82.17$   \\ \midrule
        \rowcolor{gray!5!white}
        &GCD \cite{vaze2022generalized} 
        & $69.68$ & $82.90$ & $60.43$ 
        & $56.57$ & $77.81$ & $52.52$ 
        & $50.89$ & $71.62$ & $47.27$ 
        & $59.05$ \cellcolor{blue!7!white} & $77.44$ & $53.41$  \\
        \rowcolor{gray!5!white}
        &MetaGCD \cite{wu2023metagcd} 
        & $62.07$ & $81.69$ & $48.33$ 
        & $53.29$ & $73.29$ & $49.47$ 
        & $53.10$ & $69.10$ & $50.30$ 
        & $56.15$ \cellcolor{blue!7!white} & $74.69$ & $49.37$  \\
        \rowcolor{gray!5!white}
        &PA-CGCD \cite{kim2023proxy} 
        & $56.94$ & $73.41$ & $45.41$ 
        & $47.30$ & $66.33$ & $43.66$ 
        & $52.07$ & $64.48$ & $49.90$ 
        & $52.10$ \cellcolor{blue!7!white} & $68.07$ & $46.32$  \\
        \rowcolor{gray!5!white}
        &Happy~\cite{ma2024happy} 
        & $72.17$ & $82.02$ & $65.27$ 
        & $64.16$ & $76.57$ & $61.79$  
        & $65.51$ & $77.81$ & $63.36$ 
        & $67.28$ \cellcolor{blue!7!white} & $78.80$ & $63.47$   \\
        \rowcolor{blue!3!white}
        &PromptCCD (Ours) 
        & $74.30$ & $83.69$ & $67.73$ 
        & $67.00$ & $75.86$ & $65.31$ 
        & $63.31$ & $67.14$ & $62.64$ 
        & $68.20$ \cellcolor{blue!7!white} & $75.56$ & $65.23$  \\
        \rowcolor{blue!3!white}
        \rot{\rlap{Tiny}}
        &PromptCCD++ (Ours) 
        & $74.97$ & $81.90$ & $70.13$ 
        & $70.69$ & $78.76$ & $69.15$  
        & $73.41$ & $79.48$ & $72.35$ 
        & $\textbf{73.02}$ \cellcolor{blue!7!white} & $80.15$ & $70.54$   \\ \midrule
        \rowcolor{gray!5!white}
        &GCD \cite{vaze2022generalized} 
        & $89.99$ & $95.87$ & $84.15$ 
        & $79.63$ & $91.81$ & $75.31$ 
        & $79.38$ & $78.27$ & $79.93$ 
        & $83.00$ \cellcolor{blue!7!white} & $88.65$ & $79.80$  \\
        \rowcolor{gray!5!white}
        &MetaGCD \cite{wu2023metagcd} 
        & $89.03$ & $91.18$ & $86.89$ 
        & $77.95$ & $88.89$ & $74.07$ 
        & $82.17$ & $84.19$ & $81.70$ 
        & $83.05$ \cellcolor{blue!7!white} & $88.08$ & $80.89$  \\
        \rowcolor{gray!5!white}
        &PA-CGCD \cite{kim2023proxy} 
        & $80.52$ & $91.18$ & $69.95$ 
        & $88.97$ & $95.32$ & $86.72$ 
        & $79.70$ & $95.71$ & $75.99$ 
        & $83.06$ \cellcolor{blue!7!white} & $94.07$ & $77.55$  \\
        \rowcolor{gray!5!white}
        &Happy~\cite{ma2024happy} 
        & $95.84$ & $97.21$ & $94.76$ 
        & $84.59$ & $88.24$ & $84.44$  
        & $84.39$ & $87.50$ & $83.88$ 
        & $88.27$ \cellcolor{blue!7!white} & $90.98$ & $87.69$   \\
        \rowcolor{blue!3!white}
        &PromptCCD (Ours) 
        & $90.53$ & $94.21$ & $86.89$ 
        & $76.26$ & $90.06$ & $71.37$ 
        & $84.79$ & $79.52$ & $86.01$ 
        & $83.86$ \cellcolor{blue!7!white} & $87.93$ & $81.42$  \\
        \rowcolor{blue!3!white}
         \rot{\rlap{C-101}\hskip 0.1in}
        &PromptCCD++ (Ours) 
        & $95.84$ & $93.87$ & $97.38$ 
        & $87.96$ & $95.88$ & $86.04$  
        & $91.75$ & $91.83$ & $91.73$ 
        & $\textbf{91.85}$ \cellcolor{blue!7!white} & $93.86$ & $91.72$   \\ 
        \bottomrule
        \end{tabular}
    }
    \label{tab:main_complete_result_w_dinov2}
\end{table*}

\definecolor{Gray}{gray}{0.9}
\definecolor{PaleBlue}{rgb}{0.7529, 0.9137, 0.9372}
\definecolor{BeauBlue}{rgb}{0.7686, 0.8470, 0.9529}
\definecolor{Mauve}{rgb}{0.8 , 0.7098, 0.9843}
\definecolor{PaleViolet}{rgb}{0.8156, 0.6431, 1.0}
\definecolor{Salmon}{rgb}{1.0, 0.8980, 0.6920}
\definecolor{Pink}{rgb}{1.0, 0.6902, 0.7908}
\definecolor{Mint}{rgb}{0.6902, 1.0, 0.7451}
\definecolor{cyan}{rgb}{0.906, 0.969, 0.965}
\definecolor{SoftP}{rgb}{0.945, 0.933, 0.949}

\begin{table*}[!ht]
    \caption{Breakdown results of different methods for CCD leveraging pretrained DINO model on fine-grained datasets with the \textit{known} $C$ in each unlabelled set.}
    \centering
    \resizebox{\columnwidth}{!}{%
    \centering
        \begin{tabular}{clccc @{\hskip 0.1in} ccc@{\hskip 0.1in}ccc @{\hskip 0.1in} ||ccc}
        \toprule
        \multicolumn{2}{c}{}
        & \multicolumn{3}{c}{Stage 1 \textit{ACC} (\%)} 
        & \multicolumn{3}{c}{Stage 2 \textit{ACC} (\%)} 
        & \multicolumn{3}{c}{Stage 3 \textit{ACC} (\%)} 
        & \multicolumn{3}{c}{Average \textit{ACC} (\%)} 
        \\ 
        \multicolumn{1}{l}{} &
        \multicolumn{1}{c}{Method} &
        \multicolumn{1}{c}{\textit{All}} & 
        \multicolumn{1}{c}{\textit{Old}} &
        \multicolumn{1}{c}{\textit{New}} &
        \multicolumn{1}{c}{\textit{All}} & 
        \multicolumn{1}{c}{\textit{Old}} & 
        \multicolumn{1}{c}{\textit{New}} &
        \multicolumn{1}{c}{\textit{All}} & 
        \multicolumn{1}{c}{\textit{Old}} & 
        \multicolumn{1}{c}{\textit{New}} &
        \multicolumn{1}{c}{\textit{All} \cellcolor{blue!7!white}} & 
        \multicolumn{1}{c}{\textit{Old}} & 
        \multicolumn{1}{c}{\textit{New}} 
        \\ 
        \cmidrule[0.1pt](r{0.80em}){1-2}%
        \cmidrule[0.1pt](r{0.80em}){3-5}%
        \cmidrule[0.1pt](r{0.80em}){6-8}%
        \cmidrule[0.1pt](r{0.80em}){9-11}%
        \cmidrule[0.1pt](r{0.80em}){12-14}%
        \rowcolor{gray!5!white}
        &ORCA \cite{cao22orca} 
        & $31.89$ & $31.43$ & $32.17$ 
        & $28.64$ & $27.14$ & $28.93$
        & $31.77$ & $18.57$ & $36.23$
        & $30.77$ \cellcolor{blue!7!white} & $25.71$ & $32.44$ \\
        \rowcolor{gray!5!white}
        &GCD \cite{vaze2022generalized} 
        & $47.30$ & $58.57$ & $40.43$ 
        & $48.59$ & $68.57$ & $44.66$
        & $46.21$ & $57.41$ & $42.51$
        & $47.37$ \cellcolor{blue!7!white} & $61.43$ & $42.53$ \\
        \rowcolor{gray!5!white}
        &SimGCD \cite{wen2022simple} 
        & $32.97$ & $47.86$ & $23.91$ 
        & $30.28$ & $34.29$ & $29.49$
        & $23.83$ & $25.00$ & $23.43$
        & $29.03$ \cellcolor{blue!7!white} & $35.72$ & $25.61$ \\
        \rowcolor{gray!5!white}
        &GCD $w/$replay
        & $47.03$ & $60.71$ & $38.70$ 
        & $46.71$ & $71.43$ & $41.85$
        & $43.14$ & $55.00$ & $39.13$
        & $45.63$ \cellcolor{blue!7!white} & $62.38$ & $39.89$ \\
        \rowcolor{gray!5!white}
        &SimGCD $w/$replay
        & $41.08$ & $62.86$ & $27.83$ 
        & $33.33$ & $67.14$ & $26.69$
        & $37.91$ & $54.29$ & $32.37$
        & $37.44$ \cellcolor{blue!7!white} & $61.43$ & $28.96$ \\
        \rowcolor{gray!5!white}
        &Grow \& Merge \cite{zhang2022grow} 
        & $32.43$ & $32.86$ & $32.17$ 
        & $30.05$ & $41.43$ & $27.81$
        & $30.69$ & $25.71$ & $32.37$
        & $31.06$ \cellcolor{blue!7!white} & $33.33$ & $30.78$ \\
        \rowcolor{gray!5!white}
        &MetaGCD \cite{wu2023metagcd} 
        & $47.57$ & $61.43$ & $39.13$ 
        & $45.54$ & $64.29$ & $41.85$
        & $40.79$ & $51.43$ & $37.20$
        & $44.63$ \cellcolor{blue!7!white} & $59.05$ & $39.39$ \\
        \rowcolor{gray!5!white}
        &PA-CGCD \cite{kim2023proxy} 
        & $48.11$ & $61.43$ & $40.00$ 
        & $45.54$ & $87.14$ & $37.36$
        & $51.08$ & $70.71$ & $44.44$
        & $48.24$ \cellcolor{blue!7!white} & $73.09$ & $40.60$ \\
        \rowcolor{gray!5!white}
        &Happy~\cite{ma2024happy}  
        & $50.00$ & $62.86$ & $34.48$ 
        & $45.32$ & $62.86$ & $44.08$  
        & $41.99$ & $51.43$ & $41.44$ 
        & $45.77$ \cellcolor{blue!7!white} & $59.05$ & $40.00$   \\
        \rowcolor{blue!3!white}
        \rot{\rlap{Aircraft}}
        &PromptCCD (Ours) 
        & $57.30$ & $63.57$ & $53.48$ 
        & $47.18$ & $64.29$ & $43.82$
        & $53.43$ & $53.57$ & $53.28$
        & $\textbf{52.64}$ \cellcolor{blue!7!white} & $60.48$ & $50.23$ \\ \midrule
        \rowcolor{gray!5!white}
        &ORCA \cite{cao22orca} 
        & $22.24$ & $30.35$ & $17.35$ 
        & $21.50$ & $43.61$ & $18.56$
        & $18.64$ & $26.25$ & $16.89$
        & $20.79$ \cellcolor{blue!7!white} & $33.40$ & $17.60$ \\
        \rowcolor{gray!5!white}
        &GCD \cite{vaze2022generalized} 
        & $43.78$ & $60.70$ & $33.55$ 
        & $38.58$ & $61.65$ & $35.51$
        & $35.26$ & $52.51$ & $31.29$
        & $39.21$ \cellcolor{blue!7!white} & $58.29$ & $33.45$ \\
        \rowcolor{gray!5!white}
        &SimGCD \cite{wen2022simple} 
        & $23.47$ & $31.17$ & $18.82$ 
        & $21.06$ & $52.63$ & $16.85$
        & $18.50$ & $39.00$ & $13.78$
        & $21.01$ \cellcolor{blue!7!white} & $40.93$ & $16.48$ \\
        \rowcolor{gray!5!white}
        &GCD $w/$replay
        & $43.67$ & $62.60$ & $32.24$ 
        & $40.18$ & $58.65$ & $37.71$
        & $35.77$ & $53.28$ & $31.73$
        & $39.87$ \cellcolor{blue!7!white} & $58.18$ & $33.89$ \\
        \rowcolor{gray!5!white}
        &SimGCD $w/$replay
        & $35.82$ & $36.86$ & $19.15$ 
        & $22.30$ & $63.16$ & $16.85$
        & $20.16$ & $47.10$ & $13.96$
        & $22.76$ \cellcolor{blue!7!white} & $49.04$ & $16.65$ \\
        \rowcolor{gray!5!white}
        &Grow \& Merge \cite{zhang2022grow} 
        & $24.18$ & $34.42$ & $18.00$ 
        & $22.74$ & $42.11$ & $20.16$
        & $18.79$ & $29.34$ & $16.36$
        & $21.90$ \cellcolor{blue!7!white} & $35.29$ & $18.17$ \\
        \rowcolor{gray!5!white}
        &MetaGCD \cite{wu2023metagcd} 
        & $39.80$ & $56.37$ & $29.79$ 
        & $34.25$ & $61.65$ & $30.59$
        & $33.89$ & $52.90$ & $29.51$
        & $35.98$ \cellcolor{blue!7!white} & $56.97$ & $29.96$ \\
        \rowcolor{gray!5!white}
        &PA-CGCD \cite{kim2023proxy} 
        & $49.31$ & $75.88$ & $33.86$ 
        & $39.62$ & $74.69$ & $35.11$
        & $42.70$ & $90.73$ & $31.64$
        & $43.88$ \cellcolor{blue!7!white} & $80.43$ & $33.54$ \\
        \rowcolor{gray!5!white}
        &Happy~\cite{ma2024happy} 
        & $48.33$ & $71.27$ & $32.26$ 
        & $47.55$ & $68.42$ & $40.64$  
        & $40.20$ & $67.18$ & $33.58$ 
        & $45.36$ \cellcolor{blue!7!white} & $68.96$ & $35.49$   \\
        \rowcolor{blue!3!white}
        \rot{\rlap{SCars}}
        &PromptCCD (Ours) 
        & $50.31$ & $71.82$ & $37.32$ 
        & $44.69$ & $62.41$ & $42.33$
        & $37.21$ & $64.86$ & $30.84$
        & $44.07$ \cellcolor{blue!7!white} & $66.36$ & $36.83$ \\ \midrule
        \rowcolor{gray!5!white}
        &ORCA \cite{cao22orca} 
        & $49.79$ & $66.43$ & $38.66$ 
        & $30.63$ & $63.57$ & $23.64$ 
        & $44.76$ & $68.57$ & $40.11$
        & $41.73$ \cellcolor{blue!7!white} & $66.19$ & $34.14$ \\
        \rowcolor{gray!5!white}
        &GCD \cite{vaze2022generalized} 
        & $58.80$ & $75.71$ & $47.49$ 
        & $49.25$ & $76.43$ & $43.48$ 
        & $56.88$ & $74.49$ & $53.48$
        & $54.98$ \cellcolor{blue!7!white} & $75.47$ & $48.15$ \\
        \rowcolor{gray!5!white}
        &SimGCD \cite{wen2022simple} 
        & $49.26$ & $63.81$ & $39.71$ 
        & $29.48$ & $49.64$ & $25.23$ 
        & $40.92$ & $64.29$ & $36.30$
        & $39.89$ \cellcolor{blue!7!white} & $59.25$ & $33.75$ \\
        \rowcolor{gray!5!white}
        &GCD $w/$replay
        & $59.66$ & $77.50$ & $47.73$ 
        & $50.12$ & $73.57$ & $45.15$ 
        & $54.20$ & $72.86$ & $50.56$
        & $54.66$ \cellcolor{blue!7!white} & $74.64$ & $47.81$ \\
        \rowcolor{gray!5!white}
        &SimGCD $w/$replay
        & $49.70$ & $75.75$ & $32.60$ 
        & $37.99$ & $69.34$ & $31.38$ 
        & $38.56$ & $72.86$ & $31.78$
        & $42.08$ \cellcolor{blue!7!white} & $72.65$ & $31.92$ \\
        \rowcolor{gray!5!white}
        &Grow \& Merge \cite{zhang2022grow} 
        & $44.21$ & $65.00$ & $30.31$ 
        & $29.50$ & $65.00$ & $21.97$ 
        & $42.89$ & $65.00$ & $38.58$
        & $38.87$ \cellcolor{blue!7!white} & $65.00$ & $30.29$ \\
        \rowcolor{gray!5!white}
        &MetaGCD \cite{wu2023metagcd} 
        & $50.93$ & $71.07$ & $37.47$ 
        & $39.37$ & $74.29$ & $31.97$ 
        & $43.47$ & $77.86$ & $36.77$
        & $44.59$ \cellcolor{blue!7!white} & $74.40$ & $35.40$ \\
        \rowcolor{gray!5!white}
        &PA-CGCD \cite{kim2023proxy} 
        & $55.94$ & $73.21$ & $44.39$ 
        & $46.25$ & $77.86$ & $39.55$ 
        & $55.24$ & $80.71$ & $50.28$
        & $52.48$ \cellcolor{blue!7!white} & $77.26$ & $44.74$ \\
        \rowcolor{gray!5!white}
        &Happy~\cite{ma2024happy} 
        & $58.28$ & $77.50$ & $46.44$ 
        & $50.90$ & $74.29$ & $45.72$  
        & $64.05$ & $74.49$ & $61.96$ 
        & $57.74$ \cellcolor{blue!7!white} & $75.42$ & $51.37$   \\
        \rowcolor{blue!3!white}
        \rot{\rlap{CUB}}
        &PromptCCD (Ours) 
        & $57.08$ & $75.00$ & $45.11$ 
        & $47.38$ & $75.00$ & $41.52$ 
        & $61.89$ & $76.43$ & $59.05$
        & $55.45$ \cellcolor{blue!7!white} & $75.48$ & $48.56$ \\
        \bottomrule
        \end{tabular}
    }
    \label{tab:main_complete_result_ssb}
\end{table*}
\definecolor{Gray}{gray}{0.9}
\definecolor{PaleBlue}{rgb}{0.7529, 0.9137, 0.9372}
\definecolor{BeauBlue}{rgb}{0.7686, 0.8470, 0.9529}
\definecolor{Mauve}{rgb}{0.8 , 0.7098, 0.9843}
\definecolor{PaleViolet}{rgb}{0.8156, 0.6431, 1.0}
\definecolor{Salmon}{rgb}{1.0, 0.8980, 0.6920}
\definecolor{Pink}{rgb}{1.0, 0.6902, 0.7908}
\definecolor{Mint}{rgb}{0.6902, 1.0, 0.7451}
\definecolor{cyan}{rgb}{0.906, 0.969, 0.965}
\definecolor{SoftP}{rgb}{0.945, 0.933, 0.949}

\begin{table*}[!ht]
    \caption{Breakdown results of different methods for CCD leveraging pretrained DINOv2 model on fine-grained datasets with the \textit{known} $C$ in each unlabelled set.}
    \centering
    \resizebox{1.0\columnwidth}{!}{%
    \centering
        \begin{tabular}{clccc @{\hskip 0.1in} ccc@{\hskip 0.1in}ccc @{\hskip 0.1in} ||ccc}
        \toprule
        \multicolumn{2}{c}{}
        & \multicolumn{3}{c}{Stage 1 \textit{ACC} (\%)} 
        & \multicolumn{3}{c}{Stage 2 \textit{ACC} (\%)} 
        & \multicolumn{3}{c}{Stage 3 \textit{ACC} (\%)} 
        & \multicolumn{3}{c}{Average \textit{ACC} (\%)} 
        \\ 
        \multicolumn{1}{l}{} &
        \multicolumn{1}{c}{Method} &
        \multicolumn{1}{c}{\textit{All}} & 
        \multicolumn{1}{c}{\textit{Old}} &
        \multicolumn{1}{c}{\textit{New}} &
        \multicolumn{1}{c}{\textit{All}} & 
        \multicolumn{1}{c}{\textit{Old}} & 
        \multicolumn{1}{c}{\textit{New}} &
        \multicolumn{1}{c}{\textit{All}} & 
        \multicolumn{1}{c}{\textit{Old}} & 
        \multicolumn{1}{c}{\textit{New}} &
        \multicolumn{1}{c}{\textit{All} \cellcolor{blue!7!white}} & 
        \multicolumn{1}{c}{\textit{Old}} & 
        \multicolumn{1}{c}{\textit{New}} 
        \\ 
        \cmidrule[0.1pt](r{0.80em}){1-2}%
        \cmidrule[0.1pt](r{0.80em}){3-5}%
        \cmidrule[0.1pt](r{0.80em}){6-8}%
        \cmidrule[0.1pt](r{0.80em}){9-11}%
        \cmidrule[0.1pt](r{0.80em}){12-14}%
        \rowcolor{gray!5!white}
        &GCD \cite{vaze2022generalized} 
        & $60.00$ & $70.71$ & $53.48$ 
        & $56.57$ & $67.14$ & $54.49$ 
        & $57.04$ & $53.57$ & $58.21$ 
        & $57.87$ \cellcolor{blue!7!white} & $63.80$ & $55.39$  \\
        \rowcolor{gray!5!white}
        &MetaGCD \cite{wu2023metagcd} 
        & $58.11$ & $64.29$ & $54.35$ 
        & $55.16$ & $74.29$ & $51.40$ 
        & $51.44$ & $54.29$ & $50.48$ 
        & $54.90$ \cellcolor{blue!7!white} & $64.29$ & $52.08$  \\
        \rowcolor{gray!5!white}
        &PA-CGCD \cite{kim2023proxy} 
        & $59.73$ & $75.71$ & $50.00$ 
        & $52.82$ & $72.86$ & $48.88$ 
        & $61.91$ & $84.29$ & $54.35$ 
        & $58.15$ \cellcolor{blue!7!white} & $77.62$ & $51.08$  \\
        \rowcolor{gray!5!white}
        &Happy~\cite{ma2024happy} 
        & $72.66$ & $82.86$ & $60.34$ 
        & $57.75$ & $71.43$ & $55.06$  
        & $63.18$ & $67.14$ & $61.84$ 
        & $64.53$ \cellcolor{blue!7!white} & $73.81$ & $59.08$   \\
        \rowcolor{blue!3!white}
        &PromptCCD (Ours) 
        & $61.35$ & $68.57$ & $56.96$ 
        & $58.92$ & $72.86$ & $56.18$ 
        & $67.87$ & $63.57$ & $69.32$ 
        & $62.71$ \cellcolor{blue!7!white} & $68.33$ & $60.82$  \\ 
        \rowcolor{blue!3!white}
        \rot{\rlap{Aircraft}\hskip 0.1in}
        &PromptCCD++ (Ours) 
        & $77.03$ & $83.57$ & $73.04$ 
        & $67.14$ & $78.57$ & $64.89$  
        & $70.04$ & $73.57$ & $68.84$ 
        & $\textbf{71.40}$ \cellcolor{blue!7!white} & $78.57$ & $68.92$   \\ \midrule
        \rowcolor{gray!5!white}
        &GCD \cite{vaze2022generalized} 
        & $60.51$ & $77.51$ & $50.25$ 
        & $58.41$ & $71.43$ & $56.67$ 
        & $56.65$ & $66.02$ & $54.49$ 
        & $58.52$ \cellcolor{blue!7!white} & $71.65$ & $53.80$  \\
        \rowcolor{gray!5!white}
        &MetaGCD \cite{wu2023metagcd} 
        & $59.08$ & $76.15$ & $48.77$ 
        & $57.35$ & $68.42$ & $55.87$ 
        & $55.06$ & $71.04$ & $51.38$ 
        & $57.16$ \cellcolor{blue!7!white} & $71.87$ & $52.01$  \\
        \rowcolor{gray!5!white}
        &PA-CGCD \cite{kim2023proxy} 
        & $68.51$ & $85.80$ & $58.07$ 
        & $65.08$ & $88.98$ & $61.89$ 
        & $61.15$ & $94.14$ & $53.56$ 
        & $64.91$ \cellcolor{blue!7!white} & $89.64$ & $57.84$  \\
        \rowcolor{gray!5!white}
        &Happy~\cite{ma2024happy} 
        & $70.20$ & $85.91$ & $59.91$ 
        & $68.15$ & $81.20$ & $66.81$  
        & $61.13$ & $79.15$ & $57.25$ 
        & $66.49$ \cellcolor{blue!7!white} & $76.88$ & $61.32$   \\
        \rowcolor{blue!3!white}
        &PromptCCD (Ours) 
        & $69.49$ & $85.09$ & $60.07$ 
        & $65.49$ & $73.68$ & $64.39$ 
        & $60.26$ & $71.04$ & $57.78$ 
        & $65.08$ \cellcolor{blue!7!white} & $76.60$ & $60.75$  \\
        \rowcolor{blue!3!white}
        \rot{\rlap{SCars}\hskip 0.1in}
        &PromptCCD++ (Ours) 
        & $72.04$ & $86.45$ & $63.34$ 
        & $73.10$ & $89.47$ & $70.91$  
        & $65.90$ & $81.85$ & $62.22$ 
        & $\textbf{70.35}$ \cellcolor{blue!7!white} & $85.92$ & $65.49$   \\ \midrule
        \rowcolor{gray!5!white}
        &GCD \cite{vaze2022generalized} 
        & $68.67$ & $85.00$ & $57.76$ 
        & $66.87$ & $80.71$ & $63.94$ 
        & $64.57$ & $84.29$ & $60.72$ 
        & $66.70$ \cellcolor{blue!7!white} & $83.33$ & $60.81$  \\
        \rowcolor{gray!5!white}
        &MetaGCD \cite{wu2023metagcd} 
        & $65.81$ & $84.64$ & $53.22$ 
        & $57.25$ & $80.00$ & $52.42$ 
        & $63.52$ & $82.86$ & $59.75$ 
        & $62.19$ \cellcolor{blue!7!white} & $82.50$ & $55.13$  \\
        \rowcolor{gray!5!white}
        &PA-CGCD \cite{kim2023proxy} 
        & $67.67$ & $87.86$ & $54.18$ 
        & $65.25$ & $99.29$ & $58.03$ 
        & $67.72$ & $90.71$ & $63.23$ 
        & $66.88$ \cellcolor{blue!7!white} & $92.62$ & $58.48$  \\
        \rowcolor{gray!5!white}
        &Happy~\cite{ma2024happy} 
        & $70.96$ & $87.50$ & $59.90$ 
        & $72.12$ & $82.86$ & $69.85$  
        & $73.31$ & $87.86$ & $70.47$ 
        & $72.13$ \cellcolor{blue!7!white} & $86.07$ & $66.74$   \\
        \rowcolor{blue!3!white}
        &PromptCCD (Ours) 
        & $69.10$ & $83.21$ & $59.67$ 
        & $63.00$ & $80.00$ & $59.39$ 
        & $71.33$ & $81.43$ & $69.36$ 
        & $67.81$ \cellcolor{blue!7!white} & $81.55$ & $62.81$  \\
        \rowcolor{blue!3!white}
        \rot{\rlap{CUB}}
        &PromptCCD++ (Ours) 
        & $72.68$ & $86.79$ & $63.25$ 
        & $72.75$ & $83.57$ & $70.45$  
        & $82.63$ & $88.57$ & $81.48$ 
        & $\textbf{76.02}$ \cellcolor{blue!7!white} & $86.31$ & $71.73$   \\
        \bottomrule
        \end{tabular}
    }
    \label{tab:main_complete_result_ssb_w_dinov2}
\end{table*}

\definecolor{Gray}{gray}{0.9}
\definecolor{PaleBlue}{rgb}{0.7529, 0.9137, 0.9372}
\definecolor{BeauBlue}{rgb}{0.7686, 0.8470, 0.9529}
\definecolor{Mauve}{rgb}{0.8 , 0.7098, 0.9843}
\definecolor{PaleViolet}{rgb}{0.8156, 0.6431, 1.0}
\definecolor{Salmon}{rgb}{1.0, 0.8980, 0.6920}
\definecolor{Pink}{rgb}{1.0, 0.6902, 0.7908}
\definecolor{Mint}{rgb}{0.6902, 1.0, 0.7451}
\definecolor{cyan}{rgb}{0.906, 0.969, 0.965}
\definecolor{SoftP}{rgb}{0.945, 0.933, 0.949}

\begin{table*}[!ht]
    \caption{Breakdown results of our method with different prompt pool designs for CCD leveraging pretrained DINO model on generic and CUB datasets with the \textit{known} $C$ in each unlabelled set. The experiments are conducted with \textit{seed 1}.}
    \centering
    \resizebox{1.0\columnwidth}{!}{%
    \centering
        \begin{tabular}{clcccc @{\hskip 0.1in} ccc@{\hskip 0.1in}ccc @{\hskip 0.1in} ||ccc}
        \toprule
        \multicolumn{3}{c}{}
        & \multicolumn{3}{c}{Stage 1 \textit{ACC} (\%)} 
        & \multicolumn{3}{c}{Stage 2 \textit{ACC} (\%)} 
        & \multicolumn{3}{c}{Stage 3 \textit{ACC} (\%)} 
        & \multicolumn{3}{c}{Average \textit{ACC} (\%)} 
        \\ 
        \multicolumn{1}{l}{} &
        \multicolumn{1}{c}{Method} &
        \multicolumn{1}{c}{Prompt Pool} &
        \multicolumn{1}{c}{\textit{All}} & 
        \multicolumn{1}{c}{\textit{Old}} &
        \multicolumn{1}{c}{\textit{New}} &
        \multicolumn{1}{c}{\textit{All}} & 
        \multicolumn{1}{c}{\textit{Old}} & 
        \multicolumn{1}{c}{\textit{New}} &
        \multicolumn{1}{c}{\textit{All}} & 
        \multicolumn{1}{c}{\textit{Old}} & 
        \multicolumn{1}{c}{\textit{New}} &
        \multicolumn{1}{c}{\textit{All} \cellcolor{blue!7!white}} & 
        \multicolumn{1}{c}{\textit{Old}} & 
        \multicolumn{1}{c}{\textit{New}} 
        \\ 
        \cmidrule[0.1pt](r{0.80em}){1-2}%
        \cmidrule[0.1pt](r{0.80em}){3-3}%
        \cmidrule[0.1pt](r{0.80em}){4-6}%
        \cmidrule[0.1pt](r{0.80em}){7-9}%
        \cmidrule[0.1pt](r{0.80em}){10-12}%
        \cmidrule[0.1pt](r{0.80em}){13-15}%
        \rowcolor{gray!5!white}
        &PromptCCD-B~(Ours) & L2P \cite{wang2022learning}
        & $65.93$  & $80.20$  & $55.94$ 
        & $56.72$  & $70.76$  & $54.04$ 
        & $51.55$  & $66.67$  & $48.90$ 
        & $58.07$ \cellcolor{blue!7!white}  & $72.54$  & $52.96$ \\
        \rowcolor{gray!5!white}
        &PromptCCD-B~(Ours) & DP \cite{wang2022dualprompt}
        & $69.92$  & $82.57$  & $61.06$ 
        & $55.21$  & $78.10$  & $51.75$ 
        & $54.37$  & $74.29$  & $50.88$ 
        & $59.83$ \cellcolor{blue!7!white}  & $78.32$  & $54.56$ \\
        \rowcolor{blue!3!white}
        \rot{\rlap{C100}}
        &PromptCCD~(Ours) & GMP (Ours)
        & $70.69$  & $80.90$  & $63.54$ 
        & $64.08$  & $73.14$  & $62.35$ 
        & $57.73$  & $72.67$  & $55.12$ 
        & $\textbf{64.17}$ \cellcolor{blue!7!white}  & $75.57$  & $60.34$ \\ \midrule
        \rowcolor{gray!5!white}
        &PromptCCD-B~(Ours) & L2P \cite{wang2022learning}
        & $75.80$  & $83.84$  & $70.17$ 
        & $70.95$  & $82.48$  & $68.75$ 
        & $61.19$  & $78.67$  & $58.13$ 
        & $69.31$ \cellcolor{blue!7!white}  & $81.66$  & $65.68$ \\
        \rowcolor{gray!5!white}
        &PromptCCD-B~(Ours) & DP \cite{wang2022dualprompt}
        & $76.39$  & $83.43$  & $71.46$ 
        & $67.90$  & $83.52$  & $63.96$ 
        & $61.13$  & $83.24$  & $57.57$ 
        & $68.47$ \cellcolor{blue!7!white}  & $83.40$  & $64.33$ \\
        \rowcolor{blue!3!white}
        \rot{\rlap{IN-100}}
        &PromptCCD~(Ours) & GMP (Ours)
        & $79.56$  & $84.24$  & $76.29$ 
        & $78.58$  & $79.71$  & $78.36$ 
        & $70.33$  & $81.33$  & $68.40$ 
        & $\textbf{76.16}$ \cellcolor{blue!7!white}  & $81.76$  & $74.35$ \\ \midrule
        \rowcolor{gray!5!white}
        &PromptCCD-B~(Ours) & L2P \cite{wang2022learning}
        & $69.46$  & $75.24$  & $65.41$ 
        & $54.64$  & $65.43$  & $52.58$ 
        & $44.88$  & $59.43$  & $42.33$ 
        & $56.33$ \cellcolor{blue!7!white}  & $66.70$  & $53.44$ \\
        \rowcolor{gray!5!white}
        &PromptCCD-B~(Ours) & DP \cite{wang2022dualprompt}
        & $65.91$  & $72.63$  & $61.60$ 
        & $55.73$  & $67.71$  & $53.44$ 
        & $48.28$  & $59.14$  & $46.37$ 
        & $56.64$ \cellcolor{blue!7!white}  & $66.49$  & $53.80$ \\
        \rowcolor{blue!3!white}
        \rot{\rlap{Tiny}}
        &PromptCCD~(Ours) & GMP (Ours)
        & $68.67$  & $72.84$  & $65.76$ 
        & $59.69$  & $65.67$  & $58.55$ 
        & $57.16$  & $61.10$  & $56.47$ 
        & $\textbf{61.84}$ \cellcolor{blue!7!white}  & $66.54$  & $60.26$ \\ \midrule
        \rowcolor{gray!5!white}
        &PromptCCD-B~(Ours) & L2P \cite{wang2022learning}
        & $56.65$  & $73.93$  & $45.11$ 
        & $47.75$  & $74.29$  & $42.12$ 
        & $47.32$  & $71.43$  & $42.62$ 
        & $50.57$ \cellcolor{blue!7!white}  & $73.22$  & $43.28$ \\
        \rowcolor{gray!5!white}
        &PromptCCD-B~(Ours) & DP \cite{wang2022dualprompt}
        & $60.66$  & $77.86$  & $49.16$ 
        & $47.62$  & $74.57$  & $41.91$ 
        & $57.34$  & $79.29$  & $53.06$ 
        & $55.21$ \cellcolor{blue!7!white}  & $77.24$  & $48.04$ \\
        \rowcolor{blue!3!white}
        \rot{\rlap{CUB}}
        &PromptCCD~(Ours) & GMP (Ours)
        & $57.08$  & $75.00$  & $45.11$ 
        & $47.38$  & $75.00$  & $41.52$ 
        & $61.89$  & $76.43$  & $59.05$ 
        & $\textbf{55.45}$ \cellcolor{blue!7!white}  & $75.48$  & $48.56$ \\
        \bottomrule
        \end{tabular}
    }
    \label{tab:main_complete_result_baselines}
\end{table*}
\definecolor{Gray}{gray}{0.9}
\definecolor{PaleBlue}{rgb}{0.7529, 0.9137, 0.9372}
\definecolor{BeauBlue}{rgb}{0.7686, 0.8470, 0.9529}
\definecolor{Mauve}{rgb}{0.8 , 0.7098, 0.9843}
\definecolor{PaleViolet}{rgb}{0.8156, 0.6431, 1.0}
\definecolor{Salmon}{rgb}{1.0, 0.8980, 0.6920}
\definecolor{Pink}{rgb}{1.0, 0.6902, 0.7908}
\definecolor{Mint}{rgb}{0.6902, 1.0, 0.7451}
\definecolor{cyan}{rgb}{0.906, 0.969, 0.965}
\definecolor{SoftP}{rgb}{0.945, 0.933, 0.949}

\begin{table*}[!ht]
    \caption{Breakdown results of our method with different prompt pool designs for CCD leveraging pretrained DINO model on generic and CUB datasets with the \textit{known} $C$ in each unlabelled set. The experiments are conducted with \textit{seed 7}.}
    \centering
    \resizebox{1.0\columnwidth}{!}{%
    \centering
        \begin{tabular}{clcccc @{\hskip 0.1in} ccc@{\hskip 0.1in}ccc @{\hskip 0.1in} ||ccc}
        \toprule
        \multicolumn{3}{c}{}
        & \multicolumn{3}{c}{Stage 1 \textit{ACC} (\%)} 
        & \multicolumn{3}{c}{Stage 2 \textit{ACC} (\%)} 
        & \multicolumn{3}{c}{Stage 3 \textit{ACC} (\%)} 
        & \multicolumn{3}{c}{Average \textit{ACC} (\%)} 
        \\ 
        \multicolumn{1}{l}{} &
        \multicolumn{1}{c}{Method} &
        \multicolumn{1}{c}{Prompt Pool} &
        \multicolumn{1}{c}{\textit{All}} & 
        \multicolumn{1}{c}{\textit{Old}} &
        \multicolumn{1}{c}{\textit{New}} &
        \multicolumn{1}{c}{\textit{All}} & 
        \multicolumn{1}{c}{\textit{Old}} & 
        \multicolumn{1}{c}{\textit{New}} &
        \multicolumn{1}{c}{\textit{All}} & 
        \multicolumn{1}{c}{\textit{Old}} & 
        \multicolumn{1}{c}{\textit{New}} &
        \multicolumn{1}{c}{\textit{All} \cellcolor{blue!7!white}} & 
        \multicolumn{1}{c}{\textit{Old}} & 
        \multicolumn{1}{c}{\textit{New}} 
        \\ 
        \cmidrule[0.1pt](r{0.80em}){1-2}%
        \cmidrule[0.1pt](r{0.80em}){3-3}%
        \cmidrule[0.1pt](r{0.80em}){4-6}%
        \cmidrule[0.1pt](r{0.80em}){7-9}%
        \cmidrule[0.1pt](r{0.80em}){10-12}%
        \cmidrule[0.1pt](r{0.80em}){13-15}%
        \rowcolor{gray!5!white}
        &PromptCCD-B~(Ours) & L2P \cite{wang2022learning}
        & $46.54$  & $61.06$  & $36.37$ 
        & $38.95$  & $48.57$  & $37.11$ 
        & $38.48$  & $46.29$  & $37.12$ 
        & $41.32$ \cellcolor{blue!7!white}  & $51.97$  & $36.87$ \\
        \rowcolor{gray!5!white}
        &PromptCCD-B~(Ours) & DP \cite{wang2022dualprompt}
        & $69.92$  & $82.57$  & $61.06$ 
        & $56.90$  & $78.10$  & $52.85$ 
        & $48.14$  & $70.67$  & $44.20$ 
        & $58.32$ \cellcolor{blue!7!white}  & $77.11$  & $52.70$ \\
        \rowcolor{blue!3!white}
        \rot{\rlap{C100}}
        &PromptCCD~(Ours) & GMP (Ours)
        & $74.54$  & $82.45$  & $69.00$ 
        & $65.82$  & $77.81$  & $63.53$ 
        & $58.07$  & $83.62$  & $53.60$ 
        & $\textbf{66.14}$ \cellcolor{blue!7!white}  & $81.29$  & $62.04$ \\ \midrule
        \rowcolor{gray!5!white}
        &PromptCCD-B~(Ours) & L2P \cite{wang2022learning}
        & $70.45$  & $80.37$  & $63.51$ 
        & $65.80$  & $80.38$  & $63.02$ 
        & $58.40$  & $77.05$  & $55.13$ 
        & $64.88$ \cellcolor{blue!7!white}  & $79.27$  & $60.55$ \\
        \rowcolor{gray!5!white}
        &PromptCCD-B~(Ours) & DP \cite{wang2022dualprompt}
        & $80.12$  & $83.31$  & $77.89$ 
        & $69.33$  & $82.76$  & $66.76$ 
        & $65.94$  & $84.00$  & $62.78$ 
        & $71.80$ \cellcolor{blue!7!white}  & $83.36$  & $69.14$ \\
        \rowcolor{blue!3!white}
        \rot{\rlap{IN-100}}
        &PromptCCD~(Ours) & GMP (Ours)
        & $78.40$  & $81.96$  & $75.91$ 
        & $76.66$  & $80.57$  & $75.91$ 
        & $67.60$  & $78.76$  & $65.65$ 
        & $\textbf{74.22}$ \cellcolor{blue!7!white}  & $80.43$  & $72.49$ \\ \midrule
        \rowcolor{gray!5!white}
        &PromptCCD-B~(Ours) & L2P \cite{wang2022learning}
        & $64.83$  & $73.02$  & $59.10$ 
        & $55.63$  & $65.76$  & $53.69$ 
        & $51.15$  & $57.43$  & $50.05$ 
        & $57.20$ \cellcolor{blue!7!white}  & $65.40$  & $54.28$ \\
        \rowcolor{gray!5!white}
        &PromptCCD-B~(Ours) & DP \cite{wang2022dualprompt}
        & $66.72$  & $74.51$  & $61.27$ 
        & $58.00$  & $66.52$  & $56.37$ 
        & $55.45$  & $61.38$  & $54.42$ 
        & $60.06$ \cellcolor{blue!7!white}  & $67.47$  & $57.35$ \\
        \rowcolor{blue!3!white}
        \rot{\rlap{Tiny}}
        &PromptCCD~(Ours) & GMP (Ours)
        & $68.31$  & $73.41$  & $64.74$ 
        & $58.36$  & $66.76$  & $56.75$ 
        & $55.45$  & $61.38$  & $54.42$ 
        & $\textbf{60.71}$ \cellcolor{blue!7!white}  & $63.85$  & $58.64$ \\ \midrule
        \rowcolor{gray!5!white}
        &PromptCCD-B~(Ours) & L2P \cite{wang2022learning}
        & $55.36$  & $74.29$  & $42.72$ 
        & $43.00$  & $75.00$  & $36.21$ 
        & $55.24$  & $71.43$  & $52.09$ 
        & $51.20$ \cellcolor{blue!7!white}  & $73.57$  & $43.67$ \\
        \rowcolor{gray!5!white}
        &PromptCCD-B~(Ours) & DP \cite{wang2022dualprompt}
        & $55.94$  & $77.50$  & $41.53$ 
        & $47.75$  & $80.00$  & $40.91$ 
        & $63.75$  & $77.14$  & $61.14$ 
        & $55.81$ \cellcolor{blue!7!white}  & $78.21$  & $47.86$ \\
        \rowcolor{blue!3!white}
        \rot{\rlap{CUB}}
        &PromptCCD~(Ours) & GMP (Ours)
        & $58.37$  & $78.21$  & $45.11$ 
        & $50.25$  & $77.14$  & $44.55$ 
        & $61.54$  & $85.71$  & $56.82$ 
        & $\textbf{56.72}$ \cellcolor{blue!7!white}  & $80.35$  & $48.83$ \\
        \bottomrule
        \end{tabular}
    }
    \label{tab:main_complete_result_baselines_seed_7}
\end{table*}
\definecolor{Gray}{gray}{0.9}
\definecolor{PaleBlue}{rgb}{0.7529, 0.9137, 0.9372}
\definecolor{BeauBlue}{rgb}{0.7686, 0.8470, 0.9529}
\definecolor{Mauve}{rgb}{0.8 , 0.7098, 0.9843}
\definecolor{PaleViolet}{rgb}{0.8156, 0.6431, 1.0}
\definecolor{Salmon}{rgb}{1.0, 0.8980, 0.6920}
\definecolor{Pink}{rgb}{1.0, 0.6902, 0.7908}
\definecolor{Mint}{rgb}{0.6902, 1.0, 0.7451}
\definecolor{cyan}{rgb}{0.906, 0.969, 0.965}
\definecolor{SoftP}{rgb}{0.945, 0.933, 0.949}

\begin{table*}[!ht]
    \caption{Breakdown results of our method with different prompt pool designs for CCD leveraging pretrained DINO model on generic and CUB datasets with the \textit{known} $C$ in each unlabelled set. The experiments are conducted with \textit{seed 10}.}
    \centering
    \resizebox{1.0\columnwidth}{!}{%
    \centering
        \begin{tabular}{clcccc @{\hskip 0.1in} ccc@{\hskip 0.1in}ccc @{\hskip 0.1in} ||ccc}
        \toprule
        \multicolumn{3}{c}{}
        & \multicolumn{3}{c}{Stage 1 \textit{ACC} (\%)} 
        & \multicolumn{3}{c}{Stage 2 \textit{ACC} (\%)} 
        & \multicolumn{3}{c}{Stage 3 \textit{ACC} (\%)} 
        & \multicolumn{3}{c}{Average \textit{ACC} (\%)} 
        \\ 
        \multicolumn{1}{l}{} &
        \multicolumn{1}{c}{Method} &
        \multicolumn{1}{c}{Prompt Pool} &
        \multicolumn{1}{c}{\textit{All}} & 
        \multicolumn{1}{c}{\textit{Old}} &
        \multicolumn{1}{c}{\textit{New}} &
        \multicolumn{1}{c}{\textit{All}} & 
        \multicolumn{1}{c}{\textit{Old}} & 
        \multicolumn{1}{c}{\textit{New}} &
        \multicolumn{1}{c}{\textit{All}} & 
        \multicolumn{1}{c}{\textit{Old}} & 
        \multicolumn{1}{c}{\textit{New}} &
        \multicolumn{1}{c}{\textit{All} \cellcolor{blue!7!white}} & 
        \multicolumn{1}{c}{\textit{Old}} & 
        \multicolumn{1}{c}{\textit{New}} 
        \\ 
        \cmidrule[0.1pt](r{0.80em}){1-2}%
        \cmidrule[0.1pt](r{0.80em}){3-3}%
        \cmidrule[0.1pt](r{0.80em}){4-6}%
        \cmidrule[0.1pt](r{0.80em}){7-9}%
        \cmidrule[0.1pt](r{0.80em}){10-12}%
        \cmidrule[0.1pt](r{0.80em}){13-15}%
        \rowcolor{gray!5!white}
        &PromptCCD-B~(Ours) & L2P \cite{wang2022learning}
        & $64.32$  & $78.04$  & $54.71$ 
        & $52.58$  & $69.62$  & $49.33$ 
        & $46.79$  & $60.95$  & $44.32$ 
        & $54.56$ \cellcolor{blue!7!white}  & $69.54$  & $49.45$ \\
        \rowcolor{gray!5!white}
        &PromptCCD-B~(Ours) & DP \cite{wang2022dualprompt}
        & $71.82$  & $83.51$  & $63.63$ 
        & $58.44$  & $77.90$  & $54.73$ 
        & $49.22$  & $75.90$  & $44.55$ 
        & $59.83$ \cellcolor{blue!7!white}  & $79.10$  & $54.30$ \\
        \rowcolor{blue!3!white}
        \rot{\rlap{C100}}
        &PromptCCD~(Ours) & GMP (Ours)
        & $73.82$  & $79.67$  & $69.71$ 
        & $63.48$  & $74.00$  & $61.47$ 
        & $55.46$  & $72.67$  & $52.45$ 
        & $\textbf{64.25}$ \cellcolor{blue!7!white}  & $75.45$  & $61.21$ \\ \midrule
        \rowcolor{gray!5!white}
        &PromptCCD-B~(Ours) & L2P \cite{wang2022learning}
        & $69.09$  & $83.80$  & $58.80$ 
        & $65.68$  & $81.14$  & $62.73$ 
        & $63.50$  & $79.81$  & $60.65$ 
        & $66.09$ \cellcolor{blue!7!white}  & $81.58$  & $60.73$ \\
        \rowcolor{gray!5!white}
        &PromptCCD-B~(Ours) & DP \cite{wang2022dualprompt}
        & $78.08$  & $84.12$  & $73.86$ 
        & $72.89$  & $83.24$  & $70.91$ 
        & $63.05$  & $85.14$  & $59.18$ 
        & $71.34$ \cellcolor{blue!7!white}  & $84.17$  & $67.98$ \\
        \rowcolor{blue!3!white}
        \rot{\rlap{IN-100}}
        &PromptCCD~(Ours) & GMP (Ours)
        & $79.97$  & $84.41$  & $76.86$ 
        & $77.16$  & $79.71$  & $76.67$ 
        & $69.39$  & $78.00$  & $67.88$ 
        & $\textbf{75.71}$ \cellcolor{blue!7!white}  & $80.71$  & $73.80$ \\ \midrule
        \rowcolor{gray!5!white}
        &PromptCCD-B~(Ours) & L2P \cite{wang2022learning}
        & $67.50$  & $75.53$  & $61.87$ 
        & $53.47$  & $64.81$  & $51.31$ 
        & $49.73$  & $59.05$  & $48.10$ 
        & $56.90$ \cellcolor{blue!7!white}  & $66.46$  & $53.76$ \\
        \rowcolor{gray!5!white}
        &PromptCCD-B~(Ours) & DP \cite{wang2022dualprompt}
        & $66.93$  & $74.94$  & $61.33$ 
        & $58.53$  & $63.43$  & $57.60$ 
        & $55.09$  & $59.90$  & $54.24$ 
        & $60.18$ \cellcolor{blue!7!white}  & $66.09$  & $57.72$ \\
        \rowcolor{blue!3!white}
        \rot{\rlap{Tiny}}
        &PromptCCD~(Ours) & GMP (Ours)
        & $69.95$  & $78.51$  & $63.51$ 
        & $59.22$  & $65.67$  & $56.04$ 
        & $57.88$  & $63.67$  & $54.88$ 
        & $\textbf{62.25}$ \cellcolor{blue!7!white}  & $69.28$  & $58.14$ \\ \midrule
        \rowcolor{gray!5!white}
        &PromptCCD-B~(Ours) & L2P \cite{wang2022learning}
        & $53.65$  & $70.00$  & $42.72$ 
        & $49.00$  & $72.86$  & $43.94$ 
        & $55.48$  & $71.43$  & $52.37$ 
        & $52.71$ \cellcolor{blue!7!white}  & $71.43$  & $46.34$ \\
        \rowcolor{gray!5!white}
        &PromptCCD-B~(Ours) & DP \cite{wang2022dualprompt}
        & $61.80$  & $80.36$  & $49.40$ 
        & $49.38$  & $80.71$  & $42.73$ 
        & $58.74$  & $80.00$  & $54.60$ 
        & $56.64$ \cellcolor{blue!7!white}  & $80.36$  & $48.91$ \\
        \rowcolor{blue!3!white}
        \rot{\rlap{CUB}}
        &PromptCCD~(Ours) & GMP (Ours)
        & $61.09$  & $81.07$  & $47.73$ 
        & $50.12$  & $85.71$  & $42.58$ 
        & $60.26$  & $79.29$  & $56.55$ 
        & $\textbf{57.16}$ \cellcolor{blue!7!white}  & $82.02$  & $48.95$ \\
        \bottomrule
        \end{tabular}
    }
    \label{tab:main_complete_result_baselines_seed_10}
\end{table*}
\definecolor{Gray}{gray}{0.9}
\definecolor{PaleBlue}{rgb}{0.7529, 0.9137, 0.9372}
\definecolor{BeauBlue}{rgb}{0.7686, 0.8470, 0.9529}
\definecolor{Mauve}{rgb}{0.8 , 0.7098, 0.9843}
\definecolor{PaleViolet}{rgb}{0.8156, 0.6431, 1.0}
\definecolor{Salmon}{rgb}{1.0, 0.8980, 0.6920}
\definecolor{Pink}{rgb}{1.0, 0.6902, 0.7908}
\definecolor{Mint}{rgb}{0.6902, 1.0, 0.7451}
\definecolor{cyan}{rgb}{0.906, 0.969, 0.965}
\definecolor{SoftP}{rgb}{0.945, 0.933, 0.949}

\begin{table*}[!ht]
    \caption{Breakdown results of our method with different prompt pool designs for CCD leveraging pretrained DINO model on generic and CUB datasets with the \textit{known} $C$ in each unlabelled set. The experiments are conducted with \textit{seed 2000}.}
    \centering
    \resizebox{1.0\columnwidth}{!}{%
    \centering
        \begin{tabular}{clcccc @{\hskip 0.1in} ccc@{\hskip 0.1in}ccc @{\hskip 0.1in} ||ccc}
        \toprule
        \multicolumn{3}{c}{}
        & \multicolumn{3}{c}{Stage 1 \textit{ACC} (\%)} 
        & \multicolumn{3}{c}{Stage 2 \textit{ACC} (\%)} 
        & \multicolumn{3}{c}{Stage 3 \textit{ACC} (\%)} 
        & \multicolumn{3}{c}{Average \textit{ACC} (\%)} 
        \\ 
        \multicolumn{1}{l}{} &
        \multicolumn{1}{c}{Method} &
        \multicolumn{1}{c}{Prompt Pool} &
        \multicolumn{1}{c}{\textit{All}} & 
        \multicolumn{1}{c}{\textit{Old}} &
        \multicolumn{1}{c}{\textit{New}} &
        \multicolumn{1}{c}{\textit{All}} & 
        \multicolumn{1}{c}{\textit{Old}} & 
        \multicolumn{1}{c}{\textit{New}} &
        \multicolumn{1}{c}{\textit{All}} & 
        \multicolumn{1}{c}{\textit{Old}} & 
        \multicolumn{1}{c}{\textit{New}} &
        \multicolumn{1}{c}{\textit{All} \cellcolor{blue!7!white}} & 
        \multicolumn{1}{c}{\textit{Old}} & 
        \multicolumn{1}{c}{\textit{New}} 
        \\ 
        \cmidrule[0.1pt](r{0.80em}){1-2}%
        \cmidrule[0.1pt](r{0.80em}){3-3}%
        \cmidrule[0.1pt](r{0.80em}){4-6}%
        \cmidrule[0.1pt](r{0.80em}){7-9}%
        \cmidrule[0.1pt](r{0.80em}){10-12}%
        \cmidrule[0.1pt](r{0.80em}){13-15}%
        \rowcolor{gray!5!white}
        &PromptCCD-B~(Ours) & L2P \cite{wang2022learning}
        & $60.40$  & $76.61$  & $49.06$ 
        & $50.79$  & $65.62$  & $47.20$ 
        & $41.04$  & $65.43$  & $36.77$ 
        & $50.74$ \cellcolor{blue!7!white}  & $69.22$  & $44.34$ \\
        \rowcolor{gray!5!white}
        &PromptCCD-B~(Ours) & DP \cite{wang2022dualprompt}
        & $70.97$  & $83.67$  & $62.09$ 
        & $60.33$  & $77.81$  & $57.13$ 
        & $44.52$  & $78.29$  & $38.62$ 
        & $58.61$ \cellcolor{blue!7!white}  & $79.92$  & $52.61$ \\
        \rowcolor{blue!3!white}
        \rot{\rlap{C100}}
        &PromptCCD~(Ours) & GMP (Ours)
        & $70.00$  & $80.94$  & $62.34$ 
        & $64.66$  & $74.10$  & $62.85$ 
        & $53.79$  & $72.19$  & $50.57$ 
        & $\textbf{62.82}$ \cellcolor{blue!7!white}  & $75.74$  & $58.59$ \\ \midrule
        \rowcolor{gray!5!white}
        &PromptCCD-B~(Ours) & L2P \cite{wang2022learning}
        & $74.97$  & $83.35$  & $69.11$ 
        & $70.08$  & $81.52$  & $67.89$ 
        & $56.18$  & $84.00$  & $51.32$ 
        & $67.08$ \cellcolor{blue!7!white}  & $82.96$  & $62.77$ \\
        \rowcolor{gray!5!white}
        &PromptCCD-B~(Ours) & DP \cite{wang2022dualprompt}
        & $77.51$  & $83.96$  & $73.00$ 
        & $70.21$  & $82.29$  & $67.91$ 
        & $65.65$  & $83.71$  & $62.48$ 
        & $71.12$ \cellcolor{blue!7!white}  & $83.32$  & $67.80$ \\
        \rowcolor{blue!3!white}
        \rot{\rlap{IN-100}}
        &PromptCCD~(Ours) & GMP (Ours)
        & $80.30$  & $83.10$  & $78.34$ 
        & $74.92$  & $82.29$  & $73.51$ 
        & $70.54$  & $80.48$  & $68.80$ 
        & $\textbf{75.25}$ \cellcolor{blue!7!white}  & $81.96$  & $73.55$ \\ \midrule
        \rowcolor{gray!5!white}
        &PromptCCD-B~(Ours) & L2P \cite{wang2022learning}
        & $66.36$  & $73.41$  & $61.43$ 
        & $54.65$  & $66.38$  & $52.41$ 
        & $49.02$  & $60.33$  & $47.04$ 
        & $56.68$ \cellcolor{blue!7!white}  & $66.71$  & $53.63$ \\
        \rowcolor{gray!5!white}
        &PromptCCD-B~(Ours) & DP \cite{wang2022dualprompt}
        & $66.89$  & $73.98$  & $61.93$ 
        & $55.05$  & $66.90$  & $52.78$ 
        & $50.68$  & $60.24$  & $49.01$ 
        & $57.54$ \cellcolor{blue!7!white}  & $67.04$  & $54.57$ \\
        \rowcolor{blue!3!white}
        \rot{\rlap{Tiny}}
        &PromptCCD~(Ours) & GMP (Ours)
        & $66.82$  & $72.16$  & $63.07$ 
        & $59.69$  & $66.57$  & $58.38$ 
        & $57.26$  & $60.33$  & $56.73$ 
        & $\textbf{61.26}$ \cellcolor{blue!7!white}  & $66.35$  & $59.39$ \\ \midrule
        \rowcolor{gray!5!white}
        &PromptCCD-B~(Ours) & L2P \cite{wang2022learning}
        & $52.65$  & $70.36$  & $40.81$ 
        & $46.50$  & $74.29$  & $40.61$ 
        & $51.52$  & $70.00$  & $47.91$ 
        & $50.22$ \cellcolor{blue!7!white}  & $71.55$  & $43.11$ \\
        \rowcolor{gray!5!white}
        &PromptCCD-B~(Ours) & DP \cite{wang2022dualprompt}
        & $61.95$  & $80.00$  & $49.88$ 
        & $51.25$  & $80.71$  & $45.00$ 
        & $61.19$  & $80.71$  & $57.38$ 
        & $\textbf{58.13}$ \cellcolor{blue!7!white}  & $80.47$  & $50.75$ \\
        \rowcolor{blue!3!white}
        \rot{\rlap{CUB}}
        &PromptCCD~(Ours) & GMP (Ours)
        & $56.80$  & $75.00$  & $44.63$ 
        & $48.88$  & $78.57$  & $42.58$ 
        & $62.00$  & $86.43$  & $57.24$ 
        & $55.89$ \cellcolor{blue!7!white}  & $80.00$  & $48.15$ \\
        \bottomrule
        \end{tabular}
    }
    \label{tab:main_complete_result_baselines_seed_2000}
\end{table*}
\definecolor{Gray}{gray}{0.9}
\definecolor{PaleBlue}{rgb}{0.7529, 0.9137, 0.9372}
\definecolor{BeauBlue}{rgb}{0.7686, 0.8470, 0.9529}
\definecolor{Mauve}{rgb}{0.8 , 0.7098, 0.9843}
\definecolor{PaleViolet}{rgb}{0.8156, 0.6431, 1.0}
\definecolor{Salmon}{rgb}{1.0, 0.8980, 0.6920}
\definecolor{Pink}{rgb}{1.0, 0.6902, 0.7908}
\definecolor{Mint}{rgb}{0.6902, 1.0, 0.7451}
\definecolor{cyan}{rgb}{0.906, 0.969, 0.965}
\definecolor{SoftP}{rgb}{0.945, 0.933, 0.949}

\begin{table*}[!ht]
    \caption{Breakdown results of our method with different prompt pool designs for CCD leveraging pretrained DINO model on generic and CUB datasets with the \textit{known} $C$ in each unlabelled set. The experiments are conducted with \textit{seed 2024}.}
    \centering
    \resizebox{1.0\columnwidth}{!}{%
    \centering
        \begin{tabular}{clcccc @{\hskip 0.1in} ccc@{\hskip 0.1in}ccc @{\hskip 0.1in} ||ccc}
        \toprule
        \multicolumn{3}{c}{}
        & \multicolumn{3}{c}{Stage 1 \textit{ACC} (\%)} 
        & \multicolumn{3}{c}{Stage 2 \textit{ACC} (\%)} 
        & \multicolumn{3}{c}{Stage 3 \textit{ACC} (\%)} 
        & \multicolumn{3}{c}{Average \textit{ACC} (\%)} 
        \\ 
        \multicolumn{1}{l}{} &
        \multicolumn{1}{c}{Method} &
        \multicolumn{1}{c}{Prompt Pool} &
        \multicolumn{1}{c}{\textit{All}} & 
        \multicolumn{1}{c}{\textit{Old}} &
        \multicolumn{1}{c}{\textit{New}} &
        \multicolumn{1}{c}{\textit{All}} & 
        \multicolumn{1}{c}{\textit{Old}} & 
        \multicolumn{1}{c}{\textit{New}} &
        \multicolumn{1}{c}{\textit{All}} & 
        \multicolumn{1}{c}{\textit{Old}} & 
        \multicolumn{1}{c}{\textit{New}} &
        \multicolumn{1}{c}{\textit{All} \cellcolor{blue!7!white}} & 
        \multicolumn{1}{c}{\textit{Old}} & 
        \multicolumn{1}{c}{\textit{New}} 
        \\ 
        \cmidrule[0.1pt](r{0.80em}){1-2}%
        \cmidrule[0.1pt](r{0.80em}){3-3}%
        \cmidrule[0.1pt](r{0.80em}){4-6}%
        \cmidrule[0.1pt](r{0.80em}){7-9}%
        \cmidrule[0.1pt](r{0.80em}){10-12}%
        \cmidrule[0.1pt](r{0.80em}){13-15}%
        \rowcolor{gray!5!white}
        &PromptCCD-B~(Ours) & L2P \cite{wang2022learning}
        & $65.56$  & $80.53$  & $55.09$ 
        & $52.26$  & $69.33$  & $49.00$ 
        & $41.87$  & $69.33$  & $37.07$ 
        & $53.23$ \cellcolor{blue!7!white}  & $73.06$  & $47.05$ \\
        \rowcolor{gray!5!white}
        &PromptCCD-B~(Ours) & DP \cite{wang2022dualprompt}
        & $74.79$  & $83.47$  & $68.71$ 
        & $60.34$  & $77.71$  & $57.02$ 
        & $49.18$  & $79.52$  & $43.87$ 
        & $61.44$ \cellcolor{blue!7!white}  & $80.23$  & $56.53$ \\
        \rowcolor{blue!3!white}
        \rot{\rlap{C100}}
        &PromptCCD~(Ours) & GMP (Ours)
        & $69.48$  & $82.57$  & $60.31$ 
        & $62.96$  & $73.71$  & $60.91$ 
        & $55.02$  & $69.62$  & $52.47$ 
        & $\textbf{62.49}$ \cellcolor{blue!7!white}  & $75.30$  & $57.90$ \\ \midrule
        \rowcolor{gray!5!white}
        &PromptCCD-B~(Ours) & L2P \cite{wang2022learning}
        & $68.87$  & $82.82$  & $57.51$ 
        & $64.19$  & $79.24$  & $60.36$ 
        & $56.92$  & $77.33$  & $53.30$ 
        & $63.33$ \cellcolor{blue!7!white}  & $79.80$  & $57.06$ \\
        \rowcolor{gray!5!white}
        &PromptCCD-B~(Ours) & DP \cite{wang2022dualprompt}
        & $76.82$  & $84.16$  & $71.69$ 
        & $71.74$  & $82.95$  & $69.60$ 
        & $62.89$  & $82.10$  & $59.53$ 
        & $70.48$ \cellcolor{blue!7!white}  & $83.07$  & $66.94$ \\
        \rowcolor{blue!3!white}
        \rot{\rlap{IN-100}}
        &PromptCCD~(Ours) & GMP (Ours)
        & $80.64$  & $83.35$  & $78.64$ 
        & $79.24$  & $80.86$  & $78.93$ 
        & $67.45$  & $78.67$  & $65.48$ 
        & $\textbf{75.78}$ \cellcolor{blue!7!white}  & $80.96$  & $74.35$ \\ \midrule
        \rowcolor{gray!5!white}
        &PromptCCD-B~(Ours) & L2P \cite{wang2022learning}
        & $66.03$  & $73.39$  & $60.89$ 
        & $52.70$  & $62.38$  & $50.85$ 
        & $49.88$  & $59.10$  & $48.27$ 
        & $56.20$ \cellcolor{blue!7!white}  & $64.96$  & $53.34$ \\
        \rowcolor{gray!5!white}
        &PromptCCD-B~(Ours) & DP \cite{wang2022dualprompt}
        & $63.58$  & $72.51$  & $57.33$ 
        & $56.19$  & $64.43$  & $54.62$ 
        & $56.15$  & $60.95$  & $55.31$ 
        & $58.64$ \cellcolor{blue!7!white}  & $65.96$  & $55.75$ \\
        \rowcolor{blue!3!white}
        \rot{\rlap{Tiny}}
        &PromptCCD~(Ours) & GMP (Ours)
        & $66.88$  & $72.08$  & $63.24$ 
        & $57.23$  & $63.33$  & $56.06$ 
        & $54.91$  & $60.90$  & $53.87$ 
        & $\textbf{59.67}$ \cellcolor{blue!7!white}  & $65.44$  & $57.72$ \\ \midrule
        \rowcolor{gray!5!white}
        &PromptCCD-B~(Ours) & L2P \cite{wang2022learning}
        & $52.50$  & $70.00$  & $40.81$ 
        & $47.25$  & $76.43$  & $41.06$ 
        & $55.83$  & $70.71$  & $52.92$ 
        & $51.86$ \cellcolor{blue!7!white}  & $72.38$  & $44.93$ \\
        \rowcolor{gray!5!white}
        &PromptCCD-B~(Ours) & DP \cite{wang2022dualprompt}
        & $58.66$  & $74.29$  & $48.21$ 
        & $50.00$  & $78.57$  & $43.94$ 
        & $58.51$  & $77.86$  & $54.74$ 
        & $55.72$ \cellcolor{blue!7!white}  & $76.91$  & $48.96$ \\
        \rowcolor{blue!3!white}
        \rot{\rlap{CUB}}
        &PromptCCD~(Ours) & GMP (Ours)
        & $61.37$  & $80.36$  & $48.69$ 
        & $51.75$  & $78.57$  & $46.06$ 
        & $60.96$  & $85.71$  & $56.13$ 
        & $\textbf{58.03}$ \cellcolor{blue!7!white}  & $81.55$  & $50.29$ \\
        \bottomrule
        \end{tabular}
    }
    \label{tab:main_complete_result_baselines_seed_2024}
\end{table*}

\definecolor{Gray}{gray}{0.9}
\definecolor{PaleBlue}{rgb}{0.7529, 0.9137, 0.9372}
\definecolor{BeauBlue}{rgb}{0.7686, 0.8470, 0.9529}
\definecolor{Mauve}{rgb}{0.8 , 0.7098, 0.9843}
\definecolor{PaleViolet}{rgb}{0.8156, 0.6431, 1.0}
\definecolor{Salmon}{rgb}{1.0, 0.8980, 0.6920}
\definecolor{Pink}{rgb}{1.0, 0.6902, 0.7908}
\definecolor{Mint}{rgb}{0.6902, 1.0, 0.7451}
\definecolor{cyan}{rgb}{0.906, 0.969, 0.965}
\definecolor{SoftP}{rgb}{0.945, 0.933, 0.949}

\begin{table*}[!ht]
    \caption{Breakdown results of our method with different prompt pool designs for CCD leveraging pretrained DINOv2 model on generic and CUB datasets with the \textit{known} $C$ in each unlabelled set. The experiments are conducted with \textit{seed 1}.}
    \centering
    \resizebox{1.0\columnwidth}{!}{%
    \centering
        \begin{tabular}{clcccc @{\hskip 0.1in} ccc@{\hskip 0.1in}ccc @{\hskip 0.1in} ||ccc}
        \toprule
        \multicolumn{3}{c}{}
        & \multicolumn{3}{c}{Stage 1 \textit{ACC} (\%)} 
        & \multicolumn{3}{c}{Stage 2 \textit{ACC} (\%)} 
        & \multicolumn{3}{c}{Stage 3 \textit{ACC} (\%)} 
        & \multicolumn{3}{c}{Average \textit{ACC} (\%)} 
        \\ 
        \multicolumn{1}{l}{} &
        \multicolumn{1}{c}{Method} &
        \multicolumn{1}{c}{Prompt Pool} &
        \multicolumn{1}{c}{\textit{All}} & 
        \multicolumn{1}{c}{\textit{Old}} &
        \multicolumn{1}{c}{\textit{New}} &
        \multicolumn{1}{c}{\textit{All}} & 
        \multicolumn{1}{c}{\textit{Old}} & 
        \multicolumn{1}{c}{\textit{New}} &
        \multicolumn{1}{c}{\textit{All}} & 
        \multicolumn{1}{c}{\textit{Old}} & 
        \multicolumn{1}{c}{\textit{New}} &
        \multicolumn{1}{c}{\textit{All} \cellcolor{blue!7!white}} & 
        \multicolumn{1}{c}{\textit{Old}} & 
        \multicolumn{1}{c}{\textit{New}} 
        \\ 
        \cmidrule[0.1pt](r{0.80em}){1-2}%
        \cmidrule[0.1pt](r{0.80em}){3-3}%
        \cmidrule[0.1pt](r{0.80em}){4-6}%
        \cmidrule[0.1pt](r{0.80em}){7-9}%
        \cmidrule[0.1pt](r{0.80em}){10-12}%
        \cmidrule[0.1pt](r{0.80em}){13-15}%
        \rowcolor{blue!3!white}
        & & & & & & & & & & & & \cellcolor{blue!7!white} & & \\
        \rowcolor{blue!3!white}
        &PromptCCD~(Ours) & GMP (Ours)
        & $78.24$ & $90.04$ & $69.97$ 
        & $65.27$ & $74.67$ & $63.46$ 
        & $65.69$ & $69.33$ & $65.05$ 
        & $69.73$ \cellcolor{blue!7!white} & $78.01$ & $66.16$  \\
        \rowcolor{blue!3!white}
        \rot{\rlap{C100}}
        &PromptCCD++~(Ours) & PLP (Ours)
        & $82.59$ & $87.27$ & $79.31$ 
        & $81.30$ & $82.95$ & $80.98$  
        & $69.15$ & $79.33$ & $67.37$ 
        & $\textbf{77.68}$ \cellcolor{blue!7!white} & $83.18$ & $75.89$   \\ \midrule
        \rowcolor{blue!3!white}
        & & & & & & & & & & & & \cellcolor{blue!7!white} & & \\
        \rowcolor{blue!3!white}
        &PromptCCD~(Ours) & GMP (Ours)
        & $80.35$ & $86.78$ & $75.86$ 
        & $75.65$ & $80.86$ & $74.65$ 
        & $72.85$ & $80.19$ & $73.08$ 
        & $76.28$ \cellcolor{blue!7!white} & $82.61$ & $74.53$ \\ 
        \rowcolor{blue!3!white}
        \rot{\rlap{IN-100}}
        &PromptCCD++~(Ours) & PLP (Ours)
        & $81.09$ & $86.33$ & $77.43$ 
        & $82.63$ & $84.95$ & $82.18$  
        & $86.45$ & $83.90$ & $86.90$ 
        & $\textbf{83.39}$ \cellcolor{blue!7!white} & $85.06$ & $82.17$   \\ \midrule
        \rowcolor{blue!3!white}
        & & & & & & & & & & & & \cellcolor{blue!7!white} & & \\
        \rowcolor{blue!3!white}
        &PromptCCD~(Ours) & GMP (Ours)
        & $74.30$ & $83.69$ & $67.73$ 
        & $67.00$ & $75.86$ & $65.31$ 
        & $63.31$ & $67.14$ & $62.64$ 
        & $68.20$ \cellcolor{blue!7!white} & $75.56$ & $65.23$  \\
        \rowcolor{blue!3!white}
        \rot{\rlap{Tiny}}
        &PromptCCD++~(Ours) & PLP (Ours)
        & $74.97$ & $81.90$ & $70.13$ 
        & $70.69$ & $78.76$ & $69.15$  
        & $73.41$ & $79.48$ & $72.35$ 
        & $\textbf{73.02}$ \cellcolor{blue!7!white} & $80.15$ & $70.54$   \\ \midrule
        \rowcolor{blue!3!white}
        & & & & & & & & & & & & \cellcolor{blue!7!white} & & \\
        \rowcolor{blue!3!white}
        &PromptCCD~(Ours) & GMP (Ours)
        & $69.10$ & $83.21$ & $59.67$ 
        & $63.00$ & $80.00$ & $59.39$ 
        & $71.33$ & $81.43$ & $69.36$ 
        & $67.81$ \cellcolor{blue!7!white} & $81.55$ & $62.81$  \\
        \rowcolor{blue!3!white}
        \rot{\rlap{CUB}}
        &PromptCCD++~(Ours) & PLP (Ours)
        & $72.68$ & $86.79$ & $63.25$ 
        & $72.75$ & $83.57$ & $70.45$  
        & $82.63$ & $88.57$ & $81.48$ 
        & $\textbf{76.02}$ \cellcolor{blue!7!white} & $86.31$ & $71.73$   \\
        \bottomrule
        \end{tabular}
    }
    \label{tab:main_complete_result_baselines_seed_1_dinov2}
\end{table*}
\definecolor{Gray}{gray}{0.9}
\definecolor{PaleBlue}{rgb}{0.7529, 0.9137, 0.9372}
\definecolor{BeauBlue}{rgb}{0.7686, 0.8470, 0.9529}
\definecolor{Mauve}{rgb}{0.8 , 0.7098, 0.9843}
\definecolor{PaleViolet}{rgb}{0.8156, 0.6431, 1.0}
\definecolor{Salmon}{rgb}{1.0, 0.8980, 0.6920}
\definecolor{Pink}{rgb}{1.0, 0.6902, 0.7908}
\definecolor{Mint}{rgb}{0.6902, 1.0, 0.7451}
\definecolor{cyan}{rgb}{0.906, 0.969, 0.965}
\definecolor{SoftP}{rgb}{0.945, 0.933, 0.949}

\begin{table*}[!ht]
    \caption{Breakdown results of our method with different prompt pool designs for CCD leveraging pretrained DINOv2 model on generic and CUB datasets with the \textit{known} $C$ in each unlabelled set. The experiments are conducted with \textit{seed 7}.}
    \centering
    \resizebox{1.0\columnwidth}{!}{%
    \centering
        \begin{tabular}{clcccc @{\hskip 0.1in} ccc@{\hskip 0.1in}ccc @{\hskip 0.1in} ||ccc}
        \toprule
        \multicolumn{3}{c}{}
        & \multicolumn{3}{c}{Stage 1 \textit{ACC} (\%)} 
        & \multicolumn{3}{c}{Stage 2 \textit{ACC} (\%)} 
        & \multicolumn{3}{c}{Stage 3 \textit{ACC} (\%)} 
        & \multicolumn{3}{c}{Average \textit{ACC} (\%)} 
        \\ 
        \multicolumn{1}{l}{} &
        \multicolumn{1}{c}{Method} &
        \multicolumn{1}{c}{Prompt Pool} &
        \multicolumn{1}{c}{\textit{All}} & 
        \multicolumn{1}{c}{\textit{Old}} &
        \multicolumn{1}{c}{\textit{New}} &
        \multicolumn{1}{c}{\textit{All}} & 
        \multicolumn{1}{c}{\textit{Old}} & 
        \multicolumn{1}{c}{\textit{New}} &
        \multicolumn{1}{c}{\textit{All}} & 
        \multicolumn{1}{c}{\textit{Old}} & 
        \multicolumn{1}{c}{\textit{New}} &
        \multicolumn{1}{c}{\textit{All} \cellcolor{blue!7!white}} & 
        \multicolumn{1}{c}{\textit{Old}} & 
        \multicolumn{1}{c}{\textit{New}} 
        \\ 
        \cmidrule[0.1pt](r{0.80em}){1-2}%
        \cmidrule[0.1pt](r{0.80em}){3-3}%
        \cmidrule[0.1pt](r{0.80em}){4-6}%
        \cmidrule[0.1pt](r{0.80em}){7-9}%
        \cmidrule[0.1pt](r{0.80em}){10-12}%
        \cmidrule[0.1pt](r{0.80em}){13-15}%
        \rowcolor{blue!3!white}
        & & & & & & & & & & & & \cellcolor{blue!7!white} & & \\
        \rowcolor{blue!3!white}
        &PromptCCD~(Ours) & GMP (Ours)
        & $77.13$  & $91.59$  & $67.00$ 
        & $71.77$  & $77.90$  & $70.60$ 
        & $61.56$  & $70.86$  & $59.93$ 
        & $70.15$ \cellcolor{blue!7!white}  & $80.12$  & $65.84$ \\
        \rowcolor{blue!3!white}
        \rot{\rlap{C100}}
        &PromptCCD++~(Ours) & PLP (Ours)
        & $78.17$  & $86.94$  & $72.03$ 
        & $82.95$  & $84.95$  & $82.56$ 
        & $69.12$  & $78.48$  & $67.48$ 
        & $\textbf{76.74}$ \cellcolor{blue!7!white}  & $83.46$  & $74.02$ \\ \midrule
        \rowcolor{blue!3!white}
        & & & & & & & & & & & & \cellcolor{blue!7!white} & & \\
        \rowcolor{blue!3!white}
        &PromptCCD~(Ours) & GMP (Ours)
        & $78.97$  & $85.67$  & $74.29$ 
        & $76.47$  & $84.00$  & $75.04$ 
        & $75.77$  & $72.57$  & $76.33$ 
        & $77.07$ \cellcolor{blue!7!white}  & $80.74$  & $75.22$ \\
        \rowcolor{blue!3!white}
        \rot{\rlap{IN-100}}
        &PromptCCD++~(Ours) & PLP (Ours)
        & $80.79$  & $84.20$  & $78.40$ 
        & $81.94$  & $84.67$  & $81.42$ 
        & $83.13$  & $84.95$  & $82.82$ 
        & $\textbf{81.95}$ \cellcolor{blue!7!white}  & $84.61$  & $80.88$ \\ \midrule
        \rowcolor{blue!3!white}
        & & & & & & & & & & & & \cellcolor{blue!7!white} & & \\
        \rowcolor{blue!3!white}
        &PromptCCD~(Ours) & GMP (Ours)
        & $73.41$  & $83.12$  & $66.61$ 
        & $65.55$  & $73.76$  & $63.98$ 
        & $63.85$  & $65.62$  & $63.54$ 
        & $67.60$ \cellcolor{blue!7!white}  & $74.17$  & $64.71$ \\
        \rowcolor{blue!3!white}
        \rot{\rlap{Tiny}}
        &PromptCCD++~(Ours) & PLP (Ours)
        & $72.48$  & $81.37$  & $66.26$ 
        & $70.52$  & $79.52$  & $68.80$ 
        & $71.37$  & $76.95$  & $70.39$ 
        & $\textbf{71.46}$ \cellcolor{blue!7!white}  & $79.28$  & $68.48$ \\ \midrule
        \rowcolor{blue!3!white}
        & & & & & & & & & & & & \cellcolor{blue!7!white} & & \\
        \rowcolor{gray!5!white}
        \rowcolor{blue!3!white}
        &PromptCCD~(Ours) & GMP (Ours)
        & $69.81$  & $86.79$  & $58.47$ 
        & $67.75$  & $86.43$  & $63.79$ 
        & $70.51$  & $86.43$  & $67.41$ 
        & $69.36$ \cellcolor{blue!7!white}  & $86.55$  & $63.22$ \\
        \rowcolor{blue!3!white}
        \rot{\rlap{CUB}}
        &PromptCCD++~(Ours) & PLP (Ours)
        & $74.19$  & $84.29$  & $64.36$ 
        & $73.59$  & $90.00$  & $70.79$ 
        & $76.81$  & $85.71$  & $74.74$ 
        & $\textbf{74.86}$ \cellcolor{blue!7!white}  & $86.67$  & $69.96$ \\
        \bottomrule
        \end{tabular}
    }
    \label{tab:main_complete_result_baselines_seed_7_dinov2}
\end{table*}
\definecolor{Gray}{gray}{0.9}
\definecolor{PaleBlue}{rgb}{0.7529, 0.9137, 0.9372}
\definecolor{BeauBlue}{rgb}{0.7686, 0.8470, 0.9529}
\definecolor{Mauve}{rgb}{0.8 , 0.7098, 0.9843}
\definecolor{PaleViolet}{rgb}{0.8156, 0.6431, 1.0}
\definecolor{Salmon}{rgb}{1.0, 0.8980, 0.6920}
\definecolor{Pink}{rgb}{1.0, 0.6902, 0.7908}
\definecolor{Mint}{rgb}{0.6902, 1.0, 0.7451}
\definecolor{cyan}{rgb}{0.906, 0.969, 0.965}
\definecolor{SoftP}{rgb}{0.945, 0.933, 0.949}

\begin{table*}[!ht]
    \caption{Breakdown results of our method with different prompt pool designs for CCD leveraging pretrained DINOv2 model on generic and CUB datasets with the \textit{known} $C$ in each unlabelled set. The experiments are conducted with \textit{seed 10}.}
    \centering
    \resizebox{1.0\columnwidth}{!}{%
    \centering
        \begin{tabular}{clcccc @{\hskip 0.1in} ccc@{\hskip 0.1in}ccc @{\hskip 0.1in} ||ccc}
        \toprule
        \multicolumn{3}{c}{}
        & \multicolumn{3}{c}{Stage 1 \textit{ACC} (\%)} 
        & \multicolumn{3}{c}{Stage 2 \textit{ACC} (\%)} 
        & \multicolumn{3}{c}{Stage 3 \textit{ACC} (\%)} 
        & \multicolumn{3}{c}{Average \textit{ACC} (\%)} 
        \\ 
        \multicolumn{1}{l}{} &
        \multicolumn{1}{c}{Method} &
        \multicolumn{1}{c}{Prompt Pool} &
        \multicolumn{1}{c}{\textit{All}} & 
        \multicolumn{1}{c}{\textit{Old}} &
        \multicolumn{1}{c}{\textit{New}} &
        \multicolumn{1}{c}{\textit{All}} & 
        \multicolumn{1}{c}{\textit{Old}} & 
        \multicolumn{1}{c}{\textit{New}} &
        \multicolumn{1}{c}{\textit{All}} & 
        \multicolumn{1}{c}{\textit{Old}} & 
        \multicolumn{1}{c}{\textit{New}} &
        \multicolumn{1}{c}{\textit{All} \cellcolor{blue!7!white}} & 
        \multicolumn{1}{c}{\textit{Old}} & 
        \multicolumn{1}{c}{\textit{New}} 
        \\ 
        \cmidrule[0.1pt](r{0.80em}){1-2}%
        \cmidrule[0.1pt](r{0.80em}){3-3}%
        \cmidrule[0.1pt](r{0.80em}){4-6}%
        \cmidrule[0.1pt](r{0.80em}){7-9}%
        \cmidrule[0.1pt](r{0.80em}){10-12}%
        \cmidrule[0.1pt](r{0.80em}){13-15}%
        \rowcolor{blue!3!white}
        & & & & & & & & & & & & \cellcolor{blue!7!white} & & \\
        \rowcolor{blue!3!white}
        &PromptCCD~(Ours) & GMP (Ours)
        & $76.92$  & $90.82$  & $67.20$ 
        & $63.28$  & $75.14$  & $61.02$ 
        & $61.90$  & $67.62$  & $60.90$ 
        & $67.37$ \cellcolor{blue!7!white}  & $77.86$  & $63.04$ \\
        \rowcolor{blue!3!white}
        \rot{\rlap{C100}}
        &PromptCCD++~(Ours) & PLP (Ours)
        & $80.94$  & $87.76$  & $76.17$ 
        & $83.91$  & $81.90$  & $84.29$ 
        & $70.23$  & $81.33$  & $68.28$ 
        & $\textbf{78.36}$ \cellcolor{blue!7!white}  & $83.66$  & $76.25$ \\ \midrule
        \rowcolor{blue!3!white}
        & & & & & & & & & & & & \cellcolor{blue!7!white} & & \\
        \rowcolor{blue!3!white}
        &PromptCCD~(Ours) & GMP (Ours)
        & $79.38$  & $87.22$  & $73.89$ 
        & $75.74$  & $80.57$  & $74.82$ 
        & $74.95$  & $75.90$  & $74.78$ 
        & $76.69$ \cellcolor{blue!7!white}  & $81.23$  & $74.50$ \\
        \rowcolor{blue!3!white}
        \rot{\rlap{IN-100}}
        &PromptCCD++~(Ours) & PLP (Ours)
        & $81.53$  & $85.67$  & $78.63$ 
        & $78.99$  & $82.86$  & $78.25$ 
        & $87.99$  & $82.86$  & $88.88$ 
        & $\textbf{82.84}$ \cellcolor{blue!7!white} & $83.80$  & $81.92$ \\ \midrule
        \rowcolor{blue!3!white}
        & & & & & & & & & & & & \cellcolor{blue!7!white} & & \\
        \rowcolor{blue!3!white}
        &PromptCCD~(Ours) & GMP (Ours)
        & $74.39$  & $84.96$  & $66.99$ 
        & $63.18$  & $74.48$  & $61.03$ 
        & $62.57$  & $69.05$  & $61.43$ 
        & $66.71$ \cellcolor{blue!7!white}  & $76.16$  & $63.15$ \\
        \rowcolor{blue!3!white}
        \rot{\rlap{Tiny}}
        &PromptCCD++~(Ours) & PLP (Ours)
        & $75.40$  & $82.12$  & $70.70$ 
        & $73.05$  & $80.05$  & $71.72$ 
        & $70.82$  & $77.19$  & $69.71$ 
        & $\textbf{73.09}$ \cellcolor{blue!7!white}  & $79.79$  & $70.71$ \\ \midrule
        \rowcolor{blue!3!white}
        & & & & & & & & & & & & \cellcolor{blue!7!white} & & \\
        \rowcolor{blue!3!white}
        &PromptCCD~(Ours) & GMP (Ours)
        & $64.23$  & $89.29$  & $47.49$ 
        & $60.00$  & $86.43$  & $54.39$ 
        & $74.83$  & $87.14$  & $72.42$ 
        & $66.35$ \cellcolor{blue!7!white}  & $87.62$  & $58.10$ \\
        \rowcolor{blue!3!white}
        \rot{\rlap{CUB}}
        &PromptCCD++~(Ours) & PLP (Ours)
        & $70.67$  & $88.21$  & $58.95$ 
        & $75.52$  & $90.00$  & $72.26$ 
        & $79.37$  & $89.29$  & $77.44$ 
        & $\textbf{75.19}$ \cellcolor{blue!7!white}  & $89.17$  & $69.55$ \\
        \bottomrule
        \end{tabular}
    }
    \label{tab:main_complete_result_baselines_seed_10_dinov2}
\end{table*}
\definecolor{Gray}{gray}{0.9}
\definecolor{PaleBlue}{rgb}{0.7529, 0.9137, 0.9372}
\definecolor{BeauBlue}{rgb}{0.7686, 0.8470, 0.9529}
\definecolor{Mauve}{rgb}{0.8 , 0.7098, 0.9843}
\definecolor{PaleViolet}{rgb}{0.8156, 0.6431, 1.0}
\definecolor{Salmon}{rgb}{1.0, 0.8980, 0.6920}
\definecolor{Pink}{rgb}{1.0, 0.6902, 0.7908}
\definecolor{Mint}{rgb}{0.6902, 1.0, 0.7451}
\definecolor{cyan}{rgb}{0.906, 0.969, 0.965}
\definecolor{SoftP}{rgb}{0.945, 0.933, 0.949}

\begin{table*}[!ht]
    \caption{Breakdown results of our method with different prompt pool designs for CCD leveraging pretrained DINOv2 model on generic and CUB datasets with the \textit{known} $C$ in each unlabelled set. The experiments are conducted with \textit{seed 2000}.}
    \centering
    \resizebox{1.0\columnwidth}{!}{%
    \centering
        \begin{tabular}{clcccc @{\hskip 0.1in} ccc@{\hskip 0.1in}ccc @{\hskip 0.1in} ||ccc}
        \toprule
        \multicolumn{3}{c}{}
        & \multicolumn{3}{c}{Stage 1 \textit{ACC} (\%)} 
        & \multicolumn{3}{c}{Stage 2 \textit{ACC} (\%)} 
        & \multicolumn{3}{c}{Stage 3 \textit{ACC} (\%)} 
        & \multicolumn{3}{c}{Average \textit{ACC} (\%)} 
        \\ 
        \multicolumn{1}{l}{} &
        \multicolumn{1}{c}{Method} &
        \multicolumn{1}{c}{Prompt Pool} &
        \multicolumn{1}{c}{\textit{All}} & 
        \multicolumn{1}{c}{\textit{Old}} &
        \multicolumn{1}{c}{\textit{New}} &
        \multicolumn{1}{c}{\textit{All}} & 
        \multicolumn{1}{c}{\textit{Old}} & 
        \multicolumn{1}{c}{\textit{New}} &
        \multicolumn{1}{c}{\textit{All}} & 
        \multicolumn{1}{c}{\textit{Old}} & 
        \multicolumn{1}{c}{\textit{New}} &
        \multicolumn{1}{c}{\textit{All} \cellcolor{blue!7!white}} & 
        \multicolumn{1}{c}{\textit{Old}} & 
        \multicolumn{1}{c}{\textit{New}} 
        \\ 
        \cmidrule[0.1pt](r{0.80em}){1-2}%
        \cmidrule[0.1pt](r{0.80em}){3-3}%
        \cmidrule[0.1pt](r{0.80em}){4-6}%
        \cmidrule[0.1pt](r{0.80em}){7-9}%
        \cmidrule[0.1pt](r{0.80em}){10-12}%
        \cmidrule[0.1pt](r{0.80em}){13-15}%
        \rowcolor{blue!3!white}
        & & & & & & & & & & & & \cellcolor{blue!7!white} & & \\
        \rowcolor{blue!3!white}
        &PromptCCD~(Ours) & GMP (Ours)
        & $77.58$  & $88.61$  & $69.86$ 
        & $73.45$  & $78.19$  & $72.55$ 
        & $63.50$  & $66.38$  & $63.00$ 
        & $71.51$ \cellcolor{blue!7!white}  & $77.73$  & $68.47$ \\
        \rowcolor{blue!3!white}
        \rot{\rlap{C100}}
        &PromptCCD++~(Ours) & PLP (Ours)
        & $82.66$  & $88.04$  & $78.89$ 
        & $83.83$  & $84.76$  & $83.65$ 
        & $71.66$  & $81.71$  & $69.90$ 
        & $\textbf{79.38}$ \cellcolor{blue!7!white}  & $84.84$  & $77.48$ \\ \midrule
        \rowcolor{blue!3!white}
        & & & & & & & & & & & & \cellcolor{blue!7!white} & & \\
        \rowcolor{blue!3!white}
        &PromptCCD~(Ours) & GMP (Ours)
        & $80.54$  & $86.86$  & $76.11$ 
        & $71.82$  & $81.24$  & $70.02$ 
        & $71.89$  & $80.48$  & $70.38$ 
        & $74.75$ \cellcolor{blue!7!white}  & $82.86$  & $72.17$ \\
        \rowcolor{blue!3!white}
        \rot{\rlap{IN-100}}
        &PromptCCD++~(Ours) & PLP (Ours)
        & $81.70$  & $86.04$  & $78.66$ 
        & $80.35$  & $81.62$  & $80.11$ 
        & $82.14$  & $81.43$  & $82.27$ 
        & $\textbf{81.40}$ \cellcolor{blue!7!white}  & $83.03$  & $80.35$ \\ \midrule
        \rowcolor{blue!3!white}
        & & & & & & & & & & & & \cellcolor{blue!7!white} & & \\
        \rowcolor{blue!3!white}
        &PromptCCD~(Ours) & GMP (Ours)
        & $72.32$  & $83.69$  & $64.36$ 
        & $66.55$  & $75.48$  & $64.85$ 
        & $64.26$  & $69.29$  & $63.38$ 
        & $67.71$ \cellcolor{blue!7!white}  & $75.15$  & $64.20$ \\
        \rowcolor{blue!3!white}
        \rot{\rlap{Tiny}}
        &PromptCCD++~(Ours) & PLP (Ours)
        & $71.60$  & $81.39$  & $64.74$ 
        & $69.95$  & $80.14$  & $68.00$ 
        & $70.78$  & $78.43$  & $69.44$ 
        & $\textbf{70.78}$ \cellcolor{blue!7!white}  & $79.99$  & $67.39$ \\ \midrule
        \rowcolor{blue!3!white}
        & & & & & & & & & & & & \cellcolor{blue!7!white} & & \\
        \rowcolor{blue!3!white}
        &PromptCCD~(Ours) & GMP (Ours)
        & $72.53$  & $88.21$  & $62.05$ 
        & $68.87$  & $86.43$  & $65.15$ 
        & $71.45$  & $85.00$  & $68.80$ 
        & $70.95$ \cellcolor{blue!7!white}  & $86.55$  & $65.33$ \\
        \rowcolor{blue!3!white}
        \rot{\rlap{CUB}}
        &PromptCCD++~(Ours) & PLP (Ours)
        & $71.53$  & $87.50$  & $60.86$ 
        & $77.62$  & $91.43$  & $74.70$ 
        & $77.51$  & $90.00$  & $75.07$ 
        & $\textbf{75.55}$ \cellcolor{blue!7!white}  & $89.64$  & $70.21$ \\
        \bottomrule
        \end{tabular}
    }
    \label{tab:main_complete_result_baselines_seed_2000_dinov2}
\end{table*}
\definecolor{Gray}{gray}{0.9}
\definecolor{PaleBlue}{rgb}{0.7529, 0.9137, 0.9372}
\definecolor{BeauBlue}{rgb}{0.7686, 0.8470, 0.9529}
\definecolor{Mauve}{rgb}{0.8 , 0.7098, 0.9843}
\definecolor{PaleViolet}{rgb}{0.8156, 0.6431, 1.0}
\definecolor{Salmon}{rgb}{1.0, 0.8980, 0.6920}
\definecolor{Pink}{rgb}{1.0, 0.6902, 0.7908}
\definecolor{Mint}{rgb}{0.6902, 1.0, 0.7451}
\definecolor{cyan}{rgb}{0.906, 0.969, 0.965}
\definecolor{SoftP}{rgb}{0.945, 0.933, 0.949}

\begin{table*}[!ht]
    \caption{Breakdown results of our method with different prompt pool designs for CCD leveraging pretrained DINOv2 model on generic and CUB datasets with the \textit{known} $C$ in each unlabelled set. The experiments are conducted with \textit{seed 2024}.}
    \centering
    \resizebox{1.0\columnwidth}{!}{%
    \centering
        \begin{tabular}{clcccc @{\hskip 0.1in} ccc@{\hskip 0.1in}ccc @{\hskip 0.1in} ||ccc}
        \toprule
        \multicolumn{3}{c}{}
        & \multicolumn{3}{c}{Stage 1 \textit{ACC} (\%)} 
        & \multicolumn{3}{c}{Stage 2 \textit{ACC} (\%)} 
        & \multicolumn{3}{c}{Stage 3 \textit{ACC} (\%)} 
        & \multicolumn{3}{c}{Average \textit{ACC} (\%)} 
        \\ 
        \multicolumn{1}{l}{} &
        \multicolumn{1}{c}{Method} &
        \multicolumn{1}{c}{Prompt Pool} &
        \multicolumn{1}{c}{\textit{All}} & 
        \multicolumn{1}{c}{\textit{Old}} &
        \multicolumn{1}{c}{\textit{New}} &
        \multicolumn{1}{c}{\textit{All}} & 
        \multicolumn{1}{c}{\textit{Old}} & 
        \multicolumn{1}{c}{\textit{New}} &
        \multicolumn{1}{c}{\textit{All}} & 
        \multicolumn{1}{c}{\textit{Old}} & 
        \multicolumn{1}{c}{\textit{New}} &
        \multicolumn{1}{c}{\textit{All} \cellcolor{blue!7!white}} & 
        \multicolumn{1}{c}{\textit{Old}} & 
        \multicolumn{1}{c}{\textit{New}} 
        \\ 
        \cmidrule[0.1pt](r{0.80em}){1-2}%
        \cmidrule[0.1pt](r{0.80em}){3-3}%
        \cmidrule[0.1pt](r{0.80em}){4-6}%
        \cmidrule[0.1pt](r{0.80em}){7-9}%
        \cmidrule[0.1pt](r{0.80em}){10-12}%
        \cmidrule[0.1pt](r{0.80em}){13-15}%
        \rowcolor{blue!3!white}
        & & & & & & & & & & & & \cellcolor{blue!7!white} & & \\
        \rowcolor{blue!3!white}
        &PromptCCD~(Ours) & GMP (Ours)
        & $71.90$  & $88.94$  & $59.97$ 
        & $61.11$  & $75.90$  & $58.29$ 
        & $61.94$  & $69.24$  & $60.67$ 
        & $64.98$ \cellcolor{blue!7!white}  & $78.03$  & $59.64$ \\
        \rowcolor{blue!3!white}
        \rot{\rlap{C100}}
        &PromptCCD++~(Ours) & PLP (Ours)
        & $80.17$  & $85.59$  & $76.37$ 
        & $83.79$  & $84.38$  & $83.67$ 
        & $71.11$  & $81.62$  & $69.27$ 
        & $\textbf{78.36}$ \cellcolor{blue!7!white}  & $83.86$  & $76.44$ \\ \midrule
        \rowcolor{blue!3!white}
        & & & & & & & & & & & & \cellcolor{blue!7!white} & & \\
        \rowcolor{blue!3!white}
        &PromptCCD~(Ours) & GMP (Ours)
        & $81.18$  & $86.41$  & $77.51$ 
        & $76.75$  & $81.33$  & $75.87$ 
        & $74.37$  & $74.76$  & $74.30$ 
        & $77.43$ \cellcolor{blue!7!white}  & $80.83$  & $75.89$ \\
        \rowcolor{blue!3!white}
        \rot{\rlap{IN-100}}
        &PromptCCD++~(Ours) & PLP (Ours)
        & $81.33$  & $84.82$  & $78.89$ 
        & $82.98$  & $85.43$  & $82.51$ 
        & $81.33$  & $84.82$  & $78.89$ 
        & $\textbf{81.88}$ \cellcolor{blue!7!white}  & $85.02$  & $80.10$ \\ \midrule
        \rowcolor{blue!3!white}
        & & & & & & & & & & & & \cellcolor{blue!7!white} & & \\
        \rowcolor{blue!3!white}
        &PromptCCD~(Ours) & GMP (Ours)
        & $71.88$  & $83.02$  & $64.09$ 
        & $64.26$  & $73.95$  & $62.41$ 
        & $61.11$  & $67.38$  & $60.01$ 
        & $65.75$ \cellcolor{blue!7!white}  & $74.78$  & $62.17$ \\
        \rowcolor{blue!3!white}
        \rot{\rlap{Tiny}}
        &PromptCCD++~(Ours) & PLP (Ours)
        & $75.84$  & $81.14$  & $72.13$ 
        & $70.86$  & $77.38$  & $69.62$ 
        & $68.97$  & $78.05$  & $67.38$ 
        & $\textbf{71.89}$ \cellcolor{blue!7!white}  & $78.86$  & $69.71$ \\ \midrule
        \rowcolor{blue!3!white}
        & & & & & & & & & & & & \cellcolor{blue!7!white} & & \\
        \rowcolor{blue!3!white}
        &PromptCCD~(Ours) & GMP (Ours)
        & $66.09$  & $88.21$  & $51.31$ 
        & $61.62$  & $85.71$  & $56.52$ 
        & $68.65$  & $79.29$  & $66.57$ 
        & $65.45$ \cellcolor{blue!7!white}  & $84.40$  & $58.13$ \\
        \rowcolor{blue!3!white}
        \rot{\rlap{CUB}}
        &PromptCCD++~(Ours) & PLP (Ours)
        & $74.54$  & $87.86$  & $65.63$ 
        & $73.38$  & $95.71$  & $68.64$ 
        & $80.07$  & $87.86$  & $78.55$ 
        & $\textbf{76.00}$ \cellcolor{blue!7!white}  & $90.48$  & $70.94$ \\
        \bottomrule
        \end{tabular}
    }
    \label{tab:main_complete_result_baselines_seed_2024_dinov2}
\end{table*}

\definecolor{Gray}{gray}{0.9}
\definecolor{PaleBlue}{rgb}{0.7529, 0.9137, 0.9372}
\definecolor{BeauBlue}{rgb}{0.7686, 0.8470, 0.9529}
\definecolor{Mauve}{rgb}{0.8 , 0.7098, 0.9843}
\definecolor{PaleViolet}{rgb}{0.8156, 0.6431, 1.0}
\definecolor{Salmon}{rgb}{1.0, 0.8980, 0.6920}
\definecolor{Pink}{rgb}{1.0, 0.6902, 0.7908}
\definecolor{Mint}{rgb}{0.6902, 1.0, 0.7451}
\definecolor{cyan}{rgb}{0.906, 0.969, 0.965}
\definecolor{SoftP}{rgb}{0.945, 0.933, 0.949}

\setlength\dashlinedash{1.2pt}
\setlength\dashlinegap{0.5pt}
\setlength\arrayrulewidth{0.3pt}

\begin{table*}[htbp]
    \caption{Breakdown results of different methods for CCD leveraging pretrained DINO model on generic and CUB datasets with the \textit{unknown} $C$ in each unlabelled set. The $C$s are estimated using our method described in Sec.~$3.3$, in the main paper. The estimated $C$s are applied to all other methods for comparison.}
    \centering
    \resizebox{1.0\columnwidth}{!}{%
    \centering
        \begin{tabular}{clccccccccc ||ccc}
        \toprule
        \multicolumn{2}{c}{}
        & \multicolumn{3}{c}{Stage 1 \textit{ACC} (\%)}
        & \multicolumn{3}{c}{Stage 2 \textit{ACC} (\%)} 
        & \multicolumn{3}{c}{Stage 3 \textit{ACC} (\%)} 
        & \multicolumn{3}{c}{Average \textit{ACC} (\%)}
        \\ 
        \multicolumn{1}{l}{} &
        \multicolumn{1}{c}{Method} &
        \multicolumn{3}{c}{\textit{All} {\hskip 0.08in} \textit{Old} {\hskip 0.08in} \textit{New}} &
        \multicolumn{3}{c}{\textit{All} {\hskip 0.08in} \textit{Old} {\hskip 0.08in} \textit{New}} &
        \multicolumn{3}{c}{\textit{All} {\hskip 0.08in} \textit{Old} {\hskip 0.08in} \textit{New}} &
        \multicolumn{1}{c}{\textit{All} \cellcolor{blue!7!white}} & 
        \multicolumn{1}{c}{\textit{Old}} & 
        \multicolumn{1}{c}{\textit{New}} 
        \\ 
        \cmidrule[0.1pt](r{0.80em}){1-2}%
        \cmidrule[0.1pt](r{0.80em}){3-5}%
        \cmidrule[0.1pt](r{0.80em}){6-8}%
        \cmidrule[0.1pt](r{0.80em}){9-11}%
        \cmidrule[0.1pt](r{0.80em}){12-14}%
        & Estimated $C$
        & \multicolumn{3}{c}{$(C^{\textit{EST}}$: $85$, $C^{\textit{GT}}$: $80)$}
        & \multicolumn{3}{c}{$(C^{\textit{EST}}$: $100$, $C^{\textit{GT}}$: $90)$}
        & \multicolumn{3}{c}{$(C^{\textit{EST}}$: $115$, $C^{\textit{GT}}$: $100)$} 
        & \multicolumn{3}{c}{} \\ \cdashline{2-14} 
        \rowcolor{gray!5!white}
        & GCD \cite{vaze2022generalized} 
        &\multicolumn{3}{c}{$61.63$ {\hskip 0.02in} $83.10$ {\hskip 0.02in} $46.60$}
        &\multicolumn{3}{c}{$50.29$ {\hskip 0.02in} $71.81$ {\hskip 0.02in} $46.18$} 
        &\multicolumn{3}{c}{$49.43$ {\hskip 0.02in} $67.24$ {\hskip 0.02in} $46.32$} 
        & $53.78$ \cellcolor{blue!7!white} & $74.05$ & $46.37$ \\
        \rowcolor{gray!5!white}
        &Grow \& Merge \cite{zhang2022grow} 
        &\multicolumn{3}{c}{$63.03$ {\hskip 0.02in} $70.69$ {\hskip 0.02in} $57.66$}
        &\multicolumn{3}{c}{$56.52$ {\hskip 0.02in} $66.76$ {\hskip 0.02in} $54.56$} 
        &\multicolumn{3}{c}{$40.45$ {\hskip 0.02in} $62.48$ {\hskip 0.02in} $36.60$} 
        & $53.33$ \cellcolor{blue!7!white} & $66.64$ & $49.61$ \\
        \rowcolor{gray!5!white}
        &MetaGCD \cite{wu2023metagcd} 
        &\multicolumn{3}{c}{$54.22$ {\hskip 0.02in} $81.31$ {\hskip 0.02in} $35.26$}
        &\multicolumn{3}{c}{$34.95$ {\hskip 0.02in} $68.00$ {\hskip 0.02in} $28.64$} 
        &\multicolumn{3}{c}{$53.49$ {\hskip 0.02in} $63.05$ {\hskip 0.02in} $51.82$} 
        & $47.55$ \cellcolor{blue!7!white} & $70.79$ & $38.57$ \\
        \rowcolor{gray!5!white}
        &PA-CGCD \cite{kim2023proxy} 
        &\multicolumn{3}{c}{$54.77$ {\hskip 0.02in} $80.24$ {\hskip 0.02in} $36.94$}
        &\multicolumn{3}{c}{$62.20$ {\hskip 0.02in} $94.29$ {\hskip 0.02in} $56.07$} 
        &\multicolumn{3}{c}{$50.01$ {\hskip 0.02in} $96.10$ {\hskip 0.02in} $41.95$} 
        & $55.66$ \cellcolor{blue!7!white} & $90.21$ & $44.99$ \\
        \rowcolor{blue!3!white}
        \rot{\rlap{C100}}
        &PromptCCD-U (Ours) 
        &\multicolumn{3}{c}{$70.64$ {\hskip 0.02in} $82.49$ {\hskip 0.02in} $62.34$}
        &\multicolumn{3}{c}{$58.46$ {\hskip 0.02in} $78.48$ {\hskip 0.02in} $54.64$} 
        &\multicolumn{3}{c}{$48.27$ {\hskip 0.02in} $71.90$ {\hskip 0.02in} $44.13$} 
        & $\textbf{59.12}$ \cellcolor{blue!7!white} & $77.62$ & $53.70$ \\ \midrule
        & Estimated $C$
        & \multicolumn{3}{c}{$(C^{\textit{EST}}$: $83$, $C^{\textit{GT}}: 80)$}
        & \multicolumn{3}{c}{$(C^{\textit{EST}}$: $98$, $C^{\textit{GT}}: 90)$}
        & \multicolumn{3}{c}{$(C^{\textit{EST}}$: $113$, $C^{\textit{GT}}: 100)$} 
        & \multicolumn{3}{c}{} \\ \cdashline{2-14}
        \rowcolor{gray!5!white}
        & GCD \cite{vaze2022generalized} 
        &\multicolumn{3}{c}{$72.15$ {\hskip 0.02in} $84.49$ {\hskip 0.02in} $63.51$}
        &\multicolumn{3}{c}{$70.62$ {\hskip 0.02in} $81.19$ {\hskip 0.02in} $68.60$} 
        &\multicolumn{3}{c}{$62.87$ {\hskip 0.02in} $80.48$ {\hskip 0.02in} $59.78$} 
        & $68.55$ \cellcolor{blue!7!white} & $82.05$ & $63.96$ \\
        \rowcolor{gray!5!white}
        &Grow \& Merge \cite{zhang2022grow} 
        &\multicolumn{3}{c}{$68.04$ {\hskip 0.02in} $73.29$ {\hskip 0.02in} $64.37$}
        &\multicolumn{3}{c}{$68.02$ {\hskip 0.02in} $76.95$ {\hskip 0.02in} $66.31$} 
        &\multicolumn{3}{c}{$63.13$ {\hskip 0.02in} $73.33$ {\hskip 0.02in} $61.35$} 
        & $66.40$ \cellcolor{blue!7!white} & $74.52$ & $64.01$ \\
        \rowcolor{gray!5!white}
        &MetaGCD \cite{wu2023metagcd} 
        &\multicolumn{3}{c}{$60.62$ {\hskip 0.02in} $83.88$ {\hskip 0.02in} $44.34$}
        &\multicolumn{3}{c}{$63.83$ {\hskip 0.02in} $82.48$ {\hskip 0.02in} $60.27$} 
        &\multicolumn{3}{c}{$66.00$ {\hskip 0.02in} $76.10$ {\hskip 0.02in} $64.23$} 
        & $63.48$ \cellcolor{blue!7!white} & $80.82$ & $56.28$ \\
        \rowcolor{gray!5!white}
        &PA-CGCD \cite{kim2023proxy} 
        &\multicolumn{3}{c}{$63.53$ {\hskip 0.02in} $82.61$ {\hskip 0.02in} $50.17$}
        &\multicolumn{3}{c}{$71.63$ {\hskip 0.02in} $97.24$ {\hskip 0.02in} $66.75$} 
        &\multicolumn{3}{c}{$65.06$ {\hskip 0.02in} $94.00$ {\hskip 0.02in} $60.00$} 
        & $66.74$ \cellcolor{blue!7!white} & $91.28$ & $58.97$ \\
        \rowcolor{blue!3!white}
        \rot{\rlap{IN-100}}
        &PromptCCD-U (Ours) 
        &\multicolumn{3}{c}{$73.24$ {\hskip 0.02in} $84.20$ {\hskip 0.02in} $65.57$}
        &\multicolumn{3}{c}{$75.07$ {\hskip 0.02in} $81.14$ {\hskip 0.02in} $73.91$} 
        &\multicolumn{3}{c}{$62.06$ {\hskip 0.02in} $80.19$ {\hskip 0.02in} $58.88$} 
        & $\textbf{70.12}$ \cellcolor{blue!7!white} & $81.84$ & $66.12$ \\ \midrule
        & Estimated $C$
        & \multicolumn{3}{c}{$(C^{\textit{EST}}$: $155$, $C^{\textit{GT}}: 160)$}
        & \multicolumn{3}{c}{$(C^{\textit{EST}}$: $170$, $C^{\textit{GT}}: 180)$}
        & \multicolumn{3}{c}{$(C^{\textit{EST}}$: $185$, $C^{\textit{GT}}: 200)$} 
        & \multicolumn{3}{c}{} \\ \cdashline{2-14}
        \rowcolor{gray!5!white}
        & GCD \cite{vaze2022generalized} 
        &\multicolumn{3}{c}{$63.97$ {\hskip 0.02in} $71.73$ {\hskip 0.02in} $58.54$}
        &\multicolumn{3}{c}{$52.17$ {\hskip 0.02in} $64.95$ {\hskip 0.02in} $49.73$} 
        &\multicolumn{3}{c}{$49.70$ {\hskip 0.02in} $58.43$ {\hskip 0.02in} $48.17$} 
        & $55.28$ \cellcolor{blue!7!white} & $65.04$ & $52.15$ \\
        \rowcolor{gray!5!white}
        &Grow \& Merge \cite{zhang2022grow} 
        &\multicolumn{3}{c}{$60.61$ {\hskip 0.02in} $62.94$ {\hskip 0.02in} $58.99$}
        &\multicolumn{3}{c}{$47.96$ {\hskip 0.02in} $55.81$ {\hskip 0.02in} $46.46$} 
        &\multicolumn{3}{c}{$48.64$ {\hskip 0.02in} $54.86$ {\hskip 0.02in} $47.55$} 
        & $52.40$ \cellcolor{blue!7!white} & $57.87$ & $51.00$ \\
        \rowcolor{gray!5!white}
        &MetaGCD \cite{wu2023metagcd} 
        &\multicolumn{3}{c}{$58.80$ {\hskip 0.02in} $73.41$ {\hskip 0.02in} $48.57$}
        &\multicolumn{3}{c}{$52.90$ {\hskip 0.02in} $73.21$ {\hskip 0.02in} $47.73$} 
        &\multicolumn{3}{c}{$56.93$ {\hskip 0.02in} $58.38$ {\hskip 0.02in} $56.67$} 
        & $56.21$ \cellcolor{blue!7!white} & $68.33$ & $50.99$ \\
        \rowcolor{gray!5!white}
        &PA-CGCD \cite{kim2023proxy} 
        &\multicolumn{3}{c}{$54.24$ {\hskip 0.02in} $72.69$ {\hskip 0.02in} $41.33$}
        &\multicolumn{3}{c}{$43.39$ {\hskip 0.02in} $63.86$ {\hskip 0.02in} $39.48$} 
        &\multicolumn{3}{c}{$54.02$ {\hskip 0.02in} $80.76$ {\hskip 0.02in} $49.34$} 
        & $50.55$ \cellcolor{blue!7!white} & $72.44$ & $43.38$ \\
        \rowcolor{blue!3!white}
        \rot{\rlap{Tiny}}
        &PromptCCD-U (Ours) 
        &\multicolumn{3}{c}{$65.61$ {\hskip 0.02in} $72.14$ {\hskip 0.02in} $61.40$}
        &\multicolumn{3}{c}{$52.05$ {\hskip 0.02in} $63.81$ {\hskip 0.02in} $49.81$} 
        &\multicolumn{3}{c}{$55.62$ {\hskip 0.02in} $57.76$ {\hskip 0.02in} $55.25$} 
        & $\textbf{57.76}$ \cellcolor{blue!7!white} & $64.57$ & $55.37$ \\ \midrule
        & Estimated $C$
        & \multicolumn{3}{c}{$(C^{\textit{EST}}$: $161$, $C^{\textit{GT}}$: $160)$}
        & \multicolumn{3}{c}{$(C^{\textit{EST}}$: $180$, $C^{\textit{GT}}$: $180)$}
        & \multicolumn{3}{c}{$(C^{\textit{EST}}$: $198$, $C^{\textit{GT}}$: $200)$} 
        & \multicolumn{3}{c}{} \\ \cdashline{2-14}
        \rowcolor{gray!5!white}
        & GCD \cite{vaze2022generalized} 
        &\multicolumn{3}{c}{$55.94$ {\hskip 0.02in} $75.86$ {\hskip 0.02in} $42.63$}
        &\multicolumn{3}{c}{$44.62$ {\hskip 0.02in} $70.71$ {\hskip 0.02in} $39.09$} 
        &\multicolumn{3}{c}{$51.52$ {\hskip 0.02in} $70.71$ {\hskip 0.02in} $47.77$} 
        & $50.69$ \cellcolor{blue!7!white} & $72.43$ & $43.16$ \\
        \rowcolor{gray!5!white}
        &Grow \& Merge \cite{zhang2022grow} 
        &\multicolumn{3}{c}{$42.49$ {\hskip 0.02in} $62.93$ {\hskip 0.02in} $28.83$}
        &\multicolumn{3}{c}{$29.50$ {\hskip 0.02in} $62.57$ {\hskip 0.02in} $22.48$} 
        &\multicolumn{3}{c}{$42.36$ {\hskip 0.02in} $61.14$ {\hskip 0.02in} $38.69$} 
        & $38.12$ \cellcolor{blue!7!white} & $62.21$ & $30.00$ \\
        \rowcolor{gray!5!white}
        &MetaGCD \cite{wu2023metagcd} 
        &\multicolumn{3}{c}{$47.36$ {\hskip 0.02in} $68.79$ {\hskip 0.02in} $33.71$}
        &\multicolumn{3}{c}{$39.00$ {\hskip 0.02in} $67.71$ {\hskip 0.02in} $32.91$} 
        &\multicolumn{3}{c}{$46.53$ {\hskip 0.02in} $75.57$ {\hskip 0.02in} $40.87$} 
        & $44.30$ \cellcolor{blue!7!white} & $70.69$ & $35.83$ \\
        \rowcolor{gray!5!white}
        &PA-CGCD \cite{kim2023proxy} 
        &\multicolumn{3}{c}{$54.65$ {\hskip 0.02in} $74.71$ {\hskip 0.02in} $41.24$}
        &\multicolumn{3}{c}{$46.75$ {\hskip 0.02in} $76.14$ {\hskip 0.02in} $40.52$} 
        &\multicolumn{3}{c}{$55.41$ {\hskip 0.02in} $78.29$ {\hskip 0.02in} $50.95$} 
        & $52.27$ \cellcolor{blue!7!white} & $76.38$ & $44.24$ \\
        \rowcolor{blue!3!white}
        \rot{\rlap{CUB}}
        &PromptCCD-U (Ours) 
        &\multicolumn{3}{c}{$57.23$ {\hskip 0.02in} $74.43$ {\hskip 0.02in} $45.73$}
        &\multicolumn{3}{c}{$46.50$ {\hskip 0.02in} $74.29$ {\hskip 0.02in} $40.61$} 
        &\multicolumn{3}{c}{$61.87$ {\hskip 0.02in} $70.86$ {\hskip 0.02in} $60.12$} 
        & $\textbf{55.20}$ \cellcolor{blue!7!white} & $73.19$ & $48.82$ \\
        \bottomrule
        \end{tabular}
    }
    \label{tab:main_result_with_unknown_k_gpc}
\end{table*}
\definecolor{Gray}{gray}{0.9}
\definecolor{PaleBlue}{rgb}{0.7529, 0.9137, 0.9372}
\definecolor{BeauBlue}{rgb}{0.7686, 0.8470, 0.9529}
\definecolor{Mauve}{rgb}{0.8 , 0.7098, 0.9843}
\definecolor{PaleViolet}{rgb}{0.8156, 0.6431, 1.0}
\definecolor{Salmon}{rgb}{1.0, 0.8980, 0.6920}
\definecolor{Pink}{rgb}{1.0, 0.6902, 0.7908}
\definecolor{Mint}{rgb}{0.6902, 1.0, 0.7451}
\definecolor{SoftP}{rgb}{0.945, 0.933, 0.949}

\setlength\dashlinedash{1.2pt}
\setlength\dashlinegap{0.5pt}
\setlength\arrayrulewidth{0.3pt}

\begin{table*}[!ht]
    \caption{Breakdown results of different methods for CCD leveraging pretrained DINO model on generic and CUB datasets with the \textit{unknown} $C$ in each unlabelled set. The $C$s are estimated using the $k$-means-based estimator in \cite{vaze2022generalized}. The estimated $C$s are applied to all other methods for comparison.}
    \centering
    \resizebox{1.0\columnwidth}{!}{%
    \centering
        \begin{tabular}{cl ccc ccc ccc||ccc}
        \toprule
        \multicolumn{2}{c}{}
        & \multicolumn{3}{c}{Stage 1 \textit{ACC} (\%)} 
        & \multicolumn{3}{c}{Stage 2 \textit{ACC} (\%)} 
        & \multicolumn{3}{c}{Stage 3 \textit{ACC} (\%)} 
        & \multicolumn{3}{c}{Average \textit{ACC} (\%)} 
        \\ 
        \multicolumn{1}{l}{} &
        \multicolumn{1}{c}{Method} &
        \multicolumn{3}{c}{\textit{All} {\hskip 0.08in} \textit{Old} {\hskip 0.08in} \textit{New}} &
        \multicolumn{3}{c}{\textit{All} {\hskip 0.08in} \textit{Old} {\hskip 0.08in} \textit{New}} &
        \multicolumn{3}{c}{\textit{All} {\hskip 0.08in} \textit{Old} {\hskip 0.08in} \textit{New}} &
        \multicolumn{1}{c}{\textit{All} \cellcolor{blue!7!white} } & 
        \multicolumn{1}{c}{\textit{Old}} & 
        \multicolumn{1}{c}{\textit{New}} 
        \\
        \cmidrule[0.1pt](r{0.80em}){1-2}%
        \cmidrule[0.1pt](r{0.80em}){3-5}%
        \cmidrule[0.1pt](r{0.80em}){6-8}%
        \cmidrule[0.1pt](r{0.80em}){9-11}%
        \cmidrule[0.1pt](r{0.80em}){12-14}%
        & Estimated $C$
        & \multicolumn{3}{c}{$(C^{\textit{EST}}$: $84$, $C^{\textit{GT}}$: $80)$}
        & \multicolumn{3}{c}{$(C^{\textit{EST}}$: $84$, $C^{\textit{GT}}$: $90)$}
        & \multicolumn{3}{c}{$(C^{\textit{EST}}$: $84$, $C^{\textit{GT}}$: $100)$} 
        & \multicolumn{3}{c}{} \\ \cdashline{2-14} 
        \rowcolor{gray!5!white}
        & GCD \cite{vaze2022generalized} 
        &\multicolumn{3}{c}{$62.45$ {\hskip 0.02in} $83.27$ {\hskip 0.02in} $47.89$}
        &\multicolumn{3}{c}{$52.23$ {\hskip 0.02in} $63.71$ {\hskip 0.02in} $50.04$}
        &\multicolumn{3}{c}{$49.32$ {\hskip 0.02in} $58.86$ {\hskip 0.02in} $47.65$}
        & $54.67$\cellcolor{blue!7!white} & $68.61$ & $48.53$ \\
        \rowcolor{gray!5!white}
        &Grow \& Merge \cite{zhang2022grow} 
        &\multicolumn{3}{c}{$63.45$ {\hskip 0.02in} $72.29$ {\hskip 0.02in} $57.26$}
        &\multicolumn{3}{c}{$57.51$ {\hskip 0.02in} $57.52$ {\hskip 0.02in} $57.51$}
        &\multicolumn{3}{c}{$54.90$ {\hskip 0.02in} $51.05$ {\hskip 0.02in} $55.10$}
        & $58.62$\cellcolor{blue!7!white} & $60.29$ & $56.62$ \\
        \rowcolor{gray!5!white}
        & MetaGCD \cite{wu2023metagcd} 
        &\multicolumn{3}{c}{$54.59$ {\hskip 0.02in} $81.18$ {\hskip 0.02in} $35.97$}
        &\multicolumn{3}{c}{$40.85$ {\hskip 0.02in} $62.00$ {\hskip 0.02in} $36.82$}
        &\multicolumn{3}{c}{$61.91$ {\hskip 0.02in} $57.52$ {\hskip 0.02in} $62.68$} 
        & $52.45$\cellcolor{blue!7!white} & $66.90$ & $45.16$ \\
        \rowcolor{gray!5!white}
        & PA-CGCD \cite{kim2023proxy} 
        &\multicolumn{3}{c}{$55.29$ {\hskip 0.02in} $82.49$ {\hskip 0.02in} $36.26$}
        &\multicolumn{3}{c}{$59.33$ {\hskip 0.02in} $88.29$ {\hskip 0.02in} $53.80$}
        &\multicolumn{3}{c}{$53.73$ {\hskip 0.02in} $82.76$ {\hskip 0.02in} $48.65$}
        & $56.12$\cellcolor{blue!7!white} & $84.51$ & $46.24$ \\
        \rowcolor{blue!3!white}
        \rot{\rlap{C100}}
        &PromptCCD (Ours) 
        &\multicolumn{3}{c}{$69.38$ {\hskip 0.02in} $82.78$ {\hskip 0.02in} $60.00$}
        &\multicolumn{3}{c}{$64.55$ {\hskip 0.02in} $70.67$ {\hskip 0.02in} $63.38$}
        &\multicolumn{3}{c}{$59.23$ {\hskip 0.02in} $64.76$ {\hskip 0.02in} $58.27$}
        & $\textbf{64.39}$\cellcolor{blue!7!white} & $72.74$ & $60.55$ \\ \midrule
        & Estimated $C$ 
        & \multicolumn{3}{c}{$(C^{\textit{EST}}$: $90$, $C^{\textit{GT}}$: $80)$}
        & \multicolumn{3}{c}{$(C^{\textit{EST}}$: $90$, $C^{\textit{GT}}$: $90)$}
        & \multicolumn{3}{c}{$(C^{\textit{EST}}$: $91$, $C^{\textit{GT}}$: $100)$} 
        & \multicolumn{3}{c}{} \\ \cdashline{2-14}
        \rowcolor{gray!5!white}
        &GCD \cite{vaze2022generalized} 
        &\multicolumn{3}{c}{$64.66$ {\hskip 0.02in} $84.53$ {\hskip 0.02in} $50.74$}
        &\multicolumn{3}{c}{$71.54$ {\hskip 0.02in} $76.57$ {\hskip 0.02in} $70.58$}
        &\multicolumn{3}{c}{$56.67$ {\hskip 0.02in} $74.67$ {\hskip 0.02in} $53.52$}
        & $64.29$\cellcolor{blue!7!white} & $78.59$ & $58.28$ \\
        \rowcolor{gray!5!white}
        &Grow \& Merge \cite{zhang2022grow} 
        &\multicolumn{3}{c}{$65.19$ {\hskip 0.02in} $76.29$ {\hskip 0.02in} $57.43$}
        &\multicolumn{3}{c}{$58.56$ {\hskip 0.02in} $70.95$ {\hskip 0.02in} $56.20$}
        &\multicolumn{3}{c}{$55.06$ {\hskip 0.02in} $71.71$ {\hskip 0.02in} $52.15$}
        & $59.60$\cellcolor{blue!7!white} & $72.98$ & $55.26$ \\
        \rowcolor{gray!5!white}
        & MetaGCD \cite{wu2023metagcd} 
        &\multicolumn{3}{c}{$55.58$ {\hskip 0.02in} $83.88$ {\hskip 0.02in} $35.77$}
        &\multicolumn{3}{c}{$45.76$ {\hskip 0.02in} $77.90$ {\hskip 0.02in} $39.62$}
        &\multicolumn{3}{c}{$60.00$ {\hskip 0.02in} $73.43$ {\hskip 0.02in} $57.65$} 
        & $53.78$\cellcolor{blue!7!white} & $78.40$ & $44.35$ \\
        \rowcolor{gray!5!white}
        & PA-CGCD \cite{kim2023proxy} 
        &\multicolumn{3}{c}{$58.03$ {\hskip 0.02in} $82.49$ {\hskip 0.02in} $40.91$}
        &\multicolumn{3}{c}{$51.45$ {\hskip 0.02in} $89.62$ {\hskip 0.02in} $44.16$}
        &\multicolumn{3}{c}{$51.02$ {\hskip 0.02in} $86.57$ {\hskip 0.02in} $44.80$}
        & $53.50$\cellcolor{blue!7!white} & $86.23$ & $43.29$ \\
        \rowcolor{blue!3!white}
        \rot{\rlap{IN-100}}
        &PromptCCD (Ours) 
        &\multicolumn{3}{c}{$70.77$ {\hskip 0.02in} $84.12$ {\hskip 0.02in} $61.43$}
        &\multicolumn{3}{c}{$72.03$ {\hskip 0.02in} $75.71$ {\hskip 0.02in} $71.33$}
        &\multicolumn{3}{c}{$59.90$ {\hskip 0.02in} $77.71$ {\hskip 0.02in} $56.78$}
        & $\textbf{67.57}$\cellcolor{blue!7!white} & $79.18$ & $63.18$ \\ \midrule
        & Estimated $C$ 
        & \multicolumn{3}{c}{$(C^{\textit{EST}}$: $169$, $C^{\textit{GT}}$: $160)$}
        & \multicolumn{3}{c}{$(C^{\textit{EST}}$: $169$, $C^{\textit{GT}}$: $180)$}
        & \multicolumn{3}{c}{$(C^{\textit{EST}}$: $172$, $C^{\textit{GT}}$: $200)$} 
        & \multicolumn{3}{c}{} \\ \cdashline{2-14}
        \rowcolor{gray!5!white}
        &GCD \cite{vaze2022generalized} 
        &\multicolumn{3}{c}{$65.45$ {\hskip 0.02in} $73.06$ {\hskip 0.02in} $60.11$}
        &\multicolumn{3}{c}{$50.78$ {\hskip 0.02in} $61.05$ {\hskip 0.02in} $48.82$}
        &\multicolumn{3}{c}{$47.21$ {\hskip 0.02in} $55.52$ {\hskip 0.02in} $45.75$}
        & $54.48$\cellcolor{blue!7!white} & $63.21$ & $51.56$ \\
        \rowcolor{gray!5!white}
        &Grow \& Merge \cite{zhang2022grow} 
        &\multicolumn{3}{c}{$56.78$ {\hskip 0.02in} $64.12$ {\hskip 0.02in} $51.64$}
        &\multicolumn{3}{c}{$49.03$ {\hskip 0.02in} $54.14$ {\hskip 0.02in} $48.05$}
        &\multicolumn{3}{c}{$52.39$ {\hskip 0.02in} $53.48$ {\hskip 0.02in} $52.20$}
        & $52.73$\cellcolor{blue!7!white} & $57.25$ & $50.63$ \\
        \rowcolor{gray!5!white}
        & MetaGCD \cite{wu2023metagcd} 
        &\multicolumn{3}{c}{$59.76$ {\hskip 0.02in} $74.22$ {\hskip 0.02in} $49.63$}
        &\multicolumn{3}{c}{$58.46$ {\hskip 0.02in} $58.38$ {\hskip 0.02in} $58.47$}
        &\multicolumn{3}{c}{$60.34$ {\hskip 0.02in} $56.95$ {\hskip 0.02in} $60.93$}
        & $59.52$\cellcolor{blue!7!white} & $63.18$ & $56.34$ \\
        \rowcolor{gray!5!white}
        & PA-CGCD \cite{kim2023proxy} 
        &\multicolumn{3}{c}{$53.80$ {\hskip 0.02in} $74.65$ {\hskip 0.02in} $39.20$}
        &\multicolumn{3}{c}{$41.19$ {\hskip 0.02in} $62.76$ {\hskip 0.02in} $37.07$}
        &\multicolumn{3}{c}{$51.72$ {\hskip 0.02in} $76.33$ {\hskip 0.02in} $47.42$}
        & $48.90$\cellcolor{blue!7!white} & $71.25$ & $41.23$ \\
        \rowcolor{blue!3!white}
        \rot{\rlap{Tiny}}
        &PromptCCD (Ours) 
        &\multicolumn{3}{c}{$66.73$ {\hskip 0.02in} $79.76$ {\hskip 0.02in} $57.61$}
        &\multicolumn{3}{c}{$57.49$ {\hskip 0.02in} $68.38$ {\hskip 0.02in} $55.41$}
        &\multicolumn{3}{c}{$56.35$ {\hskip 0.02in} $65.57$ {\hskip 0.02in} $54.74$}
        & $\textbf{60.19}$\cellcolor{blue!7!white} & $71.24$ & $55.92$ \\ \midrule
        & Estimated $C$ 
        & \multicolumn{3}{c}{$(C^{\textit{EST}}$: $166$, $C^{\textit{GT}}$: $160)$}
        & \multicolumn{3}{c}{$(C^{\textit{EST}}$: $192$, $C^{\textit{GT}}$: $180)$}
        & \multicolumn{3}{c}{$(C^{\textit{EST}}$: $220$, $C^{\textit{GT}}$: $200)$} 
        & \multicolumn{3}{c}{} \\ \cdashline{2-14}
        \rowcolor{gray!5!white}
        &GCD \cite{vaze2022generalized} 
        &\multicolumn{3}{c}{$58.51$ {\hskip 0.02in} $77.50$ {\hskip 0.02in} $45.82$}
        &\multicolumn{3}{c}{$51.00$ {\hskip 0.02in} $75.71$ {\hskip 0.02in} $45.76$}
        &\multicolumn{3}{c}{$53.03$ {\hskip 0.02in} $78.57$ {\hskip 0.02in} $48.05$}
        & $54.18$\cellcolor{blue!7!white} & $77.26$ & $46.54$ \\
        \rowcolor{gray!5!white}
        &Grow \& Merge \cite{zhang2022grow} 
        &\multicolumn{3}{c}{$43.20$ {\hskip 0.02in} $62.50$ {\hskip 0.02in} $30.31$}
        &\multicolumn{3}{c}{$31.62$ {\hskip 0.02in} $67.14$ {\hskip 0.02in} $24.09$}
        &\multicolumn{3}{c}{$43.12$ {\hskip 0.02in} $65.71$ {\hskip 0.02in} $38.72$}
        & $39.31$\cellcolor{blue!7!white} & $65.12$ & $31.04$ \\
        \rowcolor{gray!5!white}
        & MetaGCD \cite{wu2023metagcd} 
        &\multicolumn{3}{c}{$52.50$ {\hskip 0.02in} $71.07$ {\hskip 0.02in} $40.10$}
        &\multicolumn{3}{c}{$43.50$ {\hskip 0.02in} $77.14$ {\hskip 0.02in} $36.36$}
        &\multicolumn{3}{c}{$46.62$ {\hskip 0.02in} $70.00$ {\hskip 0.02in} $42.06$}
        & $47.54$\cellcolor{blue!7!white} & $72.74$ & $39.51$ \\
        \rowcolor{gray!5!white}
        & PA-CGCD \cite{kim2023proxy} 
        &\multicolumn{3}{c}{$58.23$ {\hskip 0.02in} $74.29$ {\hskip 0.02in} $47.49$}
        &\multicolumn{3}{c}{$51.50$ {\hskip 0.02in} $78.57$ {\hskip 0.02in} $45.76$}
        &\multicolumn{3}{c}{$56.06$ {\hskip 0.02in} $77.14$ {\hskip 0.02in} $51.95$}
        & $55.26$\cellcolor{blue!7!white} & $76.67$ & $48.40$ \\
        \rowcolor{blue!3!white}
        \rot{\rlap{CUB}}
        &PromptCCD (Ours) 
        &\multicolumn{3}{c}{$59.94$ {\hskip 0.02in} $80.00$ {\hskip 0.02in} $46.54$}
        &\multicolumn{3}{c}{$52.50$ {\hskip 0.02in} $78.57$ {\hskip 0.02in} $46.97$}
        &\multicolumn{3}{c}{$54.20$ {\hskip 0.02in} $76.43$ {\hskip 0.02in} $49.86$}
        & $\textbf{55.55}$\cellcolor{blue!7!white} & $78.33$ & $47.79$ \\
        \bottomrule
        \end{tabular}
    }
    \label{tab:main_result_with_unknown_k}
\end{table*}
\clearpage
\section{Transductive and Inductive Evaluation}
\label{supp: trans and induc}

In our main paper, we evaluate our method on the unlabelled data, which are from the train splits of the original datasets. Indeed, the model has seen the data during training, though no labels are used. Here, we further evaluate our method on the test splits of the original datasets, which were not seen by the model during training. 
In other words, we consider two evaluation protocols, namely, \textit{transductive evaluation} and \textit{inductive evaluation}. In \textit{transductive evaluation}, the model is evaluated on the unlabelled data that has been seen by the model during training, while in \textit{inductive evaluation}, the model is evaluated on the unlabelled data that has not been seen by the model during training.

Since we have reported the transductive evaluation results in the main paper, here, we further include the inductive evaluation results in Tab.~\ref{tab:supp-3-main_result}, based on the \textit{cACC} evaluation metric introduced in the main paper. 
Overall, we can see that our method is more robust to unseen data compared to other models as it consistently performs better in the \textit{`All'} and \textit{`New'} accuracy.
\definecolor{Gray}{gray}{0.9}
\definecolor{PaleBlue}{rgb}{0.7529, 0.9137, 0.9372}
\definecolor{BeauBlue}{rgb}{0.7686, 0.8470, 0.9529}
\definecolor{Mauve}{rgb}{0.8 , 0.7098, 0.9843}
\definecolor{PaleViolet}{rgb}{0.8156, 0.6431, 1.0}
\definecolor{Salmon}{rgb}{1.0, 0.8980, 0.6920}
\definecolor{Pink}{rgb}{1.0, 0.6902, 0.7908}
\definecolor{cyan}{rgb}{0.906, 0.969, 0.965}
\definecolor{Mint}{rgb}{0.6902, 1.0, 0.7451}
\definecolor{SoftP}{rgb}{0.945, 0.933, 0.949}

\begin{table*}[htbp]
\caption{Comparison using the \textit{cACC} evaluation metric under the \textit{inductive} protocol.}
    \resizebox{\columnwidth}{!}{%
    \centering
        \begin{tabular}{cl @{\hskip 0.1in} ccc @{\hskip 0.1in} ccc @{\hskip 0.1in} ccc @{\hskip 0.1in} ||ccc}
        \toprule
        \multicolumn{2}{c}{}
        & \multicolumn{3}{c}{Stage 1 \textit{ACC} (\%)}
        & \multicolumn{3}{c}{Stage 2 \textit{ACC} (\%)} 
        & \multicolumn{3}{c}{Stage 3 \textit{ACC} (\%)} 
        & \multicolumn{3}{c}{Average \textit{ACC} (\%)} 
        \\ 
        \multicolumn{1}{l}{} &
        \multicolumn{1}{c}{Method} &
        \multicolumn{1}{c}{\textit{All}} & 
        \multicolumn{1}{c}{\textit{Old}} &
        \multicolumn{1}{c}{\textit{New}} &
        \multicolumn{1}{c}{\textit{All}} & 
        \multicolumn{1}{c}{\textit{Old}} & 
        \multicolumn{1}{c}{\textit{New}} &
        \multicolumn{1}{c}{\textit{All}} & 
        \multicolumn{1}{c}{\textit{Old}} & 
        \multicolumn{1}{c}{\textit{New}} &
        \multicolumn{1}{c}{\textit{All} \cellcolor{blue!7!white}} & 
        \multicolumn{1}{c}{\textit{Old}} & 
        \multicolumn{1}{c}{\textit{New}} 
        \\ 
        \cmidrule[0.1pt](r{0.80em}){1-2}%
        \cmidrule[0.1pt](r{0.80em}){3-5}%
        \cmidrule[0.1pt](r{0.80em}){6-8}%
        \cmidrule[0.1pt](r{0.80em}){9-11}%
        \cmidrule[0.1pt](r{0.80em}){12-14}%
        \rowcolor{gray!5!white}
        &GCD \cite{vaze2022generalized} 
        & $64.37$ & $86.33$ & $53.29$ 
        & $53.89$ & $70.00$  & $50.82$
        & $51.21$ & $65.71$  & $48.67$ 
        & $56.49$ \cellcolor{blue!7!white} & $74.01$ & $50.93$ \\
        \rowcolor{gray!5!white}
        &Grow \& Merge \cite{zhang2022grow} 
        & $56.05$ & $66.53$ & $48.71$ 
        & $58.17$ & $65.71$  & $56.73$
        & $53.33$ & $63.33$  & $51.58$ 
        & $55.85$ \cellcolor{blue!7!white} & $65.19$ & $52.34$ \\
        \rowcolor{gray!5!white}
        &MetaGCD \cite{wu2023metagcd}
        & $56.47$ & $78.37$ & $41.14$ 
        & $41.22$ & $67.14$  & $36.27$
        & $56.88$ & $63.81$  & $55.67$ 
        & $51.52$ \cellcolor{blue!7!white} & $69.77$ & $44.36$ \\
        \rowcolor{gray!5!white}
        &PA-CGCD \cite{kim2023proxy}
        & $57.14$ & $79.80$ & $41.29$ 
        & $53.05$ & $74.76$  & $48.91$
        & $53.90$ & $68.57$  & $51.33$ 
        & $54.70$ \cellcolor{blue!7!white} & $\textbf{74.38}$ & $47.18$ \\
        \rowcolor{blue!3!white}
        \rot{\rlap{C100}}
        &PromptCCD (Ours) 
        & $67.98$ & $79.39$ & $60.00$ 
        & $60.31$ & $74.76$  & $57.55$
        & $57.02$ & $65.71$  & $55.50$ 
        & $\textbf{61.77}$ \cellcolor{blue!7!white} & $73.29$ & $\textbf{57.68}$ \\ \midrule
        \rowcolor{gray!5!white}
        &GCD \cite{vaze2022generalized} 
        & $74.11$ & $78.57$ & $71.43$ 
        & $71.88$ & $77.86$  & $70.36$
        & $60.95$ & $77.86$  & $57.00$ 
        & $68.98$ \cellcolor{blue!7!white} & $78.10$ & $66.26$ \\
        \rowcolor{gray!5!white}
        &Grow \& Merge \cite{zhang2022grow} 
        & $77.50$ & $77.14$ & $77.71$ 
        & $73.04$ & $75.71$  & $72.36$
        & $63.38$ & $73.57$  & $61.00$ 
        & $71.31$ \cellcolor{blue!7!white} & $75.47$ & $70.36$ \\
        \rowcolor{gray!5!white}
        &MetaGCD \cite{wu2023metagcd}
        & $66.43$ & $84.76$ & $55.43$ 
        & $61.01$ & $78.57$  & $56.55$
        & $68.51$ & $71.43$  & $67.83$ 
        & $65.32$ \cellcolor{blue!7!white} & $78.25$ & $59.94$ \\
        \rowcolor{gray!5!white}
        &PA-CGCD \cite{kim2023proxy}
        & $64.64$ & $81.43$ & $54.57$ 
        & $58.12$ & $72.86$  & $54.36$
        & $60.81$ & $77.14$  & $57.00$ 
        & $61.19$ \cellcolor{blue!7!white} & $77.14$ & $55.31$ \\
        \rowcolor{blue!3!white}
        \rot{\rlap{IN-100}}
        &PromptCCD (Ours) 
        & $78.75$ & $79.52$ & $78.29$ 
        & $75.80$ & $78.57$  & $75.09$
        & $69.46$ & $80.00$  & $67.00$ 
        & $\textbf{74.67}$ \cellcolor{blue!7!white} & $\textbf{79.36}$ & $\textbf{73.46}$ \\ \midrule
        \rowcolor{gray!5!white}
        &GCD \cite{vaze2022generalized} 
        & $60.45$ & $73.81$ & $52.43$ 
        & $46.09$ & $61.79$  & $42.09$
        & $46.49$ & $60.71$  & $43.17$ 
        & $51.01$ \cellcolor{blue!7!white} & $65.44$ & $45.90$ \\
        \rowcolor{gray!5!white}
        &Grow \& Merge \cite{zhang2022grow} 
        & $56.52$ & $65.00$ & $51.43$ 
        & $38.84$ & $55.71$  & $34.55$
        & $47.30$ & $56.07$  & $45.25$ 
        & $47.55$ \cellcolor{blue!7!white} & $58.93$ & $43.74$ \\
        \rowcolor{gray!5!white}
        &MetaGCD \cite{wu2023metagcd}
        & $50.98$ & $70.71$ & $39.14$ 
        & $49.71$ & $60.00$  & $47.09$
        & $51.82$ & $60.71$  & $49.75$ 
        & $50.84$ \cellcolor{blue!7!white} & $63.81$ & $45.33$ \\
        \rowcolor{gray!5!white}
        &PA-CGCD \cite{kim2023proxy}
        & $51.79$ & $70.71$ & $40.43$ 
        & $40.07$ & $64.64$  & $33.82$
        & $47.97$ & $64.29$  & $44.17$ 
        & $46.61$ \cellcolor{blue!7!white} & $\textbf{66.55}$ & $39.47$ \\
        \rowcolor{blue!3!white}
        \rot{\rlap{Tiny}}
        &PromptCCD (Ours) 
        & $61.96$ & $70.24$ & $57.00$ 
        & $52.46$ & $64.64$  & $49.36$
        & $49.86$ & $62.50$  & $46.92$ 
        & $\textbf{53.76}$ \cellcolor{blue!7!white} & $65.79$ & $\textbf{51.09}$ \\ \midrule
        \rowcolor{gray!5!white}
        &GCD \cite{vaze2022generalized} 
        & $62.72$ & $78.73$ & $52.21$ 
        & $50.95$ & $75.71$  & $45.69$
        & $49.17$ & $78.57$  & $43.36$ 
        & $54.28$ \cellcolor{blue!7!white} & $77.67$ & $47.09$ \\
        \rowcolor{gray!5!white}
        &Grow \& Merge \cite{zhang2022grow} 
        & $45.86$ & $64.93$ & $33.33$ 
        & $29.35$ & $60.58$  & $22.77$
        & $41.51$ & $62.86$  & $37.29$ 
        & $38.91$ \cellcolor{blue!7!white} & $62.79$ & $31.13$ \\
        \rowcolor{gray!5!white}
        &MetaGCD \cite{wu2023metagcd}
        & $54.44$ & $72.39$ & $42.65$ 
        & $45.36$ & $78.10$  & $38.46$
        & $50.47$ & $70.71$  & $46.47$ 
        & $50.09$ \cellcolor{blue!7!white} & $73.73$ & $42.53$ \\
        \rowcolor{gray!5!white}
        &PA-CGCD \cite{kim2023proxy}
        & $57.69$ & $81.34$ & $42.16$ 
        & $50.57$ & $73.72$  & $45.69$
        & $58.85$ & $78.57$  & $56.21$ 
        & $55.70$ \cellcolor{blue!7!white} & $77.88$ & $48.02$ \\
        \rowcolor{blue!3!white}
        \rot{\rlap{CUB}}
        &PromptCCD (Ours) 
        & $58.88$ & $77.99$ & $46.32$ 
        & $50.32$ & $78.83$  & $44.31$
        & $63.33$ & $77.14$  & $60.59$ 
        & $\textbf{57.51}$ \cellcolor{blue!7!white} & $\textbf{77.99}$ & $\textbf{50.41}$ \\ 
        \bottomrule
        \end{tabular}
    }
    \label{tab:supp-3-main_result}
\end{table*}
\clearpage
\section{Adapting the \textit{ACC} Metric in GCD for CCD in Each Time Step}
\label{supp: adapt gcd}

In the main paper, when evaluating our method, at each time step $t$, we consider the previously discovered categories as ``known'' (associated with the pseudo labels obtained by our method), which are included in $D^l$ in the \textit{cACC} evaluation algorithm. 
Here, we additionally show the results of applying the commonly used \textit{ACC} metric in GCD to each time step in CCD. Particularly, in each time step, we measure the \textit{ACC} based on $D^l \cup D^u_t$. The \textit{ACC} can be computed following~\cite{vaze2022generalized}. We summarize the adapted evaluation metric in Alg.~\ref{alg:gcd_eval} and report the results in Tab.~\ref{tab:supp-4-main_result}.
Overall, our method consistently outperforms other methods on all datasets on `All' and `New' splits.

\scalebox{0.75}{
\centering
\begin{minipage}{1.0\linewidth} \centering
\begin{algorithm}[H]
\begin{algorithmic}[1]
    \Statex \textbf{Input:} Models $\{f_{\theta}^{t} \mid t=1, \dots, T\}$ and datasets $\{D^l , D^u\}$.
    \Statex \textbf{Output:} \textit{ACC} value.
    \Statex \textbf{Require:} \Call{SS-$k$-means}{Model, Labelled set, Unlabelled set}.
\For{$t \in \{1, \cdots, T\}$}
    \State $ACC_t$ $\gets$ \Call{SS-$k$-means}{$f_{\theta}^{t}$, $D^{l}$, $D^u_t$}
\EndFor
\State $ACCs \gets \{ACC_{t} \mid t=1, \dots, T\}$
\State $ACC \gets \Call{Average}{ACCs}$
\State \Return $ACC$
\end{algorithmic}
\caption{standard incremental GCD evaluation metric}
\label{alg:gcd_eval}
\end{algorithm}
\end{minipage}
}

\definecolor{Gray}{gray}{0.9}
\definecolor{PaleBlue}{rgb}{0.7529, 0.9137, 0.9372}
\definecolor{BeauBlue}{rgb}{0.7686, 0.8470, 0.9529}
\definecolor{Mauve}{rgb}{0.8 , 0.7098, 0.9843}
\definecolor{PaleViolet}{rgb}{0.8156, 0.6431, 1.0}
\definecolor{Salmon}{rgb}{1.0, 0.8980, 0.6920}
\definecolor{Pink}{rgb}{1.0, 0.6902, 0.7908}
\definecolor{cyan}{rgb}{0.906, 0.969, 0.965}
\definecolor{Mint}{rgb}{0.6902, 1.0, 0.7451}
\definecolor{SoftP}{rgb}{0.945, 0.933, 0.949}

\begin{table*}[htbp]
\caption{Comparison using the adapted \textit{ACC} metric from GCD in Alg.~\ref{alg:gcd_eval}.}
    \centering
    \resizebox{0.9\columnwidth}{!}{
    \centering
        \begin{tabular}{cl @{\hskip 0.1in} ccc @{\hskip 0.1in} ccc @{\hskip 0.1in} ccc @{\hskip 0.1in} ||ccc}
        \toprule
        \multicolumn{2}{c}{}
        & \multicolumn{3}{c}{Stage 1 \textit{ACC} (\%)}
        & \multicolumn{3}{c}{Stage 2 \textit{ACC} (\%)}
        & \multicolumn{3}{c}{Stage 3 \textit{ACC} (\%)} 
        & \multicolumn{3}{c}{Average \textit{ACC} (\%)} 
        \\ 
        \multicolumn{1}{l}{} &
        \multicolumn{1}{c}{Method} &
        \multicolumn{1}{c}{\textit{All}} & 
        \multicolumn{1}{c}{\textit{Old}} &
        \multicolumn{1}{c}{\textit{New}} &
        \multicolumn{1}{c}{\textit{All}} & 
        \multicolumn{1}{c}{\textit{Old}} & 
        \multicolumn{1}{c}{\textit{New}} &
        \multicolumn{1}{c}{\textit{All}} & 
        \multicolumn{1}{c}{\textit{Old}} & 
        \multicolumn{1}{c}{\textit{New}} &
        \multicolumn{1}{c}{\textit{All} \cellcolor{blue!7!white}} & 
        \multicolumn{1}{c}{\textit{Old}} & 
        \multicolumn{1}{c}{\textit{New}} 
        \\ 
        \cmidrule[0.1pt](r{0.80em}){1-2}%
        \cmidrule[0.1pt](r{0.80em}){3-5}%
        \cmidrule[0.1pt](r{0.80em}){6-8}%
        \cmidrule[0.1pt](r{0.80em}){9-11}%
        \cmidrule[0.1pt](r{0.80em}){12-14}%
        \rowcolor{gray!5!white}
        &GCD \cite{vaze2022generalized} 
        & $67.65$ & $83.59$ & $56.49$ 
        & $49.85$ & $71.52$  & $45.71$
        & $44.51$ & $81.62$  & $38.02$ 
        & $54.00$ \cellcolor{blue!7!white} & $78.91$ & $46.74$ \\
        \rowcolor{gray!5!white}
        &Grow \& Merge \cite{zhang2022grow} 
        & $64.77$ & $70.49$  & $60.77$ 
        & $61.27$ & $64.00$  & $60.75$
        & $43.56$ & $59.90$  & $40.70$ 
        & $56.53$ \cellcolor{blue!7!white} & $64.80$ & $54.07$ \\
        \rowcolor{gray!5!white}
        &MetaGCD \cite{wu2023metagcd}
        & $56.20$ & $79.59$  & $39.83$ 
        & $54.09$ & $68.57$  & $57.51$
        & $44.77$ & $60.10$  & $42.08$ 
        & $51.69$ \cellcolor{blue!7!white} & $69.42$ & $46.47$ \\ 
        \rowcolor{gray!5!white}
        &PA-CGCD \cite{kim2023proxy}
        & $57.43$ & $80.29$  & $41.43$ 
        & $63.42$ & $86.76$  & $55.05$
        & $50.81$ & $83.24$  & $45.73$ 
        & $57.22$ \cellcolor{blue!7!white} & $\textbf{83.43}$ & $47.40$ \\ 
        \rowcolor{blue!3!white}
        \rot{\rlap{C100}}
        &PromptCCD (Ours) 
        & $70.69$ & $80.90$  & $63.54$ 
        & $65.65$ & $74.57$  & $63.95$
        & $51.94$ & $83.62$  & $46.40$ 
        & $\textbf{62.76}$ \cellcolor{blue!7!white} & $79.70$ & $\textbf{57.96}$ \\ \midrule
        \rowcolor{gray!5!white}
        &GCD \cite{vaze2022generalized} 
        & $75.65$ & $84.69$  & $69.31$ 
        & $64.39$ & $78.71$  & $62.65$
        & $53.49$ & $80.19$  & $48.82$ 
        & $64.51$ \cellcolor{blue!7!white} & $81.20$ & $60.26$ \\
        \rowcolor{gray!5!white}
        &Grow \& Merge \cite{zhang2022grow} 
        & $75.34$ & $76.78$  & $74.34$ 
        & $63.11$ & $78.00$  & $60.27$
        & $54.06$ & $73.71$  & $50.62$ 
        & $64.17$ \cellcolor{blue!7!white} & $76.16$ & $61.74$ \\
        \rowcolor{gray!5!white}
        &MetaGCD \cite{wu2023metagcd}
        & $65.61$ & $83.92$  & $52.80$ 
        & $60.23$ & $85.05$  & $55.49$
        & $65.09$ & $77.43$  & $62.93$ 
        & $63.64$ \cellcolor{blue!7!white} & $82.13$ & $57.07$ \\ 
        \rowcolor{gray!5!white}
        &PA-CGCD \cite{kim2023proxy}
        & $70.05$ & $82.61$  & $61.26$ 
        & $66.40$ & $95.52$  & $60.84$
        & $62.41$ & $93.81$  & $56.92$ 
        & $66.29$ \cellcolor{blue!7!white} & $\textbf{90.65}$ & $59.67$ \\ 
        \rowcolor{blue!3!white}
        \rot{\rlap{IN-100}}
        &PromptCCD (Ours) 
        & $79.56$ & $84.24$  & $76.29$ 
        & $66.34$ & $81.43$  & $63.45$
        & $58.91$ & $78.95$  & $64.05$ 
        & $\textbf{68.27}$ \cellcolor{blue!7!white} & $81.54$ & $\textbf{67.93}$ \\ \midrule
        \rowcolor{gray!5!white}
        &GCD \cite{vaze2022generalized} 
        & $63.62$ & $73.14$  & $56.96$ 
        & $54.09$ & $67.19$  & $51.59$
        & $47.98$ & $62.48$  & $45.44$ 
        & $55.23$ \cellcolor{blue!7!white} & $67.60$ & $51.33$ \\
        \rowcolor{gray!5!white}
        &Grow \& Merge \cite{zhang2022grow} 
        & $59.52$ & $64.24$  & $56.21$ 
        & $50.19$ & $57.95$  & $48.71$
        & $52.06$ & $54.90$  & $51.57$ 
        & $53.92$ \cellcolor{blue!7!white} & $59.03$ & $52.16$ \\
        \rowcolor{gray!5!white}
        &MetaGCD \cite{wu2023metagcd}
        & $59.41$ & $73.90$  & $49.27$ 
        & $59.90$ & $61.90$  & $59.52$
        & $53.43$ & $61.29$  & $52.06$ 
        & $57.58$ \cellcolor{blue!7!white} & $65.70$ & $53.62$ \\ 
        \rowcolor{gray!5!white}
        &PA-CGCD \cite{kim2023proxy}
        & $56.01$ & $74.96$  & $42.74$ 
        & $46.81$ & $67.14$  & $42.93$
        & $52.74$ & $88.86$  & $46.92$ 
        & $51.85$ \cellcolor{blue!7!white} & $\textbf{76.99}$ & $44.03$ \\ 
        \rowcolor{blue!3!white}
        \rot{\rlap{Tiny}}
        &PromptCCD (Ours) 
        & $68.67$ & $72.84$  & $65.76$ 
        & $60.11$ & $73.48$  & $57.56$
        & $51.51$ & $60.71$  & $49.90$ 
        & $\textbf{60.10}$ \cellcolor{blue!7!white} & $69.01$ & $\textbf{57.74}$ \\ \midrule
        \rowcolor{gray!5!white}
        &GCD \cite{vaze2022generalized} 
        & $58.80$ & $75.71$  & $47.49$ 
        & $47.50$ & $80.71$  & $40.45$
        & $47.67$ & $76.43$  & $42.06$ 
        & $51.32$ \cellcolor{blue!7!white} & $77.62$ & $43.33$ \\
        \rowcolor{gray!5!white}
        &Grow \& Merge \cite{zhang2022grow} 
        & $44.21$ & $65.00$  & $30.31$ 
        & $32.25$ & $69.29$  & $24.39$
        & $37.18$ & $67.14$  & $31.34$ 
        & $37.88$ \cellcolor{blue!7!white} & $67.14$ & $28.68$ \\
        \rowcolor{gray!5!white}
        &MetaGCD \cite{wu2023metagcd}
        & $50.93$ & $71.07$  & $37.47$ 
        & $43.50$ & $75.71$  & $36.67$
        & $45.92$ & $77.86$  & $39.69$ 
        & $46.78$ \cellcolor{blue!7!white} & $74.88$ & $37.94$ \\ 
        \rowcolor{gray!5!white}
        &PA-CGCD \cite{kim2023proxy}
        & $55.94$ & $73.21$  & $44.39$ 
        & $53.00$ & $76.43$  & $48.03$
        & $57.46$ & $86.43$  & $51.81$ 
        & $55.47$ \cellcolor{blue!7!white} & $78.69$ & $48.08$ \\ 
        \rowcolor{blue!3!white}
        \rot{\rlap{CUB}}
        &PromptCCD (Ours) 
        & $57.08$ & $75.00$  & $45.11$ 
        & $52.87$ & $85.71$  & $45.91$
        & $58.04$ & $79.29$  & $53.90$ 
        & $\textbf{56.00}$ \cellcolor{blue!7!white} & $\textbf{80.00}$ & $\textbf{48.31}$ \\ 
        \bottomrule
        \end{tabular}
    }
    \label{tab:supp-4-main_result}
\end{table*}
\clearpage
\section{Implementation Details for Augmenting Grow \& Merge with ViT}
\label{supp: gm on vit}
As the most relevant work Grow \& Merge (G\&M) \cite{zhang2022grow} uses ResNet18 \cite{he2016deep} as the backbone and the Momentum Contrast (MoCo) \cite{he2020momentum} for representation learning, to have a fair comparison, we augment G\&M from two aspects, the pretraining strategy and the dual branch network (static and dynamic branch), leveraging the more powerful ViT backbone.
First, we change the pretraining strategy MoCo to joint supervised and unsupervised contrastive learning with DINO features. Second, for the dual branch network in \cite{zhang2022grow}, originally, the ResNet18 is divided into several layers (excluding the fully connected layers) where before the last layer, G\&M divides the last layer into two branches, $\ie$, the static branch and the dynamic branch. By design, the static branch is the backbone's last layer, while the dynamic branch consists of several branches of $T-1$ layers, where $T$ is the number of stages. To maintain this design, we accordingly implement a dual-branch architecture network based on ViT backbone. Given that ViT backbone consists of several blocks, we freeze all blocks except the last block as the static branch. Moreover, before the last block, we add another $T-1$ blocks as the dynamic branches used exclusively for each stage $t$. All the rest designs are the same as \cite{zhang2022grow}. At $t=0$, \ie, during the initial stage, we optimize the static branch, and at $t>0$, we freeze the static branch and perform static-dynamic distillation while optimizing the dynamic branch $t$ for novel class discovery following  G\&M.

We compare our method with the \textit{improved} G\&M under both \textit{transductive} and \textit{inductive} evaluation protocols, using the \textit{cACC} evaluation metric. As shown in Tab.~\ref{tab:supp-6-gm}, our \textit{improved} G\&M significantly outperforms the original implementation, leading to a fair comparison with our method. 
However, our method obtains an overall accuracy of $64.17\%$, which is still substantially better.  

\textbf{Note}: For experiments in the main paper, we compare our model with the \textit{improved} Grow \& Merge (G\&M).

\definecolor{Gray}{gray}{0.9}
\definecolor{PaleBlue}{rgb}{0.7529, 0.9137, 0.9372}
\definecolor{BeauBlue}{rgb}{0.7686, 0.8470, 0.9529}
\definecolor{Mauve}{rgb}{0.8 , 0.7098, 0.9843}
\definecolor{PaleViolet}{rgb}{0.8156, 0.6431, 1.0}
\definecolor{Salmon}{rgb}{1.0, 0.8980, 0.6920}
\definecolor{Pink}{rgb}{1.0, 0.6902, 0.7908}
\definecolor{Mint}{rgb}{0.6902, 1.0, 0.7451}
\definecolor{cyan}{rgb}{0.906, 0.969, 0.965}
\definecolor{SoftP}{rgb}{0.945, 0.933, 0.949}

\begin{table*}[htbp]
\caption{Comparison with different Grow \& Merge implementations on CIFAR100 datasets.}
    \centering
    \resizebox{0.9\columnwidth}{!}{
    \centering
        \begin{tabular}{cl @{\hskip 0.1in} ccc @{\hskip 0.1in} ccc @{\hskip 0.1in} ccc @{\hskip 0.1in} ||ccc}
        \toprule
        \multicolumn{2}{c}{}
        & \multicolumn{3}{c}{Stage 1 \textit{ACC} (\%)} 
        & \multicolumn{3}{c}{Stage 2 \textit{ACC} (\%)}
        & \multicolumn{3}{c}{Stage 3 \textit{ACC} (\%)} 
        & \multicolumn{3}{c}{Average \textit{ACC} (\%)}
        \\ 
        \multicolumn{1}{l}{} &
        \multicolumn{1}{c}{Method} &
        \multicolumn{1}{c}{\textit{All}} & 
        \multicolumn{1}{c}{\textit{Old}} &
        \multicolumn{1}{c}{\textit{New}} &
        \multicolumn{1}{c}{\textit{All}} & 
        \multicolumn{1}{c}{\textit{Old}} & 
        \multicolumn{1}{c}{\textit{New}} &
        \multicolumn{1}{c}{\textit{All}} & 
        \multicolumn{1}{c}{\textit{Old}} & 
        \multicolumn{1}{c}{\textit{New}} &
        \multicolumn{1}{c}{\textit{All} \cellcolor{blue!7!white}} & 
        \multicolumn{1}{c}{\textit{Old}} & 
        \multicolumn{1}{c}{\textit{New}} 
        \\ \midrule
        \multicolumn{14}{c}{\textit{Transductive Evaluation}} \\ \midrule
        \rowcolor{gray!5!white}
        &Grow \& Merge \cite{zhang2022grow} 
        & $22.91$ & $30.20$ & $17.80$ 
        & $21.47$ & $25.71$ & $20.65$ 
        & $24.91$ & $24.00$ & $27.25$
        & $23.10$ \cellcolor{blue!7!white} & $26.64$ & $21.90$ \\
        \rowcolor{gray!5!white}
        &Grow \& Merge (\textit{improved}) 
        & $64.77$ & $70.49$ & $60.77$ 
        & $58.32$ & $62.95$ & $57.44$ 
        & $49.21$ & $57.62$ & $47.73$
        & $\textbf{57.43}$ \cellcolor{blue!7!white} & $\textbf{63.68}$ & $\textbf{55.31}$ \\
        \rowcolor{blue!3!white}
        &PromptCCD (Ours) 
        & $70.69$ & $80.90$ & $63.54$ 
        & $64.08$ & $73.14$ & $62.35$ 
        & $57.73$ & $72.67$ & $55.12$
        & $\textbf{64.17}$ \cellcolor{blue!7!white} & $\textbf{75.57}$ & $\textbf{60.34}$ \\ \midrule
        \multicolumn{14}{c}{\textit{Inductive Evaluation}} \\ \midrule
        \rowcolor{gray!5!white}
        &Grow \& Merge \cite{zhang2022grow} 
        & $38.32$ & $60.61$ & $22.71$ 
        & $29.62$ & $60.48$ & $23.73$ 
        & $31.91$ & $60.95$ & $26.83$
        & $33.28$ \cellcolor{blue!7!white} & $60.68$ & $24.42$ \\
        \rowcolor{gray!5!white}
        &Grow \& Merge (\textit{improved})  
        & $64.77$ & $70.49$  & $60.77$ 
        & $61.27$ & $64.00$  & $60.75$
        & $43.56$ & $59.90$  & $40.70$ 
        & $\textbf{56.53}$ \cellcolor{blue!7!white} & $\textbf{64.80}$ & $\textbf{54.07}$ \\
        \rowcolor{blue!3!white}
        &PromptCCD (Ours) 
        & $67.98$ & $79.39$ & $60.00$ 
        & $60.31$ & $74.76$  & $57.55$
        & $57.02$ & $65.71$  & $55.50$ 
        & $\textbf{61.77}$ \cellcolor{blue!7!white} & $\textbf{73.29}$ & $\textbf{57.68}$ \\
        \bottomrule
        \end{tabular}
    }
    \label{tab:supp-6-gm}
\end{table*}
\clearpage
\section{Additional Comparison under Other CCD Settings and Metrics}
\label{supp: recent works}
In this section, we provide additional comparison with PA-CGCD~\cite{kim2023proxy} and MetaGCD~\cite{wu2023metagcd}, following their experimental settings, including data splits and evaluation protocols. 
In Tab.~\ref{tab: rebuttal_pa_iccv}, we follow PA-CGCD's experimental setting and use their evaluation metrics (described in Sec.~$4.2$ of~\cite{kim2023proxy}). As can be seen, our method achieves the best performance following the setting of~\cite{kim2023proxy}.
In addition, we also follow MetaGCD's data splits. The common GCD evaluation metric, $ACC$, is adopted in their original paper. We experiment under their setting, and report the results in Tab.~\ref{tab: metagcd}. As can be seen, our method outperforms MetaGCD and all other methods. 
These results further demonstrate the superiority of our methods. 

\definecolor{Gray}{gray}{0.9}
\definecolor{PaleBlue}{rgb}{0.7529, 0.9137, 0.9372}
\definecolor{BeauBlue}{rgb}{0.7686, 0.8470, 0.9529}
\definecolor{Mauve}{rgb}{0.8 , 0.7098, 0.9843}
\definecolor{PaleViolet}{rgb}{0.8156, 0.6431, 1.0}
\definecolor{Salmon}{rgb}{1.0, 0.8980, 0.6920}
\definecolor{Pink}{rgb}{1.0, 0.6902, 0.7908}
\definecolor{Mint}{rgb}{0.6902, 1.0, 0.7451}
\definecolor{SoftP}{rgb}{0.945, 0.933, 0.949}

\begin{table*}[!h]
\caption{Comparison with PA-CGCD \cite{kim2023proxy} on CUB. For experiment settings and evaluation metrics, please refer to the original paper's Sec.~$4.2$ (Tab.~$4$, DINO ViT-B/16 experiments).} 
    \centering
    \resizebox{0.45\columnwidth}{!}{%
    \centering
        \begin{tabular}{lcccc}
        \toprule
        \multicolumn{1}{l}{Method} &
        \multicolumn{1}{c}{$\mathcal{M}_{all} \uparrow$ \cellcolor{blue!7!white}} & 
        \multicolumn{1}{c}{$\mathcal{M}_{o} \uparrow$} &
        \multicolumn{1}{c}{$\mathcal{M}_{f} \downarrow$} &
        \multicolumn{1}{c}{$\mathcal{M}_{d} \uparrow$} 
        \\ \midrule
            \rowcolor{gray!5!white}
            GCD \cite{vaze2022generalized}   & $62.70$ \cellcolor{blue!7!white} & $71.40$ &  $09.57$ & $56.01$\\
            \rowcolor{gray!5!white}
            Grow \& Merge \cite{zhang2022grow} & $42.12$ \cellcolor{blue!7!white} & $60.21$ & $23.24$ & $27.63$\\
            \rowcolor{gray!5!white}
            PA-CGCD \cite{kim2023proxy} & $72.51$ \cellcolor{blue!7!white} & $74.28$ & $09.49$ & $65.60$ \\ \hline
            \rowcolor{blue!3!white}
            PromptCCD (Ours)  & \textbf{76.23} \cellcolor{blue!7!white} & \textbf{78.44} & \textbf{06.07} & \textbf{74.46}\\
        \bottomrule
        \end{tabular}
    }
    \label{tab: rebuttal_pa_iccv}
    \vspace{-6mm}
\end{table*}

\vspace{5.0mm}

\definecolor{Gray}{gray}{0.9}
\definecolor{PaleBlue}{rgb}{0.7529, 0.9137, 0.9372}
\definecolor{BeauBlue}{rgb}{0.7686, 0.8470, 0.9529}
\definecolor{Mauve}{rgb}{0.8 , 0.7098, 0.9843}
\definecolor{PaleViolet}{rgb}{0.8156, 0.6431, 1.0}
\definecolor{Salmon}{rgb}{1.0, 0.8980, 0.6920}
\definecolor{Pink}{rgb}{1.0, 0.6902, 0.7908}
\definecolor{cyan}{rgb}{0.906, 0.969, 0.965}
\definecolor{Mint}{rgb}{0.6902, 1.0, 0.7451}
\definecolor{SoftP}{rgb}{0.945, 0.933, 0.949}

\begin{table*}[htbp]
\caption{Comparison with MetaGCD \cite{wu2023metagcd} on CIFAR100. For experimental settings and evaluation metric, please refer to the original paper's Sec.~$4$.}
    \resizebox{\columnwidth}{!}{%
    \centering
        \begin{tabular}{l @{\hskip 0.1in} ccc @{\hskip 0.1in} ccc @{\hskip 0.1in} ccc @{\hskip 0.1in} ccc @{\hskip 0.1in} ||ccc}
        \toprule
        \multicolumn{1}{c}{}
        & \multicolumn{3}{c}{Stage 1 \textit{ACC} (\%)}
        & \multicolumn{3}{c}{Stage 2 \textit{ACC} (\%)}
        & \multicolumn{3}{c}{Stage 3 \textit{ACC} (\%)}
        & \multicolumn{3}{c}{Stage 4 \textit{ACC} (\%)}
        & \multicolumn{3}{c}{Average \textit{ACC} (\%)}
        \\ 
        \multicolumn{1}{c}{Method} &
        \multicolumn{1}{c}{\textit{All}} & 
        \multicolumn{1}{c}{\textit{Old}} &
        \multicolumn{1}{c}{\textit{New}} &
        \multicolumn{1}{c}{\textit{All}} & 
        \multicolumn{1}{c}{\textit{Old}} & 
        \multicolumn{1}{c}{\textit{New}} &
        \multicolumn{1}{c}{\textit{All}} & 
        \multicolumn{1}{c}{\textit{Old}} & 
        \multicolumn{1}{c}{\textit{New}} &
        \multicolumn{1}{c}{\textit{All}} & 
        \multicolumn{1}{c}{\textit{Old}} & 
        \multicolumn{1}{c}{\textit{New}} &
        \multicolumn{1}{c}{\textit{All} \cellcolor{blue!7!white}} & 
        \multicolumn{1}{c}{\textit{Old}} & 
        \multicolumn{1}{c}{\textit{New}} 
        \\ 
        \cmidrule[0.1pt](r{0.80em}){1-1}%
        \cmidrule[0.1pt](r{0.80em}){2-4}%
        \cmidrule[0.1pt](r{0.80em}){5-7}%
        \cmidrule[0.1pt](r{0.80em}){8-10}%
        \cmidrule[0.1pt](r{0.80em}){11-13}%
        \cmidrule[0.1pt](r{0.80em}){14-16}%
        \rowcolor{gray!5!white}
        RankStats \cite{han2020automatically}
        & $62.33$  & $64.22$ & $31.60$ 
        & $55.01$  & $58.55$ & $26.85$ 
        & $51.77$  & $56.70$ & $25.47$ 
        & $47.51$  & $54.59$ & $17.20$
        & $54.16$  \cellcolor{blue!7!white} & $58.52$ & $25.28$  \\
        \rowcolor{gray!5!white}
        FRoST \cite{roy2022class}
        & $67.14$  & $68.57$ & $50.73$ 
        & $67.01$  & $68.82$ & $52.60$ 
        & $62.35$  & $65.48$ & $45.67$ 
        & $55.84$  & $59.06$ & $42.95$
        & $63.09$  \cellcolor{blue!7!white} & $65.48$ & $47.99$  \\
        \rowcolor{gray!5!white}
        GCD \cite{vaze2022generalized}
        & $76.78$  & $77.91$ & $58.60$ 
        & $73.67$  & $75.29$ & $60.70$ 
        & $72.77$  & $74.72$ & $62.33$ 
        & $71.44$  & $74.75$ & $58.20$
        & $73.67$  \cellcolor{blue!7!white} & $75.67$ & $59.96$  \\
        \rowcolor{gray!5!white}
        Grow \& Merge \cite{zhang2022grow}
        & $78.29$  & $79.91$ & $66.00$ 
        & $77.58$  & $79.64$ & $61.13$ 
        & $74.56$  & $77.60$ & $58.14$ 
        & $72.02$  & $75.98$ & $56.32$
        & $75.61$  \cellcolor{blue!7!white} & $78.28$ & $60.40$  \\
        \rowcolor{gray!5!white}
        MetaGCD \cite{wu2023metagcd}
        & $78.96$  & $79.36$ & $72.60$ 
        & $78.67$  & $79.41$ & $66.81$
        & $76.06$  & $78.20$ & $64.87$
        & $74.56$  & $77.60$ & $61.14$
        & $77.06$ \cellcolor{blue!7!white} & $78.64$ & $66.35$  \\ 
        \rowcolor{blue!3!white}
        PromptCCD (Ours) 
        & $90.06$  & $90.50$ & $89.47$ 
        & $82.67$  & $88.80$ & $76.23$ 
        & $81.48$  & $84.60$ & $78.80$ 
        & $70.30$  & $75.87$ & $67.64$ 
        & $\textbf{81.13}$ \cellcolor{blue!7!white} & $\textbf{84.94}$ & $\textbf{78.04}$ \\ 
        \bottomrule
        \end{tabular}
    }
    \label{tab: metagcd}
\end{table*}
\clearpage
\section{Experiments on Additional Data Splits}
\label{supp: diff class ratio}

The data splits in the main paper follow~\cite{zhang2022grow}, as reported in Tab.~$2$. The classes are split into 7:1:1:1, while the samples (${D}^l$, ${D}^u_1$, ${D}^u_2$, ${D}^u_3$) in each stage are divided following the percentages in Tab.~$2$.
To further mimic the real-world scenario, which is characterized by an abrupt increase or decrease in the number of classes of each stage, we experiment on another 3 different class splits: (1) 4:2:2:2 -- the number of the unseen classes is greater than that of the seen classes; (2) 4:3:2:1 -- the number of the unseen classes is decreasing for each stage; (3) 1:2:3:4 -- the number of the unseen class is increasing for each stage. As shown in Tab.~\ref{tab: different data split}, we compare our model with  GCD and Grow \& Merge on the CIFAR100 dataset. Our model consistently outperforms others by a large margin across the board.

\definecolor{Gray}{gray}{0.9}
\definecolor{PaleBlue}{rgb}{0.7529, 0.9137, 0.9372}
\definecolor{BeauBlue}{rgb}{0.7686, 0.8470, 0.9529}
\definecolor{Mauve}{rgb}{0.8 , 0.7098, 0.9843}
\definecolor{PaleViolet}{rgb}{0.8156, 0.6431, 1.0}
\definecolor{Salmon}{rgb}{1.0, 0.8980, 0.6920}
\definecolor{Pink}{rgb}{1.0, 0.6902, 0.7908}
\definecolor{Mint}{rgb}{0.6902, 1.0, 0.7451}
\definecolor{cyan}{rgb}{0.906, 0.969, 0.965}
\definecolor{SoftP}{rgb}{0.945, 0.933, 0.949}

\begin{table*}[!h]
\caption{Experiments on different class splits scenarios on CIFAR100.}
    \centering
    \resizebox{1.0\columnwidth}{!}{
    \centering
        \begin{tabular}{cl @{\hskip 0.1in} ccc @{\hskip 0.1in} ccc @{\hskip 0.1in} ccc @{\hskip 0.1in} ||ccc}
        \toprule
        \multicolumn{2}{c}{}
        & \multicolumn{3}{c}{Stage 1 \textit{ACC} (\%)}
        & \multicolumn{3}{c}{Stage 2 \textit{ACC} (\%)} 
        & \multicolumn{3}{c}{Stage 3 \textit{ACC} (\%)}
        & \multicolumn{3}{c}{Average \textit{ACC} (\%)}
        \\ 
        \multicolumn{1}{l}{} &
        \multicolumn{1}{c}{Method} &
        \multicolumn{1}{c}{\textit{All}} & 
        \multicolumn{1}{c}{\textit{Old}} &
        \multicolumn{1}{c}{\textit{New}} &
        \multicolumn{1}{c}{\textit{All}} & 
        \multicolumn{1}{c}{\textit{Old}} & 
        \multicolumn{1}{c}{\textit{New}} &
        \multicolumn{1}{c}{\textit{All}} & 
        \multicolumn{1}{c}{\textit{Old}} & 
        \multicolumn{1}{c}{\textit{New}} &
        \multicolumn{1}{c}{\textit{All} \cellcolor{blue!7!white}} & 
        \multicolumn{1}{c}{\textit{Old}} & 
        \multicolumn{1}{c}{\textit{New}} 
        \\ \midrule
        \multicolumn{14}{c}{\textit{Class Split: 4:2:2:2}} \\ \midrule
        \rowcolor{gray!5!white}
        &GCD \cite{vaze2022generalized}
        & $78.25$ & $55.36$ & $80.54$ 
        & $65.79$ & $39.83$ & $66.98$ 
        & $38.72$ & $39.83$ & $38.50$ 
        & $60.92$ \cellcolor{blue!7!white} & $45.01$ & $62.01$ \\
        \rowcolor{gray!5!white}
        &Grow \& Merge \cite{zhang2022grow}
        & $51.17$ & $41.86$ & $52.10$ 
        & $45.90$ & $31.50$ & $46.57$ 
        & $34.72$ & $45.17$ & $32.63$ 
        & $43.93$ \cellcolor{blue!7!white} & $39.51$ & $43.77$ \\ 
        \rowcolor{blue!3!white}
        &PromptCCD (Ours)
        & $78.53$ & $50.50$ & $81.34$ 
        & $74.22$ & $59.67$ & $74.89$ 
        & $52.64$ & $43.50$ & $54.47$ 
        & $\textbf{68.46}$ \cellcolor{blue!7!white} & $\textbf{51.22}$ & $\textbf{70.23}$ \\ \midrule
        \multicolumn{14}{c}{\textit{Class Split: 4:3:2:1 (decreasing)}} \\ \midrule
        \rowcolor{gray!5!white}
        &GCD \cite{vaze2022generalized}
        & $58.62$ & $59.79$ & $58.50$ 
        & $51.99$ & $42.17$ & $52.44$ 
        & $40.69$ & $40.83$ & $40.67$ 
        & $50.43$ \cellcolor{blue!7!white} & $47.59$ & $50.54$ \\
        \rowcolor{gray!5!white}
        &Grow \& Merge \cite{zhang2022grow}
        & $41.89$ & $52.83$ & $39.70$ 
        & $44.25$ & $42.83$ & $44.32$ 
        & $34.97$ & $35.50$ & $34.87$ 
        & $40.37$ \cellcolor{blue!7!white} & $43.72$ & $39.63$ \\ 
        \rowcolor{blue!3!white}
        &PromptCCD (Ours)
        & $57.10$ & $61.14$ & $55.70$ 
        & $64.10$ & $51.00$ & $64.70$ 
        & $47.67$ & $38.67$ & $49.47$ 
        & $\textbf{56.29}$ \cellcolor{blue!7!white} & $\textbf{50.27}$ & $\textbf{56.62}$ \\
        \midrule
        \multicolumn{14}{c}{\textit{Class Split: 1:2:3:4 (increasing)}} \\ \midrule
        \rowcolor{gray!5!white}
        &GCD \cite{vaze2022generalized}
        & $52.89$ & $63.21$ & $51.86$ 
        & $53.94$ & $53.67$ & $53.95$ 
        & $45.49$ & $33.00$ & $45.21$ 
        & $50.77$ \cellcolor{blue!7!white} & $49.96$ & $50.34$ \\
        \rowcolor{gray!5!white}
        &Grow \& Merge \cite{zhang2022grow}
        & $50.40$ & $44.64$ & $50.98$ 
        & $44.48$ & $36.33$ & $44.85$ 
        & $41.89$ & $52.83$ & $39.70$ 
        & $45.59$ \cellcolor{blue!7!white} & $44.60$ & $45.18$ \\ 
        \rowcolor{blue!3!white}
        &PromptCCD (Ours)
        & $50.21$ & $63.57$ & $48.88$ 
        & $49.96$ & $60.50$ & $49.47$ 
        & $57.00$ & $58.00$ & $56.80$ 
        & $\textbf{52.39}$ \cellcolor{blue!7!white} & $\textbf{60.69}$ & $\textbf{51.72}$ \\
        \bottomrule
        \end{tabular}
    }
    \label{tab: different data split}
\end{table*}
\clearpage
\section{Why Finetune the Final Block of DINO for CCD?}
\label{supp: learn parameters}

We analyze the number of learning parameters for each compared model and explain why the final block of our backbone is finetuned. Our motivation is to repurpose self-supervised vision foundation models for CCD. We choose DINO \cite{caron2021emerging, oquab2023dinov2} as our vision foundation model to tackle CCD. DINO is a transformer-based vision foundation model pretrained on ImageNet-1K \cite{russakovsky2015imagenet} with a resolution of $224*224$ pixels. The model is trained in a self-supervised manner (no label information) with around $86M$ parameters. 
Self-supervised models have been widely adopted in both NCD \cite{han2021autonovel} and GCD~\cite{vaze2022generalized} literature so far. 
Thus, we follow the GCD literature to use the strong DINO’s self-supervised pretrained model for all compared models.
We finetune the final block of its backbone and report the number of learnable parameters for each model in Tab.~\ref{tab: parameter}. Our PromptCCD’s learnable parameters consist of two parts: the final block of the backbone and the parameter from GMP’s GMM. The latter only accommodates $\{(2*|\hat{z}_i| + 1) * C\}$ parameters, where $C$ is the number of components, and $|\hat{z}_i|$ is the feature size of the $\texttt{[CLS]}$ token, which in this case is $768$. Compared with PromptCCD-B~$w/$$\{$L2P, DP$\}$, our model’s learnable parameters are only $0.33\%$ higher when $C=100$, which is still efficient. For PromptCCD++, the learnable parameters further include the part-level prompt pools and the part router. Since the number of part pools $P$ depends on the dataset, the total parameter count is correspondingly larger than PromptCCD, but the overhead remains modest given the improved discovery performance. In our setting, each part pool contributes $\{0.169M\}$ parameters, giving $\{0.169M * P\}$ PLP parameters in total; additionally, the part router contributes about $14.7M$ parameters and is trained only at $t{=}0$ before being frozen in later stages. Thus, for $P{=}8$, PromptCCD++ has about $46.3M$ learnable parameters at $t{=}0$ and $31.6M$ learnable parameters at $t{>}0$. Moreover, in terms of prompt tokens embedded to the backbone model, PromptCCD’s GMP is still efficient as we only embed $\{|\text{top-k}| * |\hat{z}_i|\}$, which is notably smaller compared to L2P’s method, $\ie$, $\{|\text{top-k}| * |\hat{z}_i| * L_{pp}\}$, where $L_{pp}$ is the prompt pool’s token length, and DP’s method, $\ie$, $\{|\text{top-k}| * |\hat{z}_i| * L^{G}_{pp}\}$ for the $\textit{G-prompt}$ and $\{|\text{top-k}| * |\hat{z}_i| * L^{E}_{pp}\}$ for its $\textit{E-Prompt}$, where $L^{G}_{pp}$ is the prompt pool’s token length for task-invariant prompt while $L^{E}_{pp}$ is the token length for task-specific prompts. Here, we highlight that PromptCCD uses a compact prompt representation, while PromptCCD++ introduces a slightly larger but more expressive part-level memory, leading to stronger category discovery performance. Each prompt token represents the class prototype for each category, providing strong guidance for CCD.
 
\definecolor{Gray}{gray}{0.9}
\definecolor{PaleBlue}{rgb}{0.7529, 0.9137, 0.9372}
\definecolor{BeauBlue}{rgb}{0.7686, 0.8470, 0.9529}
\definecolor{Mauve}{rgb}{0.8 , 0.7098, 0.9843}
\definecolor{PaleViolet}{rgb}{0.8156, 0.6431, 1.0}
\definecolor{Salmon}{rgb}{1.0, 0.8980, 0.6920}
\definecolor{Pink}{rgb}{1.0, 0.6902, 0.7908}
\definecolor{Mint}{rgb}{0.6902, 1.0, 0.7451}

\begin{table*}[ht]
\caption{Information on learnable parameters for each compared model.} 
    \centering
    \scalebox{0.75}{
        \begin{tabular}{llcc}
        \toprule
        \multicolumn{1}{c}{Method} &
        \multicolumn{1}{c}{Learnable Parameters} & 
        \multicolumn{1}{c}{$\approx$ Total Parameters}
        \\ \midrule
            \rowcolor{gray!5!white}
            Orca \cite{cao22orca} & $7.1 M$ \textcolor{red}{$f_{\theta}$}; $6.5 M$ \textcolor{pink}{Classification head} & $13.6 M$ \\ 
            \rowcolor{gray!5!white}
            GCD \cite{vaze2022generalized}   &  $7.1 M$ \textcolor{red}{$f_{\theta}$}; $23.1 M$ \textcolor{purple}{$\phi$}  & $30.2 M$ \\
            \rowcolor{gray!5!white}
            SimGCD \cite{wen2022simple} & $7.1 M$ \textcolor{red}{$f_{\theta}$}; $6.5 M$ \textcolor{pink}{Classification head} & $13.6 M$ \\
            \rowcolor{gray!5!white}
            Grow \& Merge \cite{zhang2022grow} & $7.1 M$ \textcolor{red}{$f_{\theta}$}; $23.1 M$ \textcolor{purple}{$\phi$}; $0.031 M$ \textcolor{teal}{Cluster head} & $30.2 M$ \\
            \rowcolor{gray!5!white}
            PA-CGCD \cite{kim2023proxy} & $7.1 M$ \textcolor{red}{$f_{\theta}$}; $0.077 M$ \textcolor{Magenta}{Proxy anchor}  & $7.2 M$ \\
            \rowcolor{gray!5!white}
            MetaGCD \cite{wu2023metagcd} &  $7.1 M$ \textcolor{red}{$f_{\theta}$}; $23.1 M$ \textcolor{purple}{$\phi$}  & $30.2 M$ \\
            \rowcolor{gray!5!white}
            PromptCCD-B~$w/$L2P (Ours) & $7.1 M$ \textcolor{red}{$f_{\theta}$}; $23.1 M$ \textcolor{purple}{$\phi$}; $0.046 M$ \textcolor{orange}{L2P} & $30.2 M$ \\
            \rowcolor{gray!5!white}
            PromptCCD-B~$w/$DP (Ours) & $7.1 M$ \textcolor{red}{$f_{\theta}$}; $23.1 M$ \textcolor{purple}{$\phi$}; $0.045 M$ \textcolor{orange}{DP} & $30.2 M$ \\ \hline
            \rowcolor{blue!3!white}
            PromptCCD (Ours)  & $7.1 M$ \textcolor{red}{$f_{\theta}$}; $23.1 M$ \textcolor{purple}{$\phi$}; $\{1537 * C\}$ \textcolor{blue}{GMP} & $30.3 M$ @ $C=100$ \\
            \rowcolor{blue!3!white} PromptCCD++ (Ours) & $7.1 M$ \textcolor{red}{$f_{\theta}$}; $23.1 M$ \textcolor{purple}{$\phi$}; $\{0.169M * P\}$  \textcolor{ForestGreen}{PLP}; $14.7 M$; \textcolor{violet}{Part router} ($t{=}0$ only) & $31.6M$ ($t{>}0$, $P{=}8$); $46.3M$ ($t{=}0$, $P{=}8$) \\
            \bottomrule
        \end{tabular}
        }
    \label{tab: parameter}
\end{table*}

L2P and DualPrompt \cite{wang2022learning, wang2022dualprompt} are prompt-based models designed for the supervised continual learning task. Both models freeze the backbone model and train the linear classifier in a supervised manner. Our CCD model $\mathcal{H}_{\theta}: \{\phi, f_{\theta}\}$ consists of $\phi$, an MLP projection head, and $f_{\theta}: \{f_e, f_b\}$ a transformer-based feature backbone that includes an input embedding layer $f_e$ and self-attention blocks $f_b$. 
During training, we optimize both the final block of $f_b$ and the projection head $\phi$. 
The projection head serves its purpose solely during contrastive learning and is omitted in the final categorization process, where only the features from $f_b$ are utilized.
Hence, freezing the backbone entirely is not feasible since it would keep the backbone unchanged even after training.
To validate the necessity of finetuning the final block of the backbone, we experiment with two frozen DINO models. 
The first model is the default frozen DINO backbone with no prompt module. For this model, we do not perform any training strategy and directly use it to extract $\hat{z}$ features. The second model is the frozen DINO backbone coupled with a learnable L2P prompt pool. For this model, we follow the exact training procedure similar to the baseline model but keep the backbone frozen. We compare these two frozen models with both our finetuned baseline and proposed models as shown in Tab.~\ref{tab: frozen}. By comparing the performance of the frozen models and the finetuned models, we can see that our finetuned model substantially outperforms the frozen models. 
Furthermore, based on the results obtained from finetuning our models on the CUB dataset, we observe that our models exhibit improved generalization compared to the DINO foundation model when applied to previously unseen datasets. This further validates the design choice of our method.

\definecolor{Gray}{gray}{0.9}
\definecolor{PaleBlue}{rgb}{0.7529, 0.9137, 0.9372}
\definecolor{BeauBlue}{rgb}{0.7686, 0.8470, 0.9529}
\definecolor{Mauve}{rgb}{0.8 , 0.7098, 0.9843}
\definecolor{PaleViolet}{rgb}{0.8156, 0.6431, 1.0}
\definecolor{Salmon}{rgb}{1.0, 0.8980, 0.6920}
\definecolor{Pink}{rgb}{1.0, 0.6902, 0.7908}
\definecolor{Mint}{rgb}{0.6902, 1.0, 0.7451}
\definecolor{cyan}{rgb}{0.906, 0.969, 0.965}
\definecolor{SoftP}{rgb}{0.945, 0.933, 0.949}

\begin{table}[ht]
    \caption{Comparison between the fully frozen models and the finetuned (final block) models.}
    \resizebox{\columnwidth}{!}{
    \centering
        \begin{tabular}{cl @{\hskip 0.1in} ccc @{\hskip 0.1in} ccc @{\hskip 0.1in} ccc @{\hskip 0.1in} ||ccc}
        \toprule
        \multicolumn{2}{c}{}
        & \multicolumn{3}{c}{Stage 1 \textit{ACC} (\%)}
        & \multicolumn{3}{c}{Stage 2 \textit{ACC} (\%)}
        & \multicolumn{3}{c}{Stage 3 \textit{ACC} (\%)}
        & \multicolumn{3}{c}{Average \textit{ACC} (\%)}
        \\ 
        \multicolumn{1}{l}{} &
        \multicolumn{1}{c}{Method} &
        \multicolumn{1}{c}{\textit{All}} & 
        \multicolumn{1}{c}{\textit{Old}} &
        \multicolumn{1}{c}{\textit{New}} &
        \multicolumn{1}{c}{\textit{All}} & 
        \multicolumn{1}{c}{\textit{Old}} & 
        \multicolumn{1}{c}{\textit{New}} &
        \multicolumn{1}{c}{\textit{All}} & 
        \multicolumn{1}{c}{\textit{Old}} & 
        \multicolumn{1}{c}{\textit{New}} &
        \multicolumn{1}{c}{\textit{All} \cellcolor{blue!7!white}} & 
        \multicolumn{1}{c}{\textit{Old}} & 
        \multicolumn{1}{c}{\textit{New}} 
        \\ 
        \cmidrule[0.1pt](r{0.80em}){1-2}%
        \cmidrule[0.1pt](r{0.80em}){3-5}%
        \cmidrule[0.1pt](r{0.80em}){6-8}%
        \cmidrule[0.1pt](r{0.80em}){9-11}%
        \cmidrule[0.1pt](r{0.80em}){12-14}%
        \rowcolor{gray!5!white}
        &Frozen DINO \cite{caron2021emerging}
        & $64.87$ & $71.43$ & $60.29$ 
        & $55.42$ & $66.67$ & $53.27$ 
        & $49.08$ & $66.19$ & $46.08$
        & $56.45$ \cellcolor{blue!7!white} & $68.10$ & $53.21$ \\
        \rowcolor{gray!5!white}
        &Frozen DINO $w/$L2P 
        & $65.08$ & $73.39$ & $59.26$ 
        & $55.43$ & $64.10$ & $53.69$ 
        & $49.52$ & $67.05$ & $46.17$
        & $56.67$ \cellcolor{blue!7!white} & $68.18$ & $53.04$ \\
        \rowcolor{gray!5!white}
        &PromptCCD-B~$w/$L2P (Ours) 
        & $65.93$  & $80.20$  & $55.94$ 
        & $56.72$  & $70.76$  & $54.04$ 
        & $51.55$  & $66.67$  & $48.90$ 
        & $58.07$ \cellcolor{blue!7!white} & $72.54$  & $52.96$ \\
        \rowcolor{blue!3!white}
        \rot{\rlap{C100}}
        &PromptCCD (Ours) 
        & $70.69$ & $80.90$ & $63.54$ 
        & $64.08$ & $73.14$ & $62.35$ 
        & $57.73$ & $72.67$ & $55.12$
        & $\textbf{64.17}$ \cellcolor{blue!7!white} & $75.57$ & $60.34$ \\ \midrule
        \rowcolor{gray!5!white}
        &Frozen DINO \cite{caron2021emerging}
        & $68.75$ & $71.90$ & $66.86$ 
        & $70.43$ & $73.57$ & $69.64$ 
        & $62.57$ & $74.29$ & $59.83$
        & $67.25$ \cellcolor{blue!7!white} & $73.25$ & $65.44$ \\
        \rowcolor{gray!5!white}
        &Frozen DINO $w/$L2P 
        & $76.71$ & $77.80$ & $75.77$ 
        & $64.33$ & $67.05$ & $63.24$ 
        & $63.70$ & $76.86$ & $61.40$
        & $68.24$ \cellcolor{blue!7!white} & $73.90$ & $66.80$ \\
        \rowcolor{gray!5!white}
        &PromptCCD-B~$w/$L2P (Ours) 
        & $75.80$  & $83.84$  & $70.17$ 
        & $70.95$  & $82.48$  & $68.75$ 
        & $61.19$  & $78.67$  & $58.13$ 
        & $69.31$ \cellcolor{blue!7!white} & $81.66$  & $65.68$ \\
        \rowcolor{blue!3!white}
        \rot{\rlap{IN-100}}
        &PromptCCD (Ours) 
        & $79.56$ & $84.24$ & $76.29$ 
        & $78.58$ & $79.71$ & $78.36$ 
        & $70.33$ & $81.33$ & $68.40$
        & $\textbf{76.16}$ \cellcolor{blue!7!white} & $81.76$ & $74.35$ \\ \midrule
        \rowcolor{gray!5!white}
        &Frozen DINO \cite{caron2021emerging}
        & $55.71$ & $65.00$ & $52.00$ 
        & $45.80$ & $56.79$ & $43.00$ 
        & $46.23$ & $55.85$ & $42.83$
        & $49.25$ \cellcolor{blue!7!white} & $59.21$ & $45.94$ \\
        \rowcolor{gray!5!white}
        &Frozen DINO $w/$L2P
        & $62.02$ & $66.31$ & $59.01$ 
        & $52.20$ & $61.00$ & $50.52$ 
        & $46.42$ & $54.81$ & $44.95$
        & $53.54$ \cellcolor{blue!7!white} & $60.70$ & $51.49$ \\
        \rowcolor{gray!5!white}
        &PromptCCD-B~$w/$L2P (Ours) 
        & $69.46$  & $75.24$  & $65.41$ 
        & $54.64$  & $65.43$  & $52.58$ 
        & $44.88$  & $59.43$  & $42.33$ 
        & $56.33$ \cellcolor{blue!7!white} & $66.70$  & $53.44$ \\
        \rowcolor{blue!3!white}
        \rot{\rlap{Tiny}}
        &PromptCCD (Ours) 
        & $68.67$ & $72.84$ & $65.76$ 
        & $59.69$ & $65.67$ & $58.55$ 
        & $57.16$ & $61.10$ & $56.47$
        & $\textbf{61.84}$ \cellcolor{blue!7!white} & $66.54$ & $60.26$ \\ \midrule
        \rowcolor{gray!5!white}
        &Frozen DINO \cite{caron2021emerging}
        & $41.60$ & $74.22$ & $31.37$ 
        & $31.27$ & $68.57$ & $23.23$ 
        & $44.77$ & $62.09$ & $37.99$
        & $39.21$ \cellcolor{blue!7!white} & $68.29$ & $30.86$ \\
        \rowcolor{gray!5!white}
        &Frozen DINO $w/$L2P
        & $40.25$ & $75.71$ & $28.40$ 
        & $30.63$ & $73.57$ & $21.52$ 
        & $45.99$ & $65.36$ & $38.44$
        & $38.95$ \cellcolor{blue!7!white} & $71.54$ & $29.45$ \\
        \rowcolor{gray!5!white}
        &PromptCCD-B~$w/$L2P (Ours) 
        & $56.65$  & $73.93$  & $45.11$ 
        & $47.75$  & $74.29$  & $42.12$ 
        & $47.32$  & $71.43$  & $42.62$ 
        & $50.57$ \cellcolor{blue!7!white} & $73.22$  & $43.28$ \\
        \rowcolor{blue!3!white}
        \rot{\rlap{CUB}}
        &PromptCCD (Ours) 
        & $57.08$ & $75.00$ & $45.11$ 
        & $47.38$ & $75.00$ & $41.52$ 
        & $61.89$ & $76.43$ & $59.05$
        & $\textbf{55.45}$ \cellcolor{blue!7!white} & $75.48$ & $48.56$ \\
        \bottomrule
        \end{tabular}
    }
    \label{tab: frozen}
\end{table}
\clearpage
\section{More Qualitative Results}
\label{supp: extra quali}

We further visualize the feature representation generated by our method on 
ImageNet-100 \cite{russakovsky2015imagenet}, TinyImageNet \cite{le2015tiny}, and CUB \cite{welinder2010caltech} datasets, using t-SNE \cite{van2008visualizing} to project the high-dimensional features of $\{D^l, D^{u}_{t}\}$ in each stage into a low-dimensional space. The qualitative visualization can be seen in Fig.~\ref{fig:supp_ProCCD_qual}, where data points of the same color indicate that the instances belong to the same category. Moreover, for stage $t > 0$, we only highlight the data points belonging to unknown novel categories. It is observed that across stages and datasets, our cluster features are discriminative.

\begin{figure}[h]
    \centering
    \includegraphics[width=1.0\linewidth]{fig/supp_ProCCD_qual.pdf}
    \caption{TSNE visualization of ImageNet-100, TinyImageNet, and CUB datasets with features from our PromptCCD on each stage.}
    \label{fig:supp_ProCCD_qual}
\end{figure}
\clearpage
\section{Broader Impacts and Limitations}
\label{supp: limitations}

Category discovery technologies can significantly impact various industries and applications, such as drug discovery and materials discovery. Our proposed framework has been shown to reduce forgetting while being robust enough to discover new classes in a continual learning setting. 
However, there may be potential negative social impacts, such as when the model learns improper prior knowledge or the data contains unwanted bias, leading to misinformation in society. 
Currently, there is still no reliable mechanism to prevent such situations from happening.
Therefore, having proper priors and managing data distribution is important to prevent the model from making corrupted predictions. 
Additionally, like other efforts on handling sequential unlabelled data, our system may accumulate errors over time as we do not have any specific regulation when dealing with longer time steps and potential categories with few samples at a given time step.

\end{document}